\documentclass{article}

\usepackage{PRIMEarxiv}
\usepackage{color}
\usepackage[utf8]{inputenc} % allow utf-8 input
\usepackage[T1]{fontenc}    % use 8-bit T1 fonts
\usepackage{hyperref}       % hyperlinks
\usepackage{url}            % simple URL typesetting
\usepackage{booktabs}       % professional-quality tables
\usepackage{amsfonts}       % blackboard math symbols
\usepackage{nicefrac}       % compact symbols for 1/2, etc.
\usepackage{microtype}      % microtypography
\usepackage{lipsum}
\usepackage{fancyhdr}       % header
\usepackage{graphicx}       % graphics
\graphicspath{{media/}}     % organize your images and other figures under media/ folder
\usepackage{amsmath}   % Core AMS mathematics package
\usepackage{amssymb}   % Provides additional math symbols, including \mathbb
\usepackage{amsfonts}  % Provides additional math fonts, often included by amssymb
\usepackage{subfigure}
\usepackage{algorithm}
\usepackage{algpseudocode}

\title{Multimodal atmospheric super-resolution with deep generative models}

 \author{
   Dibyajyoti Chakraborty \\
   Information Sciences and Technology \\
   The Pennsylvania State University \\
   University Park, Pennsylvania\\
   \texttt{d.chakraborty@psu.edu} \\
   \And
   Haiwen Guan \\
   Information Sciences and Technology \\
   The Pennsylvania State University \\
   University Park, Pennsylvania\\
   \texttt{hzg18@psu.edu} \\
   \And
   Jason Stock \\
   Environmental Science Division \\
   Argonne National Laboratory \\
   Lemont, Illinois \\
   \texttt{jason.stock@anl.gov} \\
   \And
   Troy Arcomano \\
   Environmental Science Division \\
   Argonne National Laboratory \\
   Lemont, Illinois \\
   \texttt{troya@allenai.org} \\
   \And
   Guido Cervone \\
   Department of Geography \\
   The Pennsylvania State University \\
   University Park, Pennsylvania\\
   \texttt{guc18@psu.edu} \\
   \And
   Romit Maulik \\
   Information Sciences and Technology \\
   The Pennsylvania State University \\
   University Park, Pennsylvania\\
   \texttt{rmaulik@psu.edu} \\
 }

\begin{document}
\maketitle

\begin{abstract}
Diffusion models are a class of generative machine learning algorithms that can be used to sample from complex distributions. They achieve this by learning a score function, i.e., the gradient of the log-probability density of the data, and reversing a noising process using the same. Once trained, these diffusion models not only generate new samples but also enable zero-shot conditioning of the generated samples on observed data. This promises a novel paradigm for data and model fusion, wherein the implicitly learned distributions of pretrained diffusion models can be updated given the availability of online data in a Bayesian formulation. In this article, we apply such a concept to the super-resolution of a high-dimensional dynamical system, given the real-time availability of low-resolution and experimentally observed sparse measurements from multimodal data.  Our experiments are performed for a super-resolution task that generates the ERA5 atmospheric dataset given sparse observations from a coarse-grained representation of the same and/or from unstructured experimental observations of the IGRA radiosonde dataset. We also perform a data fusion task that leverages predictions from a data-driven atmospheric emulator as well. We discover that the generative model can balance the influence of multiple dataset modalities during spatiotemporal state reconstructions. Additional analysis on how score-based sampling can be used for uncertainty estimates is also provided.
\end{abstract}

% keywords can be removed

\section{Introduction}

Several real-world forecasting applications involve high-dimensional dynamical systems that exhibit multiscale behavior. A classical example of such a system is the Earth's atmosphere which exhibits rich spatiotemporal behavior \cite{arakawa2011toward,neugebauer2008dynamics,tao2009multiscale}. The modeling and forecasting of such a system is complicated by the challenges involved in simulating and observing them. In the case of the atmosphere, a variety of instrumentation are leveraged to partially sense the atmosphere \cite{thepaut2003satellite,bouttier2001observing}. These partial observations are used to enhance understanding of atmospheric processes and improve forecasts, for example, by correcting dynamical models with parameterizations or improved initial conditions\cite{swinbank1994stratosphere,wang2000data,wang2000data,reichle2008data,law2012evaluating}. However, when spatial and temporal resolution are limited for sensing, an inverse problem emerges for recovering the full state of the dynamical system in the presence of partial observations. The dynamical models themselves, based on numerical discretizations of nonlinear partial differential equations, require vast computational resources for generating forecasts \cite{satoh2017outcomes}. Ultimately, the most accurate forecasting paradigm relies on data and model fusion, wherein first-principles-based numerical models are adaptively corrected in real-time within a Bayesian statistics formulation \cite{wikle2007bayesian,law2012evaluating}. In such an approach, the numerical models are assumed to provide a prior (or background) state which is updated given partial observations that are used to compute a likelihood. In this article, we investigate the aforementioned computational framework from the perspective of generative machine learning. Generative machine learning \cite{salakhutdinov2015learning}, in contrast with discriminative models, represents a paradigm where data is used to learn an underlying probability density function from which novel samples can be drawn (or `generated') \cite{gao2024generative,voina2024deep,regenwetter2022deep}. In particular, we continue to frame the concept of data and model fusion within a Bayesian setting but leverage deep learning for implicit learning and sampling from an underlying distribution that can be updated in real-time given sparse observations. 

Given a dataset sampled from an unknown distribution, the goal of a generative model is to approximate this distribution with a model that allows sampling, inference, or density estimation \cite{salakhutdinov2015learning}. Broadly, generative models fall into two categories: explicit density models, which define a likelihood function and aim to estimate it directly, and implicit density models, which focus on generating samples without requiring an explicit form of the likelihood. A state-of-the-art generative machine learning algorithm is the so-called score-based diffusion model \cite{song2021maximum,batzolis2021conditional,tashiro2021csdi,vahdat2021score} -- a focus of this study. In this approach, one starts with data (such as images) and gradually adds noise over time through a forward stochastic process—typically a diffusion process—until the data becomes nearly indistinguishable from pure Gaussian noise. This forward process is mathematically described using stochastic differential equations (SDEs), and it defines a continuous trajectory from clean data to random noise \cite{tang2024score}. It is important to note that score-based models learn to approximate, using a neural network, the score function—the gradient of the log-probability density of the data at various noise levels. This score function tells the model how to adjust a noisy sample to make it more likely under the data distribution. Finally, to generate new data, the model starts with a sample of pure noise and simulates the reverse-time SDE using the learned score function, effectively ``denoising" the sample step-by-step. In essence, a trained diffusion model implicitly learns an unknown distribution given collected samples. Furthermore, these models can also be used to adapt samples given sparse observations of the generated samples. This is achieved using `zero-shot' score matching \cite{li2023your,clark2023text}, where a pretrained score-based diffusion model is conditioned on new, sparse observations—without retraining the model. The core idea is to modify the sampling process using observations (constraints) to guide the generation, effectively sampling from a conditional distribution.

In this article, we develop a diffusion model for generating realistic states of a high-dimensional multiscale system (the atmosphere) from sampled noise. Subsequently, we utilize our pretrained model for rapidly generating samples that assimilate observations from coarse grids or from different sensor measurements. We outline specific contributions in the following
\begin{enumerate}
    \item We train a diffusion model using ERA5 reanalysis data \cite{Hersbach2020} to generate realistic samples corresponding to the global atmosphere for  13 variables on a 1.4-degree resolution grid. 

    \item We implement a stochastic sampler for our diffusion model that seamlessly performs multi-modal super-resolution and data fusion along with uncertainty quantification.
    
    \item We perform zero-shot score-matching using sparse observations to demonstrate a super-resolution task for full-state recovery with quantified uncertainty.

    \item We perform zero-shot score-matching using partial observations of different datasets, including radiosonde observations and a data-driven atmospheric emulator, to demonstrate a multi-fidelity super-resolution task with quantified uncertainty.
\end{enumerate}

Our results demonstrate that diffusion models can be used to generate high-dimensional states of dynamical systems (i.e., in a super-resolution task) from a variety of data sources without retraining using zero-shot score-matching. Moreover, we also demonstrate efficient `data-fusion' where zero-shot samples can balance the influence of multiple sources of data that are partially observed. This allows for the utilization of different sources of data available at different spatial resolutions for posterior updates during the generative sampling process.

\subsection{Related work}

In recent studies, deep learning-based models have provided exciting results in the state-reconstruction of high-dimensional multiscale dynamical systems \cite{fukami2019super,kelshaw2022physics,fukami2023super}. For example, in applications related to the reconstruction of turbulent flows and for magnetic resonance imaging data, customized physics-informed deep learning methods have led to the remarkable recovery of fine-scaled features from coarse-grained observations \cite{gao2021super,fathi2020super}. Most work in this emergent area of data-driven super-resolution has relied on structured grid representations of both the coarse and fine representation of the state which has allowed for the use of convolutional neural network techniques \cite{ren2023physr,fukami2024single}. For unstructured representations, graph neural network applications have also been used with significant success \cite{barwey2025mesh}. Methods that bridge convolutional architectures and arbitrary locations of sparse observations using Voronoi tessellation have also led to impressive state reconstructions \cite{fukami2021global}. However, a significant majority of these super-resolution demonstrations possess two limitations: first, they frequently rely on coarse-grid representations extracted from the fine state (i.e., they are obtained from the same dataset). Second, the proposed models need retraining when the source or representation of coarse-grid data is significantly changed for example when it is obtained from a different dataset. Finally, since super-resolution is an inverse problem, several deterministic techniques face limitations in deployments without some form of uncertainty quantification.

Some deep learning super-resolution methods that deploy Uncertainty Quantification (UQ) have leveraged probabilistic neural networks for quantifying aleatoric uncertainty \cite{maulik2020probabilistic} and deep ensembles for quantifying epistemic uncertainty \cite{maulik2023quantifying}. However, the former typically relies on the assumption of a Gaussian likelihood and the latter requires significant computational costs for the parallel training of multiple neural networks. Stochastic weight averaging \cite{izmailov2018averaging} is a competitive alternative for computationally efficient UQ in arbitrary deep learning tasks but requires restrictive assumptions on the use of the optimizer as well as the nature of the loss surface (requiring an approximately quadratic loss surface upon convergence) \cite{morimoto2022assessments}. This motivates the use of generative modeling for our present application.

The landscape of generative modeling for complex physical systems has been reshaped by the advent of deep learning, with score-based diffusion models emerging as a particularly potent paradigm. This evolution began with the foundational work on diffusion probabilistic models \cite{sohl2015deep}, which was significantly advanced by Denoising Diffusion Probabilistic Model (DDPM) \cite{ho2020denoising}, demonstrating its ability for high-fidelity image synthesis. Currently, the development of score-based generative models, which learn the log density gradient of the data distribution, provided a powerful alternative \cite{song2019generative}. These two perspectives were unified and generalized through the framework of stochastic differential equations (SDEs) \cite{song2020score}, establishing a continuous-time formulation for the diffusion and reverse-time generation processes. Further refinements to the design space, including preconditioning, noise scheduling, and sampling, led to Elucidating Diffusion Models (EDM), which achieved state-of-the-art results with improved efficiency \cite{karras2022elucidating}. More recently, Flow Matching has been introduced as a simulation-free training paradigm for continuous normalizing flows that is more stable and efficient, generalizing diffusion paths and allowing for novel transport paths, such as those based on optimal transport \cite{lipman2022flow, liu2022flow, albergo2023stochastic, tong2023improving}. These powerful generative priors have been adeptly applied to solve a wide range of inverse problems, often in a zero-shot or plug-and-play manner. Techniques such as Diffusion Posterior Sampling (DPS) \cite{chung2022diffusion} and Denoising Diffusion Restoration Models (DDRM) \cite{kawar2022denoising} enable sampling from the posterior distribution conditioned on measurements, facilitating tasks like super-resolution, inpainting, and deblurring without retraining the base model \cite{zhou2024enhancing,wu2024diffusion,dou2024diffusion}.

In this work, we utilize generative deep learning using EDM models to enable the super-resolution of high-dimensional states from dynamical systems from arbitrary coarse-grid observations without retraining. We also jointly provide UQ without the aforementioned restrictive assumptions through sampling from the generative process. Before proceeding, we review related work that has leveraged generative models for super-resolution of high-dimensional dynamical systems.

In the domain of atmospheric and climate science, these generative models are revolutionizing weather forecasting and data assimilation. Numerous studies have demonstrated their use for downscaling and super-resolution \cite{mardani2023generative, martinuu2024enhancing,srivastava2024precipitation,watt2024generative,tomasi2025can}, probabilistic and ensemble forecasting \cite{li2024generative, price2023gencast, andrae2024continuous,hua2024weather, schmidt2024spatiotemporally}, and data assimilation \cite{huang2024diffda, manshausen2024generative, yang2025generative}. Specifically, models like SEEDS \cite{li2024generative} have shown the ability to generate large, skillful ensembles for weather forecasting at a fraction of the computational cost of traditional methods. \cite{manshausen2024generative} demonstrates the viability of score-based data assimilation for incorporating sparse weather station data at kilometer scales, a task highly relevant to our work. Other similar applications include precipitation nowcasting \cite{asperti2025precipitation, gao2023prediff}, emulation of Earth system models (DiffESM - \cite{bassetti2024diffesm}), and generating climate simulations \cite{brenowitz2025climate}. These notable climate diffusion models along with FuXi-Extreme \cite{zhong2024fuxi}, DYffusion \cite{ruhling2023dyffusion} and Appa \cite{andry2025appa} underscore the transformative potential of generative models to handle the high dimensionality, uncertainty, and multiscale nature of atmospheric dynamics. In this study, we utilize diffusion models for downscaling from sparse observational data, while investigating the influence of multiple sources of data at different fidelities.

\section{Methods}
\label{sec:diffusion_models}

Diffusion models represent a new paradigm in generative modeling, achieving remarkable success in generating high-fidelity data across various modalities, including images, audio, and more \cite{ho2020denoising, song2020score, dhariwal2021diffusion}. These models are inspired by concepts from non-equilibrium thermodynamics \cite{sohl2015deep} and typically involve a two-stage process: a forward process that gradually injects noise into data, and a learned reverse process that aims to denoise and generate data. We develop the mathematical formulation of the various steps of the algorithm in the following.

\subsection{Forward and Reverse Diffusion Processes}

The forward diffusion process systematically corrupts an initial data sample $\mathbf{x}_0$ (drawn from the true data distribution $q(\mathbf{x})$) with noise. In the continuous-time formulation central to our work, this can be described by a forward stochastic differential equation (SDE) \cite{karras2022elucidating,song2020score}: 
\begin{equation}
    d\mathbf{x} = \mathbf{f}(\mathbf{x}, t)dt + g(t)d\mathbf{w}
\end{equation}
The drift term $\mathbf{f}(\mathbf{x}, t)$ governs the deterministic evolution of the data, while the diffusion term $g(t)$ scales the noise added via a Wiener process $d\mathbf{w}$. If the drift is zero, often referred to as the Variance Preserving (VP) SDE, the process simplifies to the addition of Gaussian noise with a specific standard deviation $\sigma$:
\begin{equation}
\mathbf{x}(\sigma) = \mathbf{x}_0 + \sigma \mathbf{n}
\label{eq:forward_noise_sigma}
\end{equation}
where $\mathbf{n} \sim \mathcal{N}(\mathbf{0}, \mathbf{I})$. The noise level $\sigma$ ranges from $\sigma_{\text{min}} \approx 0$ (no noise) to $\sigma_{\text{max}}$ (pure noise), at which point the data distribution $p(\mathbf{x}(\sigma_{\text{max}}))$ converges to a simple isotropic Gaussian prior. The generative power of these models comes from learning to reverse this process which has a corresponding reverse-time SDE \cite{anderson1982reverse}. Solving this reverse SDE requires the score function, $\nabla_{\mathbf{x}(\sigma)} \log p(\mathbf{x}(\sigma))$, which points in the direction of increasing data density. For the Gaussian noise model, a key relationship is:
\begin{equation}
S(x(\sigma),\sigma) = \nabla_{\mathbf{x}(\sigma)} \log p(\mathbf{x}(\sigma)) = -\frac{\mathbb{E}_{\mathbf{x}_0 | \mathbf{x}(\sigma)}[\mathbf{x}(\sigma) - \mathbf{x}_0]}{\sigma^2}
\label{eq:score_noise_relation}
\end{equation}
A neural network, $D_{\theta}$,  that estimates the clean data for different noise levels is used to indirectly approximate this score. If the network is trained to predict the clean data, $D_{\theta}(\mathbf{x}(\sigma); \sigma) \approx \mathbf{x}_0$, its output can be used to estimate the score:
\begin{equation}
\nabla_{\mathbf{x}(\sigma)} \log p(\mathbf{x}(\sigma); \sigma) \approx \frac{D_{\theta}(\mathbf{x}(\sigma); \sigma) - \mathbf{x}(\sigma)}{\sigma^2} = \mathbf{s}_{\theta}(x(\sigma),\sigma)
\label{eq:score_from_denoiser_x0}
\end{equation}

This allows the reverse process to be formulated as an ordinary differential equation (ODE), known as the probability flow ODE, which can be solved numerically to generate samples. 

\subsection{EDM Formulation and Training}\label{subsec:edm}

The Elucidating Diffusion Model (EDM) framework \cite{karras2022elucidating} provides a principled design space for diffusion models. A key aspect is network preconditioning, where the overall denoising model $F_{\theta}$ is defined as:
\begin{equation}
D_{\theta}(\mathbf{x}(\sigma); \sigma) = c_{\text{skip}}(\sigma) \mathbf{x}(\sigma) + c_{\text{out}}(\sigma) F_{\theta}\left(\frac{\mathbf{x}(\sigma)}{c_{\text{in}}(\sigma)}; c_{\text{noise}}(\sigma)\right)
\label{eq:edm_network_preconditioned_general}
\end{equation}
where $F_{\theta}$ is the neural network parameterized by $\theta$. The scaling factors ($c_{\text{skip}}, c_{\text{out}}, c_{\text{in}}, c_{\text{noise}}$) are crucial for ensuring the network operates on inputs with approximately unit variance. We use the specific coefficient definitions from Table 1 in \cite{karras2022elucidating}. The training objective is a weighted mean squared error:
\begin{equation}
L(\theta) = \mathbb{E}_{\mathbf{x}_0, \mathbf{n}, \sigma} \left[ \lambda(\sigma) || \mathbf{x}_0 - D_{\theta}(\mathbf{x}_0 + \sigma \mathbf{n}; \sigma) ||^2 \right]
\label{eq:edm_loss}
\end{equation}
where $\sigma$ is sampled from a distribution $p(\sigma)$ and $\lambda(\sigma)$ is a weighting function, typically $\lambda(\sigma) = 1/c_{\text{out}}(\sigma)^2$, that balances the loss contribution across different noise levels. Following Karras et al. (2022) \cite{karras2022elucidating}, we use $\sigma$ and $t$ interchangeably as $\sigma(t)=t$. 

\subsection{Sampling}
Once trained, the EDM model generates samples by numerically solving the probability flow ODE \cite{karras2022elucidating},
\begin{equation}
    \frac{d\mathbf{x}}{dt} = \frac{D_{\theta}(\mathbf{x}(t); t) - \mathbf{x}(t)}{t}
\end{equation}
Starting from a noise sample $\mathbf{x}_{t_{\text{max}}} \sim \mathcal{N}(\mathbf{0}, t_{\text{max}}^2 \mathbf{I})$, an ODE solver iteratively refines the sample across a sequence of decreasing noise levels $t_i$. A first-order Euler step from $\mathbf{x}(t_i)$ to $\mathbf{x}(t_{i+1})$ is:
\begin{enumerate}
    \item Predict denoised data: $\hat{\mathbf{x}}_0 = D_{\theta}(\mathbf{x}(t_i); t_i)$.
    \item Calculate drift: $\mathbf{d} = (\mathbf{x}(t_i) -\hat{\mathbf{x}}_0) / t_i$.
    \item Take step: $\mathbf{x}(t_{i+1}) = \mathbf{x}(t_i) + \mathbf{d} (t_{i+1} - t_i)$.
\end{enumerate}
More sophisticated samplers are often used to improve quality and efficiency \cite{karras2022elucidating, lu2022dpm}. Several applications require guiding generation with a condition $\mathbf{y}$ to sample from $p_{\theta}(\mathbf{x}|\mathbf{y})$. This is often done by feeding an embedding of $\mathbf{y}$ into the network $D_{\theta}(\mathbf{x}(t), t, \mathbf{y})$. However, it requires the conditions to be known a priori for training.

\subsection{Diffusion Posterior Sampling and Data Fusion}
\label{subsec:posterior_sampling_sda}
Another method to sample from the posterior $p_{\theta}(\mathbf{x}|\mathbf{y})$ is by combining a pre-trained generative (unconditional) prior $p_{\theta}(\mathbf{x})$ with a measurement likelihood $p(\mathbf{y}|\mathbf{x})$. From Bayes' theorem, the posterior score is the sum of the prior and likelihood scores:
\begin{equation}
\nabla_{\mathbf{x}(t)} \log p_{\theta}(\mathbf{x}(t)|\mathbf{y}) = \nabla_{\mathbf{x}(t)} \log p_{\theta}(\mathbf{x}(t)) + \nabla_{\mathbf{x}(t)} \log p(\mathbf{y}|\mathbf{x}(t))
\label{eq:posterior_score_main}
\end{equation}
This formulation is structurally analogous to guidance methods like CFG \cite{ho2022classifier}, where the prior score (from the unconditional model) is modified by a guidance term (the difference between conditional and unconditional scores). Here, the guidance term is the gradient of the log-likelihood. A key challenge is estimating this likelihood score $\nabla_{\mathbf{x}(t)} \log p(\mathbf{y} | \mathbf{x}(t))$.

For a differentiable measurement model $M$ and Gaussian observation noise with covariance $\Sigma_y$, DPS \cite{chung2022diffusion} approximates the likelihood $p(\mathbf{y} | \mathbf{x}(t))$ by first estimating the clean data $\hat{\mathbf{x}}_0(\mathbf{x}(t), t)$ from the noisy sample $\mathbf{x}(t)$ and then evaluating the likelihood at this estimate:
\begin{equation}
p(\mathbf{y} | \mathbf{x}(t)) \approx \mathcal{N} (\mathbf{y} | M(\hat{\mathbf{x}}_0(\mathbf{x}(t))), \Sigma_y)
\label{eq:likelihood_approx_dps}
\end{equation}
The estimate $\hat{\mathbf{x}}_0$ is obtained via Tweedie's formula \cite{efron2011tweedie}, which connects the posterior mean to the score of the noisy data distribution:
\begin{equation}
\hat{\mathbf{x}}_0(\mathbf{x}(t)) = \mathbb{E}[\mathbf{x}_0|\mathbf{x}(t)] \approx \frac{\mathbf{x}(t) + \sigma (t)^2 \mathbf{s}_{\theta}(\mathbf{x}(t), t)}{\mu(t)}
\label{eq:x0_hat_from_score}
\end{equation}
where $\mathbf{s}_{\theta}$ is the learned score model, and $p(x(t)|x_0) = \mathcal{N}(x(t)|\mu(t)x_0, \sigma(t)^2I)$. Score-based Data Assimilation (SDA) \cite{rozet2023score} refines this by accounting for the variance of the estimate $\hat{\mathbf{x}}_0$, yielding a more stable perturbed likelihood:
\begin{equation}
p(\mathbf{y} | \mathbf{x}(t)) \approx \mathcal{N} \left( \mathbf{y} \mid M(\hat{\mathbf{x}}_0), \Sigma_y + \frac{\sigma(t)^2}{(\mu(t))^2} \mathbf{J}_{M}(\hat{\mathbf{x}}_0) \mathbf{\Gamma} \mathbf{J}_{M}(\hat{\mathbf{x}}_0)^T \right)
\label{eq:sda_perturbed_likelihood_final_short}
\end{equation}
where $\mathbf{J}_{M}$ is the Jacobian of the measurement function and $\mathbf{\Gamma}$ is the term that depends on the eigen-decomposition of the covariance of prior $p(x)$, given by $\Sigma_x$. In practice, to simplify the approximation, the term $\mathbf{J}_{M}\mathbf{\Gamma} \mathbf{J}_{M}^T$
is replaced by a constant (diagonal) matrix (refer section 3.2 in \cite{rozet2023score}). 

\subsection{Diffusion Data Fusion for EDM}
In our work, we adapt the data fusion process for the EDM framework. As the network $F_{\theta}$ is trained to predict $\mathbf{x}_0$, the score is given by Eq. \ref{eq:score_from_denoiser_x0}. In the SDE convention used by EDM ($\mu(t)=1$), substituting this into Tweedie's formula (Eq. \ref{eq:x0_hat_from_score}) gives a direct and intuitive estimate for the clean data:
\begin{equation}
\hat{\mathbf{x}}_0 = D_\theta(\mathbf{x}(t); \sigma(t))
\end{equation}
The likelihood score, $\mathbf{s}_l$ is then computed based on the SDA approximation (Eq. \ref{eq:sda_perturbed_likelihood_final_short}), simplified for the EDM case:
\begin{equation}\label{eq:likelihood_Score_edm}
    \mathbf{s}_l(\mathbf{x}(t), \sigma(t); \mathbf{y}) = \nabla_{\mathbf{x}(t)} \log p(\mathbf{y}|\mathbf{x}(t)) \propto \nabla_{\mathbf{x}(t)} \frac{-||\mathbf{y} - M(D_\theta(\mathbf{x}(t);\sigma(t)))||^2}{\Sigma_y + \sigma(t)^2\hat{\Gamma}}
\end{equation}
where $\Sigma_y$ is the variance of observation noise and $\hat{\Gamma}$ approximates $\mathbf{J}_{M}\mathbf{\Gamma} \mathbf{J}_{M}^T$. These hyperparameters can be approximated as constants that are tuned based on the precision of the measurements. In highly non-linear measurements, a trade-off is observed between the desired accuracy and the quality of the generated sample. This likelihood gradient is then used to guide the standard EDM sampling process. For sampling, a first-order update can be given at each discrete timestep (or noise level) \textit{$t_i$} by the following steps:

\begin{enumerate}
    \item \textbf{Predict clean data:} $\hat{\mathbf{x}}_0 = D_\theta(\mathbf{x}_{t_i}; t_i)$.
    \item \textbf{Calculate prior drift:} $\mathbf{d}_{\text{prior}} = (\mathbf{x}_{t_i} - \hat{\mathbf{x}}_0 ) / t_i$.
    \item \textbf{Evaluate likelihood gradient:} $\mathbf{s}_l(\mathbf{x}_{t_i}, t_i; \mathbf{y})$ using Eq. \ref{eq:likelihood_Score_edm}.
    \item \textbf{Calculate posterior drift:} $\mathbf{d}_{\text{pos}} = \mathbf{d}_{\text{prior}} - \lambda_g \cdot t_i \cdot \mathbf{s}_l$, where $\lambda_g$ is a guidance scale and $t_i$ is multiplied to scale the drift term of the probability flow ODE.
    \item \textbf{Perform sampling step:} $\mathbf{x}_{t_{i+1}} = \mathbf{x}_{t_i} + \mathbf{d}_{\text{pos}} \cdot (t_{i+1} - t_i)$.
\end{enumerate}
This iterative procedure generates samples that adhere to both the learned data prior and the provided observations $\mathbf{y}$. Finally, we modify the stochastic sampler proposed in the EDM paper \cite{karras2022elucidating} to define Algorithm \ref{alg:stochastic_posterior_corrected}, which we employ in this work.

\begin{algorithm}[H]
\caption{Stochastic Posterior Sampler for data fusion}
\label{alg:stochastic_posterior_corrected}
\begin{algorithmic}[1]
\State \textbf{Require:} Pre-trained denoiser $D_\theta(\mathbf{x}; t)$, time steps $\{t_i\}_{i=0}^{N}$ where $t_N=0$, observation $\mathbf{y}$, guidance scale $\lambda_g$.
\State \textbf{Require:} Churn schedule parameters $S_{\text{churn}}, S_{t_{\min}}, S_{t_{\max}}$. (refer \cite{karras2022elucidating})
\State Sample $\mathbf{x}_{t_0} \sim \mathcal{N}(0, t_0^2 \mathbf{I})$
\For{$i = 0, \dots, N-1$}
    \State $\gamma_i \leftarrow \min\left(\frac{S_{\text{churn}}}{N}, \sqrt{2}-1\right) \text{ if } t_i \in [S_{t_{\min}}, S_{t_{\max}}] \text{ else } 0$ \Comment{Determine amount of noise injection (churn)}
    \State $\hat{t}_i \leftarrow t_i + \gamma_i t_i$
    \State Sample $\boldsymbol{\epsilon}_i \sim \mathcal{N}(0, \mathbf{I})$
    \State $\mathbf{x}_{\hat{t}_i} \leftarrow \mathbf{x}_{t_i} + \sqrt{\hat{t}_i^2 - t_i^2} \boldsymbol{\epsilon}_i$ \Comment{Add noise to move from $t_i$ to $\hat{t}_i$}
    % \State $\hat{\mathbf{x}}_0^{\text{pred}} \leftarrow D_\theta(\mathbf{x}_{\hat{t}_i}; \hat{t}_i)$
    \State $\mathbf{d}_{\text{prior}}^{\text{pred}} \leftarrow (\mathbf{x}_{\hat{t}_i} - D_\theta(\mathbf{x}_{\hat{t}_i}; \hat{t}_i)) / \hat{t}_i$
    \State $\mathbf{s}_l^{\text{pred}} \leftarrow \nabla_{\mathbf{x}_{\hat{t}_i}} \log p(\mathbf{y} | \mathbf{x}_{\hat{t}_i})$ \Comment{Evaluate likelihood gradient using Equation \ref{eq:likelihood_Score_edm}}
    \State $\mathbf{d}_{\text{pos}}^{\text{pred}} \leftarrow \mathbf{d}_{\text{prior}}^{\text{pred}} - \lambda_g \cdot \hat{t}_i \cdot \mathbf{s}_l^{\text{pred}}$
    \State $\mathbf{x}_{t_{i+1}}^{\text{pred}} \leftarrow \mathbf{x}_{\hat{t}_i} + \mathbf{d}_{\text{pos}}^{\text{pred}} \cdot (t_{i+1} - \hat{t}_i)$ \Comment{Initial Euler step}
    \If{$t_{i+1} \neq 0$}
        % \State $\hat{\mathbf{x}}_0^{\text{corr}} \leftarrow D_\theta(\mathbf{x}_{t_{i+1}}^{\text{pred}}; t_{i+1})$
        \State $\mathbf{d}_{\text{prior}}^{\text{corr}} \leftarrow (\mathbf{x}_{t_{i+1}}^{\text{pred}} - D_\theta(\mathbf{x}_{t_{i+1}}^{\text{pred}}; t_{i+1})) / t_{i+1}$
        \State $\mathbf{s}_l^{\text{corr}} \leftarrow \nabla_{\mathbf{x}_{t_{i+1}}^{\text{pred}}} \log p(\mathbf{y} | \mathbf{x}_{t_{i+1}}^{\text{pred}})$ \Comment{Re-evaluate likelihood gradient using Equation \ref{eq:likelihood_Score_edm} (optional)}
        \State $\mathbf{d}_{\text{pos}}^{\text{corr}} \leftarrow \mathbf{d}_{\text{prior}}^{\text{corr}} - \lambda_g \cdot t_{i+1} \cdot \mathbf{s}_l^{\text{corr}}$
        \State $\mathbf{x}_{t_{i+1}} \leftarrow \mathbf{x}_{\hat{t}_i} + \frac{1}{2}(\mathbf{d}_{\text{pos}}^{\text{pred}} + \mathbf{d}_{\text{pos}}^{\text{corr}}) \cdot (t_{i+1} - \hat{t}_i)$ \Comment{2nd-order correction step}
    \Else
        \State $\mathbf{x}_{t_{i+1}} \leftarrow \mathbf{x}_{t_{i+1}}^{\text{pred}}$
    \EndIf
\EndFor
\State \textbf{return} $\mathbf{x}_{t_N}$
\end{algorithmic}
\end{algorithm}

\section{Experimental configuration}
\vspace{-0.1cm}
\subsection{Motivation: Atmospheric data assimilation}
\vspace{-0.1cm}
Data assimilation for atmospheric science has traditionally been performed by using a short-term forecast from a numerical-based model (e.g., weather forecasting or climate models) as the prior, which is then updated with observations using numerical techniques (e.g., Kalman filtering or variational methods) to provide a best estimate of the atmospheric (or ocean, land surface, etc.) state at a particular time on a regular grid \cite{Szunyogh2014}. This helps overcome the problem with observations (e.g., from satellites, weather stations, buoys) that are irregular in time and space, noisy, and often incomplete.  Data assimilation combines the strengths of both: observational accuracy with model-based physical consistency and allows for the estimation of the state on a regular grid. This approach has been successful and sits at the heart of modern-day weather prediction. One particularly important use case of data assimilation has been the ability to generate physically consistent, comprehensive, and temporally continuous depiction of the Earth system, call reanalysis (e.g., ERA5) \cite{Hersbach2020}. However, classical methods for generating reanalysis data from observations and a numerical model tend to be computationally expensive require significant use of high-performance computing resources. This can be seen with ERA5 which required massive amounts of computational resources over multiple years to produce the nearly 80 years of reanalysis data \cite{Hersbach2020}. If machine learning could produce similar quality estimates of the atmosphere using observations, this could represent orders of magnitude improvement in computational efficiency for data assimilation.

\vspace{-0.1cm}
\subsection{Data}
\vspace{-0.1cm}
We use ERA5, a state-of-the-art global reanalysis dataset produced by the European Centre for Medium-Range Weather Forecasts (ECMWF), which provides hourly estimates of a large number of atmospheric, land, and oceanic climate variables at a resolution of $\approx$31 km \cite{Hersbach2020}. For this work, we use the 1.40525-degree (128 × 256 grid points) ERA5 dataset from WeatherBench \cite{Weatherbench1}, for which we used bilinear interpolation to regrid from the native resolution. The choice to use a lower resolution version of ERA5 was due to the computational limitations of training an unconditional diffusion model without large scale compute access.

For observational data, we rely on the Integrated Global Radiosonde Archive (IGRA), a comprehensive dataset of weather balloon observations from several stations worldwide, providing upper-air measurements like temperature, humidity, geopotential, pressure, and wind at different pressure levels \cite{durre2006overview}. Furthermore, we also construct an observation data set that is a further coarsened ERA5 at 11.242°(16x32 grid points) using bicubic interpolation. This is mentioned as `LR' further in the results.

Next, we also assume access to a dynamical core comprised of a machine learning emulator for the atmosphere that can be used for guiding posterior sampling. We employ the Lightweight Uncoupled Climate Emulator (LUCIE) \cite{guan2024lucie}, which was trained on ERA5 reanalysis covering the period 1980–2000 at 3.75$^\circ$(48x96 grid points). LUCIE is trained on the same set of variables used in the diffusion model, with the addition of top-of-atmosphere incident solar radiation as an external forcing variable and makes forecasts at 6 hour resolutions. We emphasize that this dynamical core is a \emph{climate model}, and therefore represents a source that generates observations from a fundamentally different platform than weather models. This is exemplified, for example, by the extremely coarse grid on which variables are resolved.

\vspace{-0.1cm}
\subsection{Measurement Operators}
\vspace{-0.1cm}
For our super-resolution tasks,  we assume access to data on a coarse or sparse and unstructured grid from which the full state must be reconstructed. A first experiment involves the artificial construction of such data, from the ground truth, using a measurement operation. This measurement operation involves a downsampling operator that reduces the resolution of the data using interpolation techniques such as bicubic interpolation. These downsampled values are compared with the available low-resolution data and used to guide our diffusion model. We mention these samples as `Sample-LR'. In contrast, real-world data fusion tasks may involve irregular or sparse observational locations. Consequently, our IGRA operator interpolates values at arbitrary spatiotemporal coordinates (e.g., latitude and longitude) using grid-based sampling. It defines a non-uniform measurement operator $M$ that evaluates the high-resolution field at the queried coordinates via bicubic interpolation in normalized coordinate space. This enables the comparison between model predictions and sparse ground truth measurements, forming the basis of our diffusion data assimilation frameworks. We mention these samples as `Sample-IGRA'.  For a first multimodal data fusion experiment, where we are performing super-resolution given access to both LR and IGRA information, we compute the misfit for both low-resolution and the IGRA measurements and use a weighted combination of the misfits to guide our diffusion sampling. We mention these samples as `Sample IGRA+LR' henceforth. For instance, the operator $M$ in Equation \ref{eq:likelihood_Score_edm} can be the IGRA operator, which takes the generated full field as input and returns the corresponding values at the locations where we have the true IGRA measurements $\mathbf{y}$.

Next, we detail how observations from the LUCIE atmospheric emulator may be introduced as an additional likelihood to guide posterior sampling while also observing IGRA measurements. In particular, the combined score function can be obtained by adding the individual scores with a hyperparameter $\lambda$ to adjust the weights given to the IGRA score.
\begin{equation}\label{eq:likelihood_Score_lucie}
    \mathbf{s}_{IGRA+LUCIE} = \nabla_{\mathbf{x}(t)} \left(\frac{-||\mathbf{y}_{L} - M_{L}(D_\theta(\mathbf{x}(t);\sigma(t)))||^2}{\Sigma_{y_L} + \sigma(t)^2\hat{\Gamma_L}} + \lambda  \frac{-||\mathbf{y}_{I} - M_{I}(D_\theta(\mathbf{x}(t);\sigma(t)))||^2}{\Sigma_{y_{I}} + \sigma(t)^2\hat{\Gamma_{I}}}\right)
\end{equation}
where the subscript $L$ stands for LUCIE and $I$ stands for IGRA in each term. This likelihood score is used in Algorithm \ref{alg:stochastic_posterior_corrected}. Optimal choices of the $\lambda$ parameter require an ablation study which is provided in Section \ref{sec:multimodal_sr}, Table \ref{tab:lambda_ablation}.

\subsection{Network architecture and hyperparameters}
\vspace{-0.1cm}
We used the architecture popularly called SongUNet \cite{song2020score} as our $D_\theta$ in Equation \ref{eq:edm_network_preconditioned_general}. The hyperparameters used for the model include a base embedding dimension of 64, channel multipliers of [1, 2, 4, 8, 16], and attention applied at resolutions [32, 64], [16, 32], and [8, 16], four residual blocks per resolution level, dropout rate of 0.10, and sinusoidal timestep embedding. We used circular padding along the x-axis and zero padding along the y-axis. For the EDM hyperparameters, we used a P$_{mean} = - 0.5$ and P$_{std} =1.5$ and $\sigma_{data} = 1$ (refer to \cite{karras2022elucidating} for details of these). Using a batch size of 512 we trained for approximately 35 million images in total. The values of $\Sigma_y$ and $\hat{\Gamma}$ in equation \ref{eq:likelihood_Score_edm} are selected as $1\times10^{-1}$ and $5 \times 10^{-3}$ for super-resolution using the LR dataset and $5\times 10^{-4}$ and $1\times10^{-5}$ for the assimilation of IGRA data, respectively. These values can be tuned based on the reduction of the negative log probability of the likelihood term while sampling. We perform an ablation study for these hyperparameters in Table \ref{tab:rmse_era5_ablation} for the super-resolution from structured coarse-grained fields case (Section \ref{sec:SR}). The value of $\lambda_g$ in Algorithm \ref{alg:stochastic_posterior_corrected} is set to 1 in all our experiments. We also use gradient clipping to stabilize sampling.  Other hyperparameters not explicitly mentioned here are obtained from the EDM paper \cite{karras2022elucidating}.

\begin{table}[!ht]
\centering
\begin{tabular}{ccc}
\hline
$\Sigma_y$ & $\hat{\Gamma}$ & 2m Temp RMSE (Mean $\pm$ Std)\\
\hline
$1\times10^{-1}$ & $1\times10^{-2}$ & 2.3511 $\pm$ 0.1246 \\
$1\times10^{-1}$ & $1\times10^{-3}$ & 1.7273 $\pm$ 0.0529 \\
$1\times10^{-1}$ & $5\times10^{-3}$ & \textbf{1.6883 $\pm$ 0.0224} \\
$5\times10^{-1}$ & $1\times10^{-3}$ & 2.3673 $\pm$ 0.0834 \\
$5\times10^{-1}$ & $5\times10^{-3}$ & 2.3359 $\pm$ 0.1129 \\
$5\times10^{-2}$ & $5\times10^{-3}$ & 54.4904 $\pm$ 7.0866 \\
\hline
\end{tabular}
\caption{RMSE of 2m temperature from ERA5 for different $\Sigma_y$ and $\hat{\Gamma}$ settings. Mean $\pm$ Std is reported. Lowest RMSE mean is shown in \textbf{bold}.}
\label{tab:rmse_era5_ablation}
\end{table}

\subsection{Comparisons to similar frameworks}

Our proposed framework is conceptually similar to the Appa-Weather approach \cite{andry2025appa} which also uses EDM-based posterior sampling. However, there are a few key differences in the present study. First, we study the assimilation of IGRA observations, as well as AI-emulator forecasts, in the reanalysis reconstruction state representing a data and model fusion exercise that leverages multiple sources of data generation that leverage both static and dynamical information about the evolving atmospheric state. Secondly, the emphasis of this paper is to assess the relative importance of these data in both the spatial and temporal sense while performing reconstructions. Compared to DiffDA \cite{huang2024diffda}, our approach is unconditional; therefore, our diffusion model does not need to be re-trained when new data or dynamical states become available. We also acknowledge previous efforts for data assimilation using diffusion models using observation data \cite{manshausen2024generative}. However, the notable difference in our work is the ability to perform multimodal data fusion with a low-resolution dynamical core and IGRA observations to obtain temporally consistent trajectories. Our approach, therefore, is a novel combination of super-resolution and data assimilation across multiple sources and resolutions of atmospheric observations.

\section{Results and discussion}

We outline results from our various experiments in the following. The training of the score-based generative model required approximately 30 hours on 16 Nvidia A100 GPUs, while generating a sample conditionally or with observation of sparse information requires only 2 seconds per ensemble on 1 A100 GPU. The results are generated for the year 2020 which is unseen data for training. We use Root Mean Square Error (RMSE) as the metric for comparing the diffusion samples with ERA5 and also for ablation studies.

\subsection{Super-resolution from structured coarse-grained fields}\label{sec:SR}
\vspace{-0.1cm}
\begin{figure}[!ht]
    \mbox{
    \subfigure[500mb Geopotential ($m^2/s^2$)]{\includegraphics[width=0.5\textwidth]{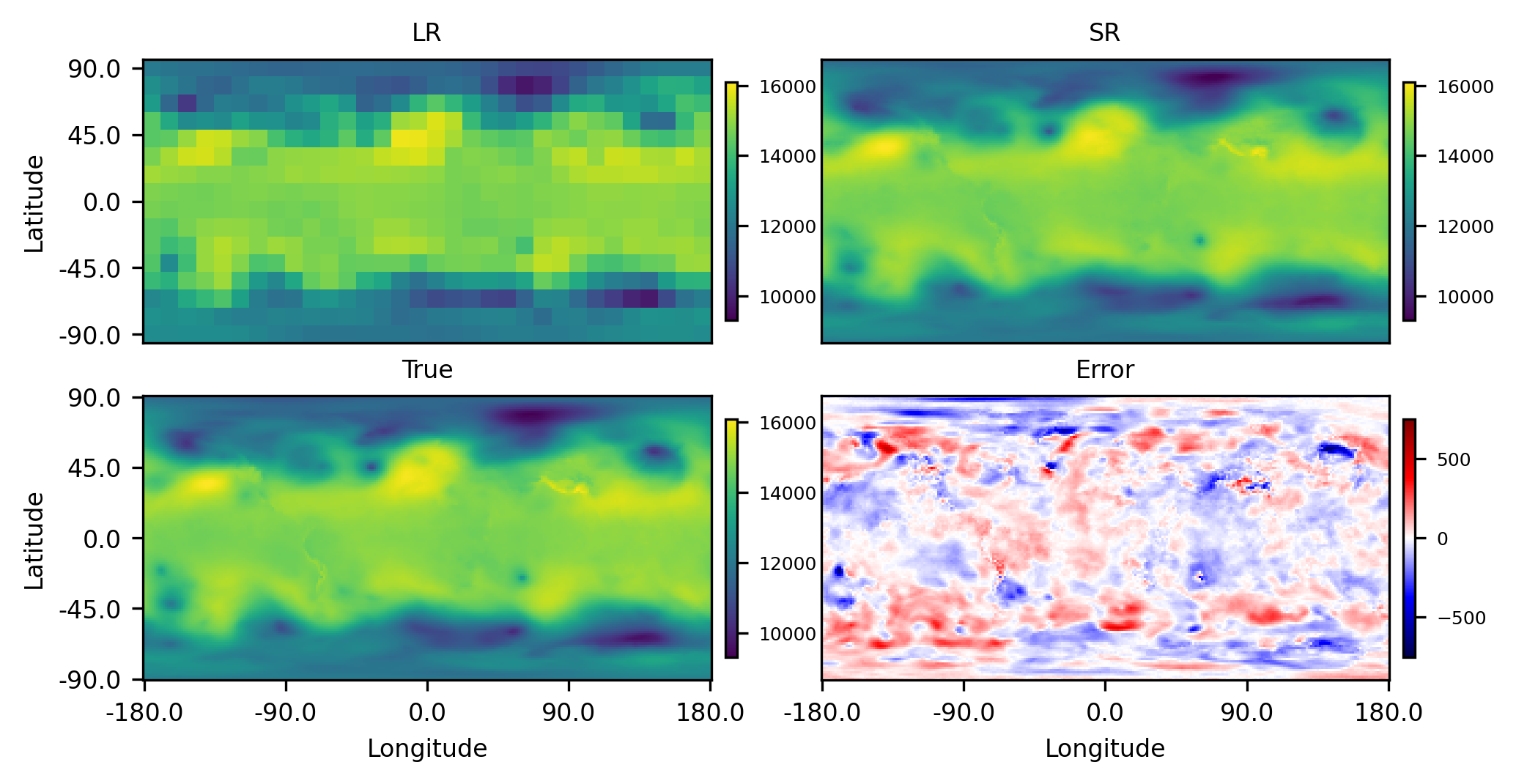}}
    \subfigure[10m U-Wind (m/s)]{\includegraphics[width=0.5\textwidth]{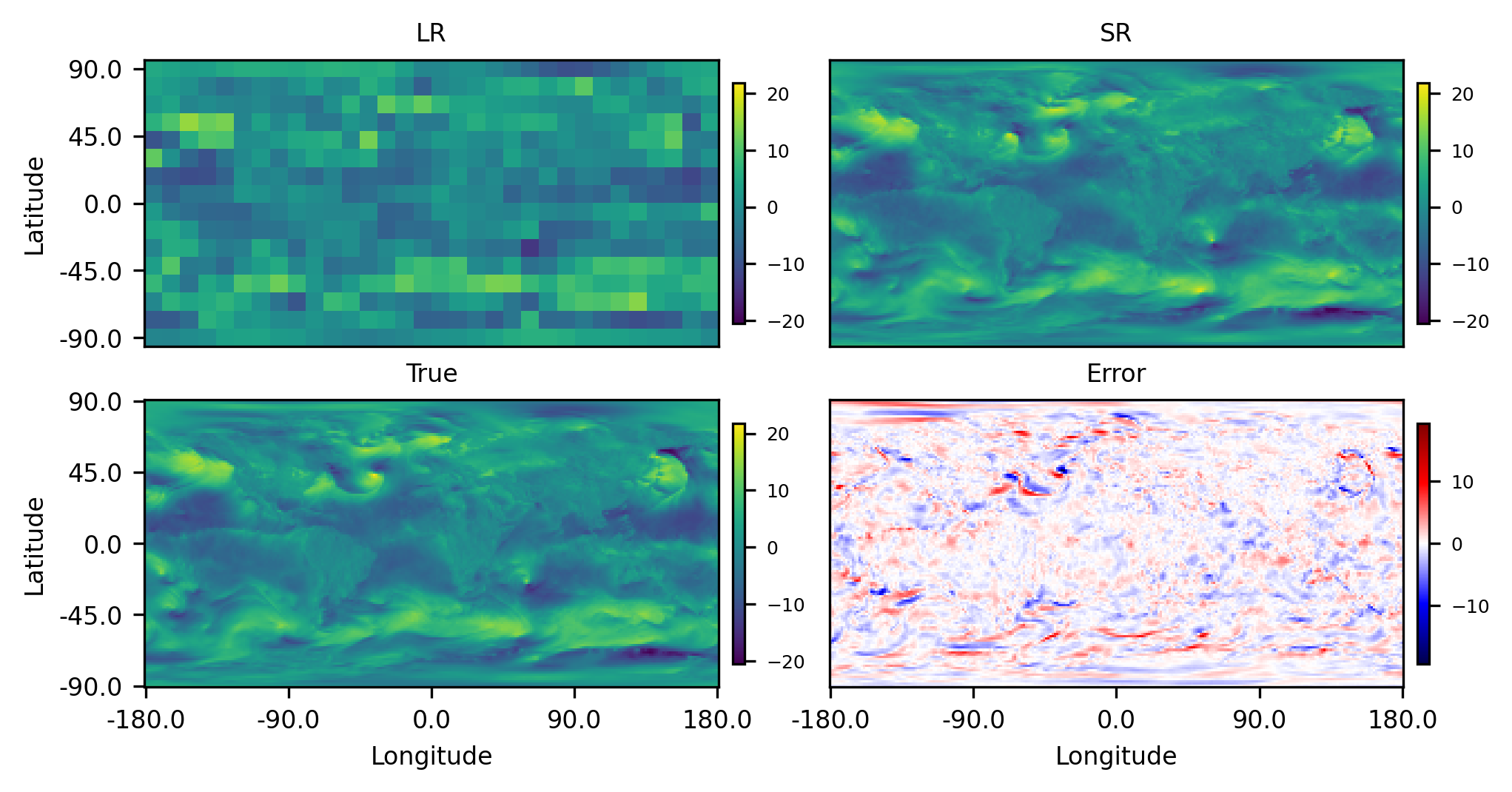}}
    } \\
    \mbox{
    \subfigure[10m V-Wind (m/s)]{\includegraphics[width=0.5\textwidth]{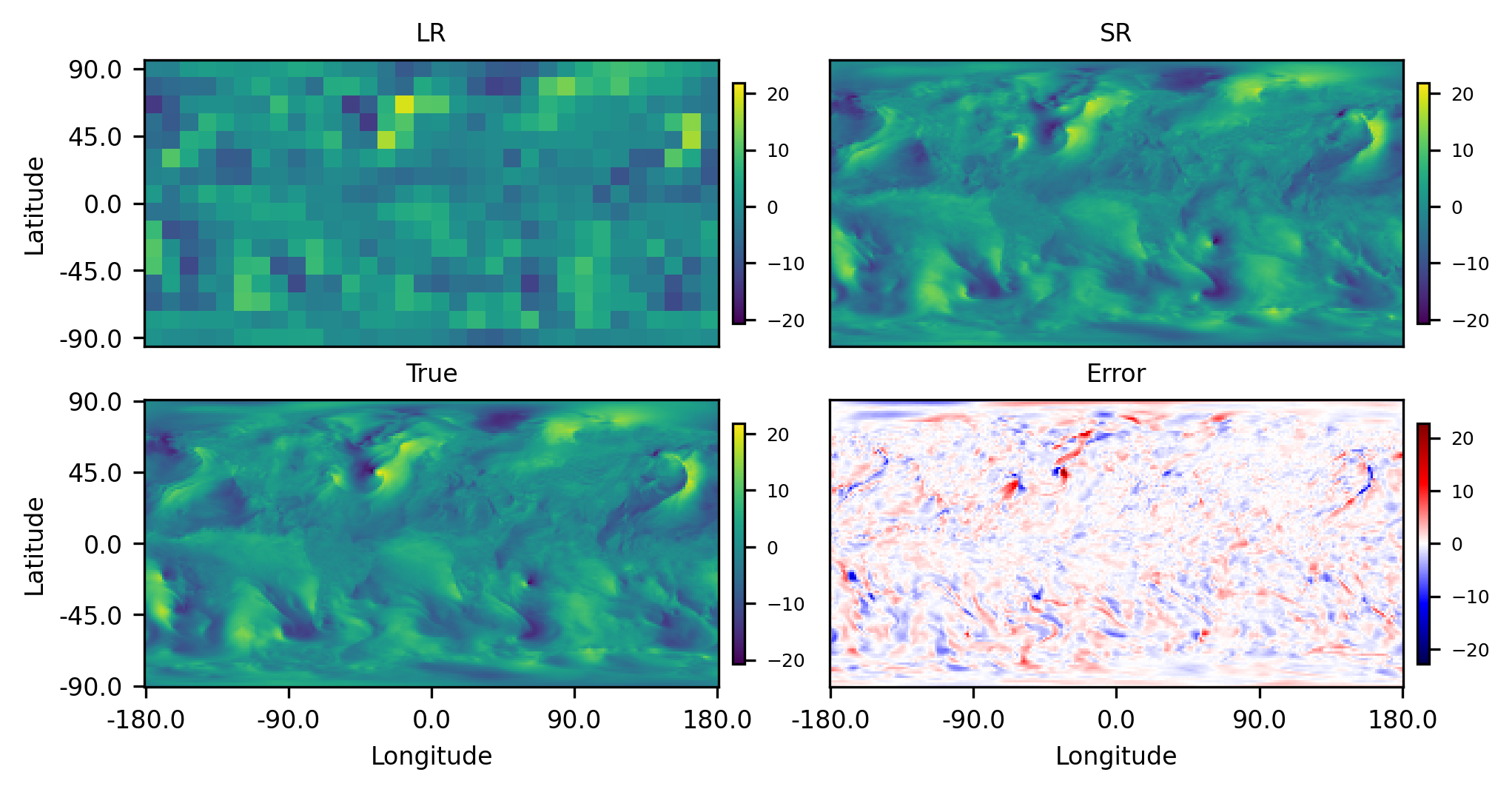}}
    \subfigure[850mb Specific humidity (kg/kg)]{\includegraphics[width=0.5\textwidth]{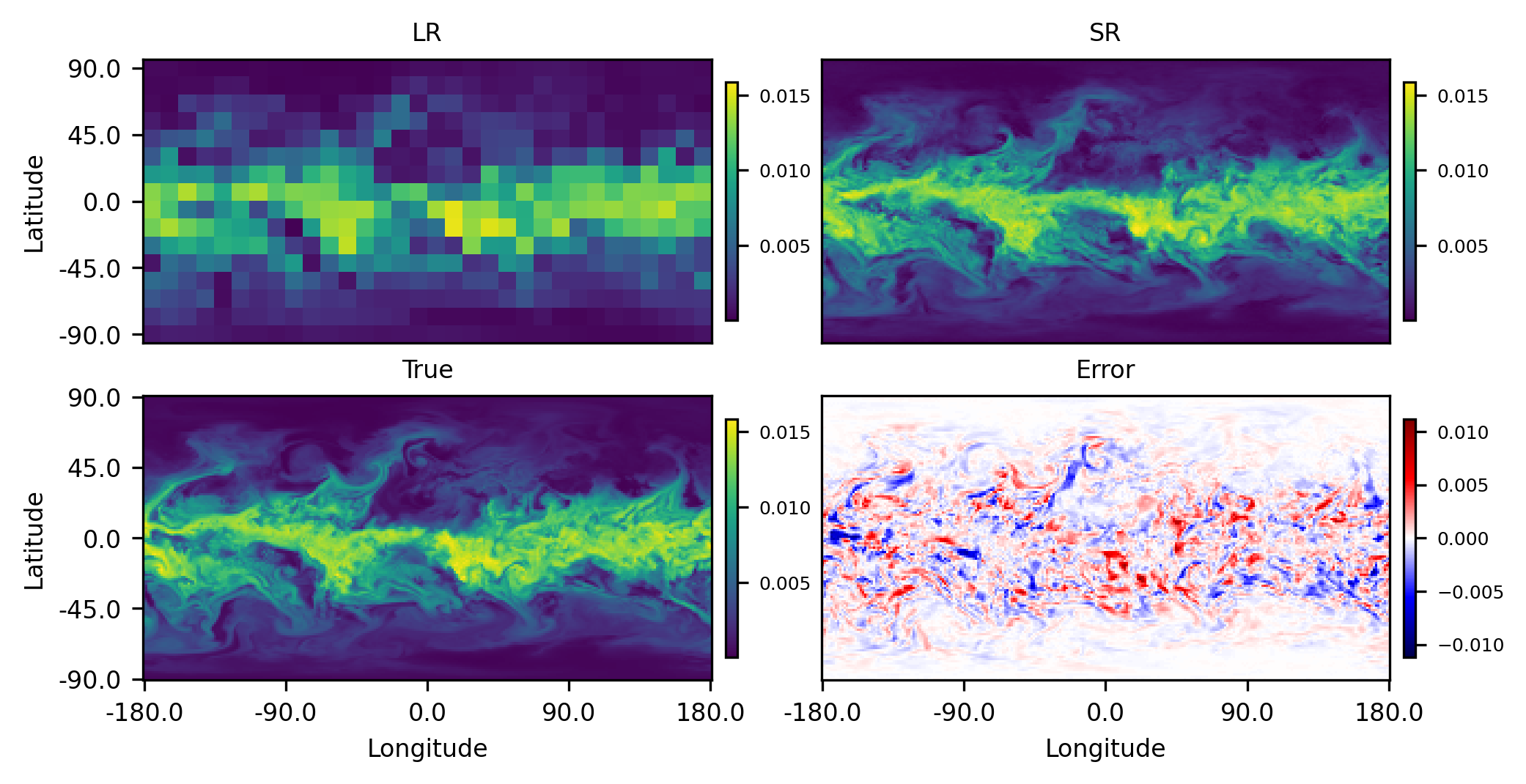}}
    }
    \centering
    \caption{Super-resolution using zero-shot posterior sampling given structured grid low-resolution observations (sample-LR). Contours showing mean of samples at specific time instances.}
    \label{fig:lr_hr_sr}
\end{figure}

\begin{figure}
    \centering
    \includegraphics[width=0.99\linewidth]{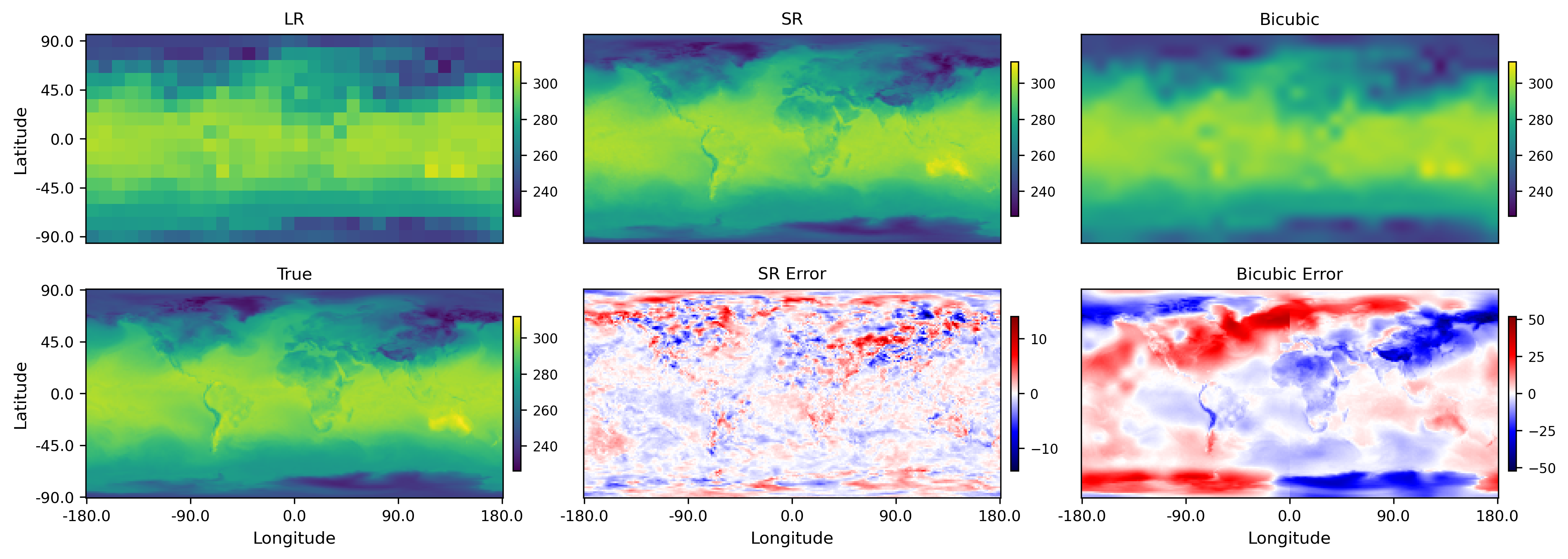}
    \caption{Comparison of 2m temperature fields from low-resolution input (LR), super-resolution using zero-shot posterior sampling (SR), and bicubic super resolution. The bottom row shows the true field and the corresponding error maps for SR and bicubic methods relative to the true field.}
    \label{fig:bicubic1}
\end{figure}

\begin{figure}
    \centering
    \includegraphics[width=0.99\linewidth]{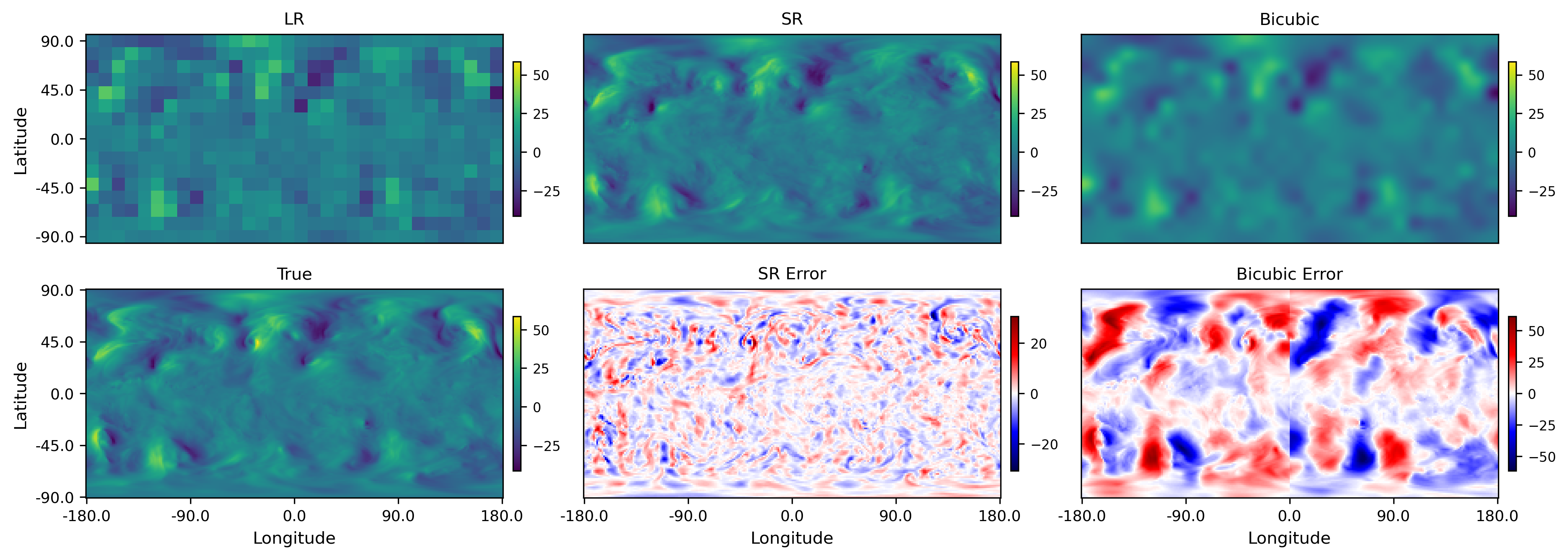}
    \caption{Comparison of 500mb V-wind fields from low-resolution input (LR), super-resolution using zero-shot posterior sampling (SR), and bicubic interpolation. The bottom row shows the true field and the corresponding error maps for SR and bicubic methods relative to the true field.}
    \label{fig:bicubic2}
\end{figure}

\begin{figure}[!ht]
\vspace{-0.3cm}
    \centering
    \includegraphics[trim={40 200 30 250},clip,width=0.8\textwidth]{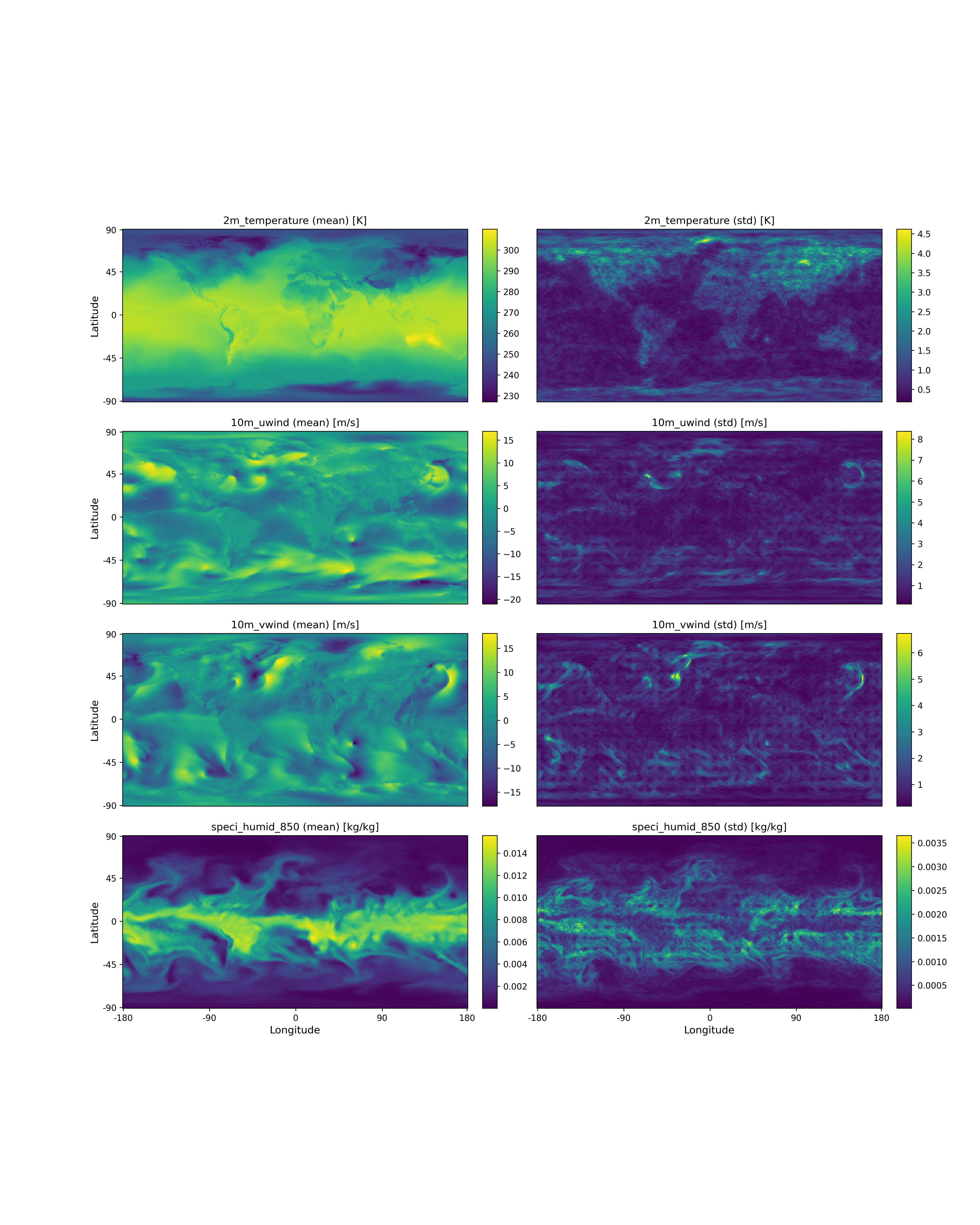}
    \vspace{-0.3cm}
    \caption{Ensemble based mean (left) and standard deviations (right) obtained from ensembles obtained with sample-LR zero-shot sampling. Showing 2m temperature, 10m u component of wind, 10m v component of wind, and specific humidity at 850mb pressure from top to bottom. One can observe a fine pattern of low uncertainty on a uniform grid, which corresponds to locations for the coarsely sampled ERA5.}
    \label{fig:lr_hr_uq}
\end{figure}
\begin{figure}[!h]
\vspace{-0.35cm}
    \centering
    \mbox{
    \subfigure[2m Temperature]{\includegraphics[width=0.45\textwidth]{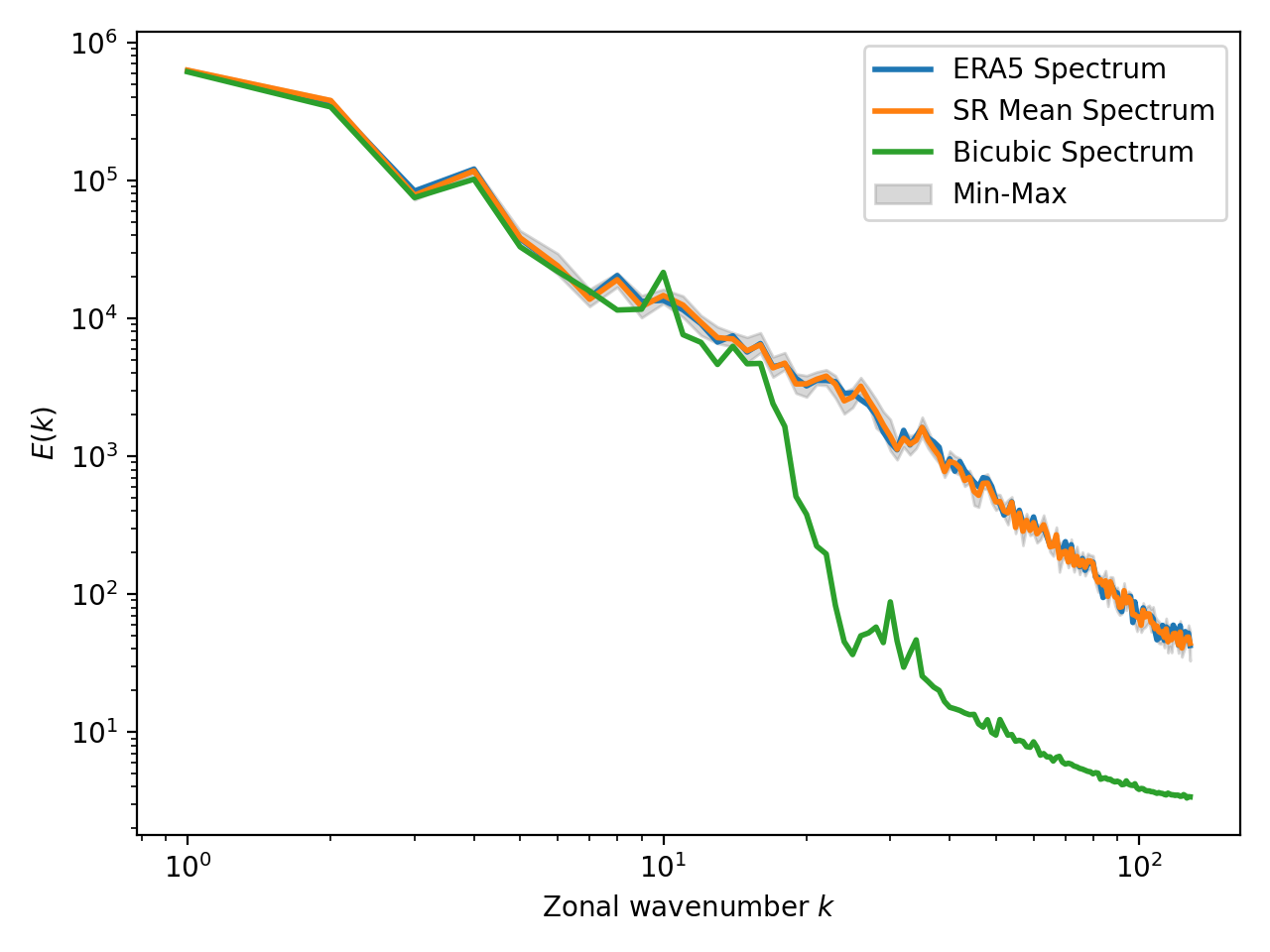}}
    \subfigure[10m u-wind]{\includegraphics[width=0.45\textwidth]{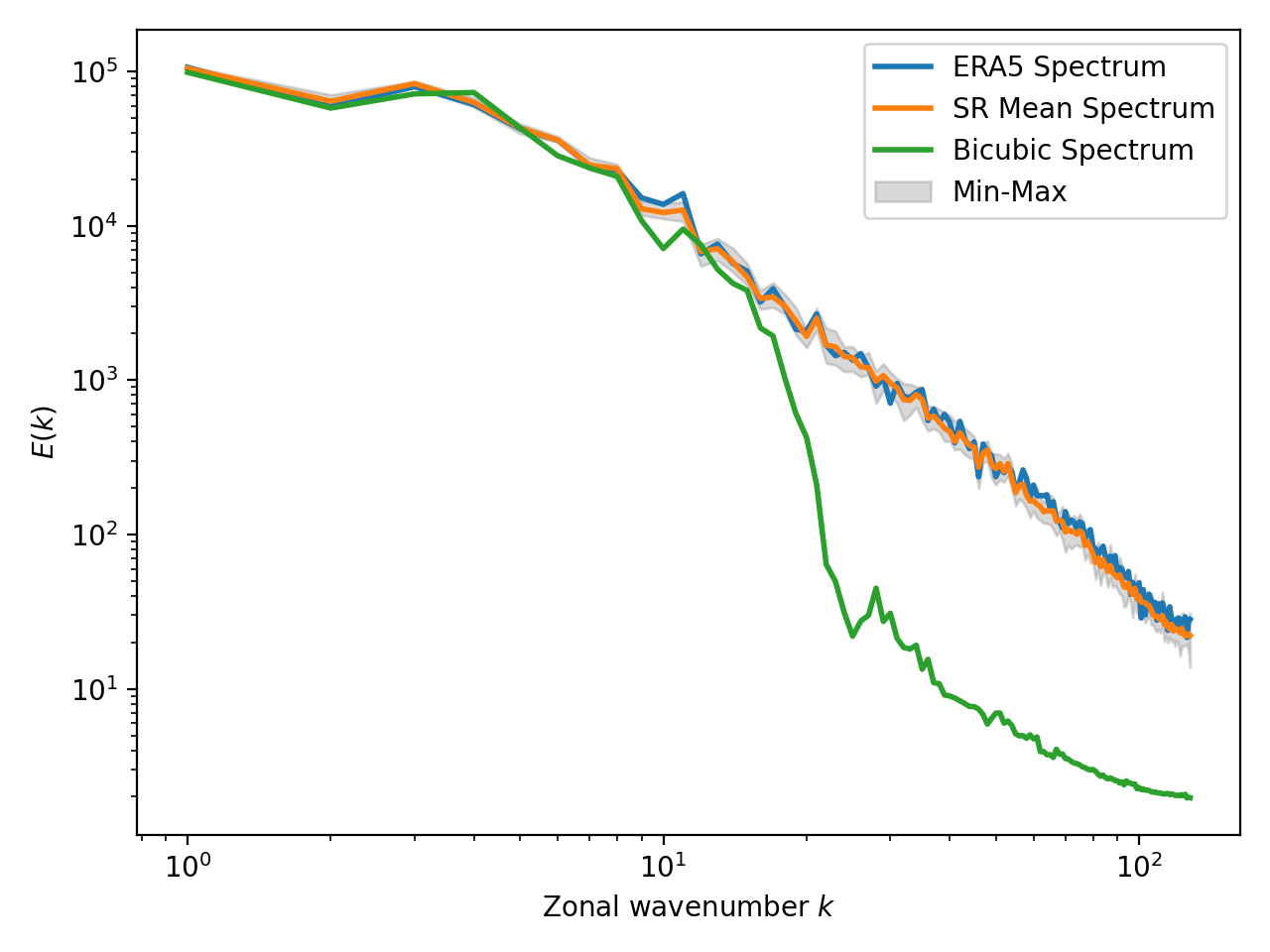}}
    }\\
    \mbox{
    \subfigure[10m v-wind]{\includegraphics[width=0.45\textwidth]{images_2020/LR_to_HR/SR_spec_10m_u_component_of_wind.png}}
    \subfigure[850mb specific humidity]{\includegraphics[width=0.45\textwidth]{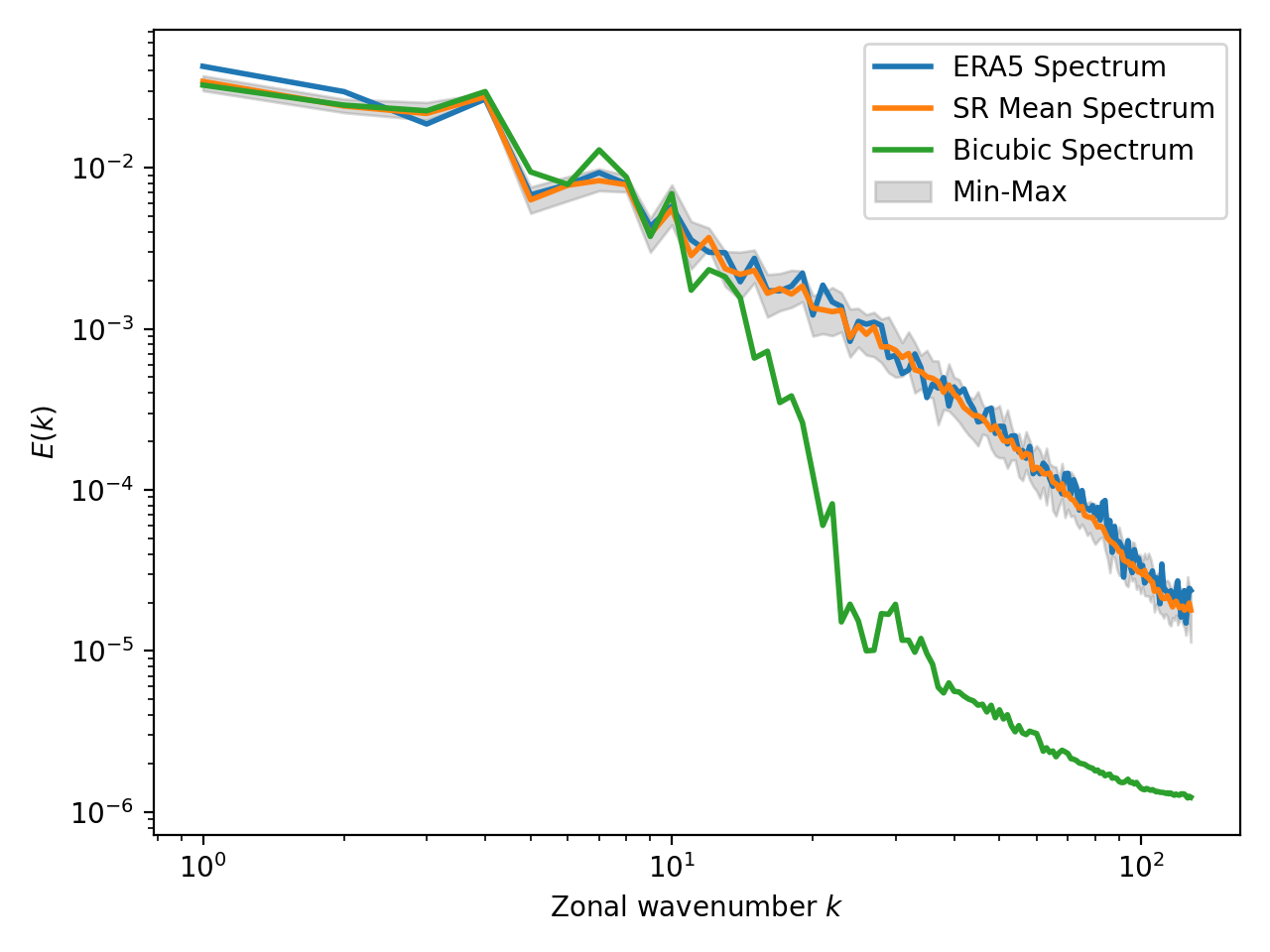}}
    }
    \vspace{-0.15cm}
    \caption{Spectral recovery exhibited by zero-shot sampling from ERA5 generative model when using sample-LR for zero-shot sampling. Note how the proposed approach recovers the right spectral trend as against that of bicubic interpolation}.
    \label{fig:lr_hr_spec}
    \vspace{-0.35cm}
\end{figure}

\begin{figure}[!ht]
    \centering
    \includegraphics[width=0.95\linewidth]{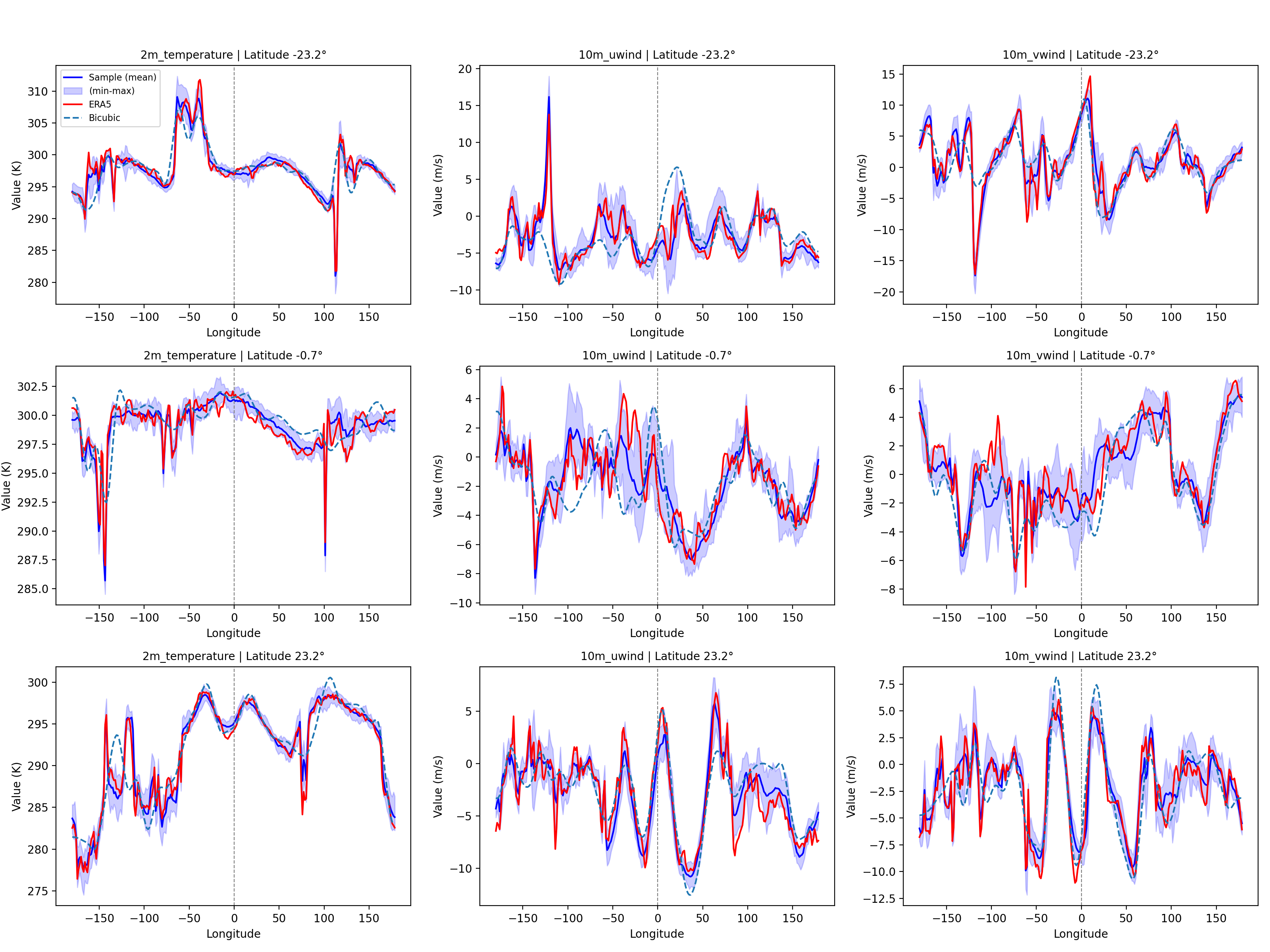}
    \caption{Traces for reconstructed flow-fields (including mean and standard deviation) compared with ground truth using sample-LR for zero-shot sampling.}
    \label{fig:lr_hr_lineplot}
    \vspace{-0.5cm}
\end{figure}
In the first experiment, we assess the ability to reconstruct the base-resolution ERA5 state given low-resolution observations from the same dataset (sample-LR). We qualitatively assess the reconstruction ability via test-time score-matching using visualizations of the fully recovered flow-fields for some selected variables in Figure \ref{fig:lr_hr_sr} where coarse-grained sparse observations as well as the ground truth flow-fields at a particular instant are available. Error contours are obtained from using the ensemble mean as the predicted full state recovery. The contours indicate the ability to recover fine-scaled features using the score-matching technique effectively. We also assess the mean and variance of the generated samples from the score-matching procedure in Figure \ref{fig:lr_hr_uq} where we can see correlations between variables and regions of high or low uncertainty. In particular, for recovering the 2m-temperature variable, we observe a higher uncertainty over the continents in contrast with wind speeds at 10m heights. For recovering specific humidity at 850mb, we observe no coherent regions of high uncertainty aside from the general region of the lower latitudes where high values of specific humidity are commonly observed.

Quantitative analyses are also performed by looking at the quality of spectral recovery from angle-averaged kinetic energy spectra for the various variables as shown in Figure \ref{fig:lr_hr_spec}. In these plots, it can be seen that the recovered spectra from the zero-shot score-matching procedure leads to good agreement with the ground truth spectra for different variables all the way through to the cut-off wavenumber. Additionally, we also assess the spatial variability of the generated fields by looking at traces along select latitudes across longitude in Figure \ref{fig:lr_hr_lineplot}. These latitudes are selected to reflect large variations that may be challenging for sample generation. Here, the ensemble mean and variance are also plotted showing certain regions with higher uncertainty. In general, a good recovery of the trends is observed, including the capture of sharp peaks and troughs accurately across different variables.

Next, we compare our super-resolution results against bicubic interpolation in Figures \ref{fig:bicubic1} and \ref{fig:bicubic2}. It is clearly observed that the diffusion samples are far superior, both in accuracy and spatial consistency. We observe the same for the spectrum (Figure \ref{fig:lr_hr_spec}) and traces along latitudes (Figure \ref{fig:lr_hr_lineplot}). We remark that the purpose of this experiment is not to propose a novel super-resolution algorithm for the atmosphere, of which there are several examples \cite{mardani2025residual,leinonen2020stochastic,sundar2024taudiff}. Instead, we assess the promise of super-resolution when interpreted as a posterior sampling process when likelihoods are generated from coarse data and the prior is generated from fine-scaled training data. Indeed, we note that most deep learning super-resolution frameworks build a direct map between coarse and fine-grid data which may provide more accurate reconstructions of the fine scales at the cost of flexibility of the coarse-grid representation. We remind the reader that the posterior sampling framework of diffusion allows one to perform reconstructions with various modalities and resolutions of data without retraining the diffusion model. These are demonstrated in the rest of the article.

\subsection{Super-resolution from unstructured data and different modality}\label{sec:IGRA}
\vspace{-0.1cm}
\begin{figure}[!ht]
    \centering
    \mbox{
    \subfigure[Temperature (K)]{\includegraphics[width=0.30\linewidth]{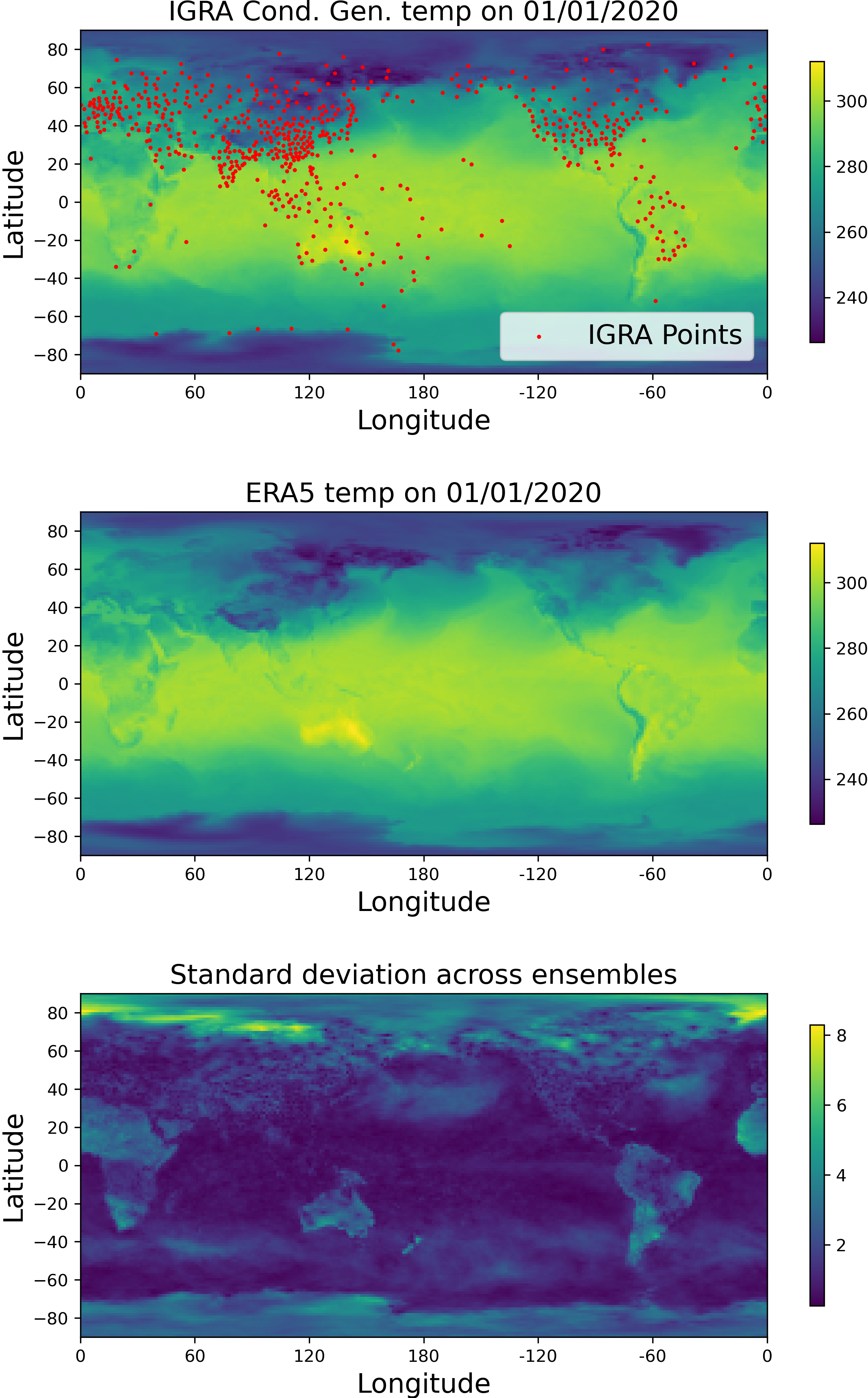}}
    \subfigure[U-wind (m/s)]{\includegraphics[width=0.3\linewidth]{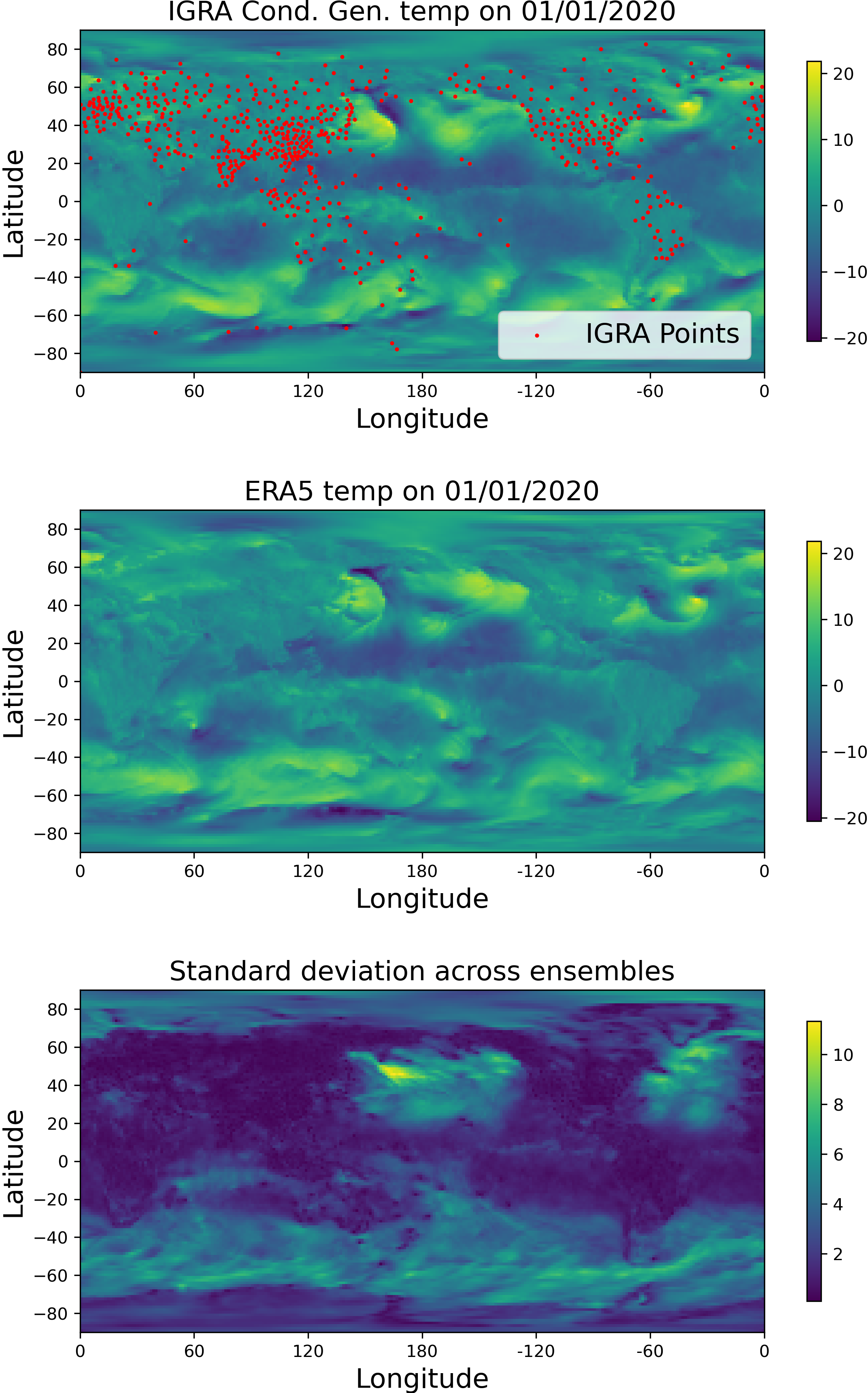}}
    \subfigure[V-wind (m/s)]{\includegraphics[width=0.3\linewidth]{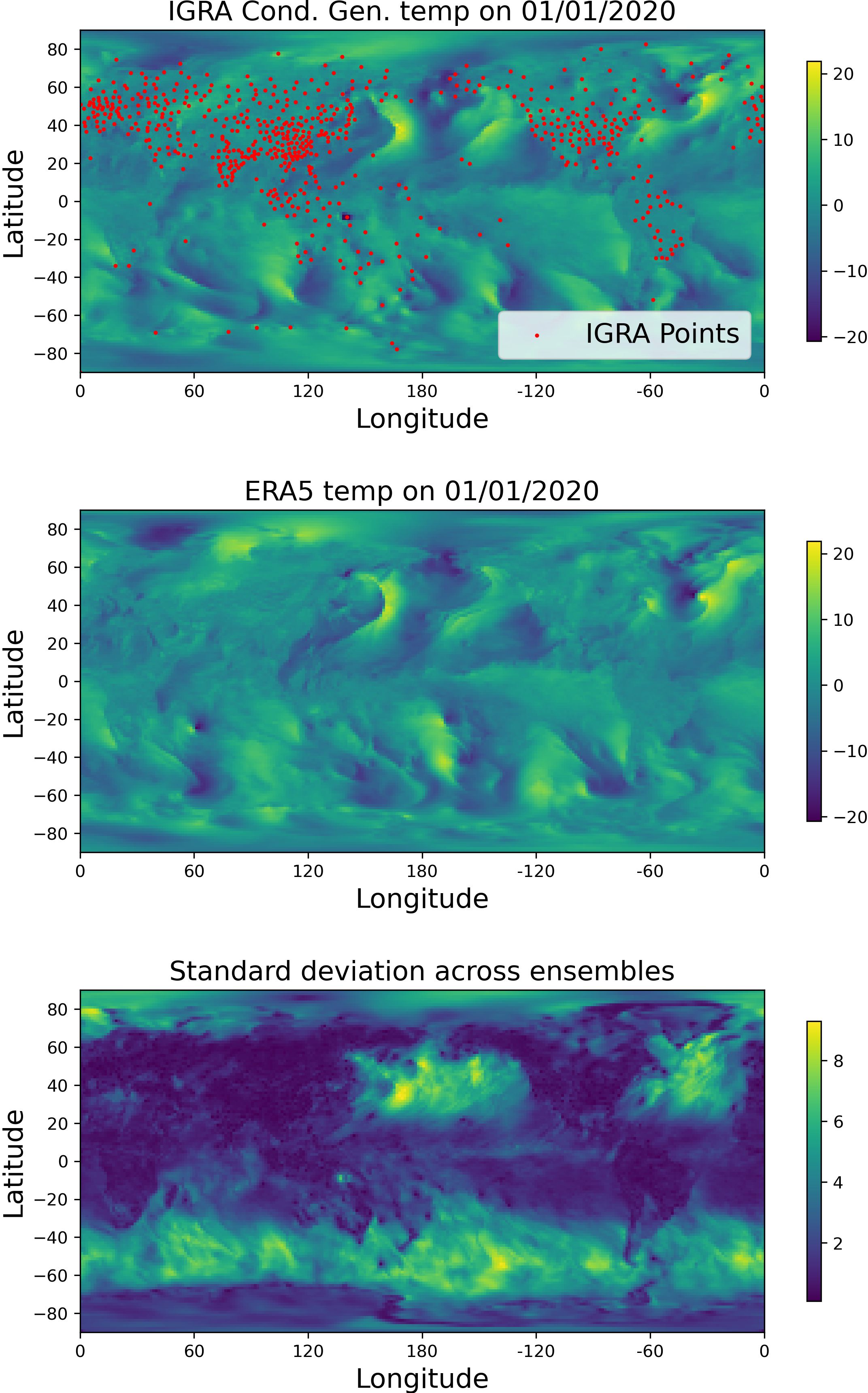}}
    }
    \vspace{-0.25cm}
    \caption{Super-resolution using zero-shot sampling with posterior updates based on IGRA observations (sample-IGRA). The rows indicate ensemble mean (top), ground truth (middle), and ensemble standard deviation from multiple zero-shot samples. Continents show lower uncertainty since IGRA radiosondes are predominantly based on land.}
    \label{fig:igra_hr_temp}
    \vspace{-0.25cm}
\end{figure}
Our next experiment is to extend the setting of ERA5 state recovery given sparse and unstructured observations from a source of data that is \emph{different} to that used for training. This can be contrasted to the previous section, where it was assumed that sparse observations were obtained from the original training data set distribution. Specifically, we construct a problem statement where our unconditional ERA5 generation is informed, during test, using sparse observations of the IGRA radiosonde dataset (i.e., the sample-IGRA configuration). Notably, this dataset is available only at land locations and therefore represents a sensing platform that is constrained. Figure \ref{fig:igra_hr_temp} shows sparse locations where IGRA readings are sampled and their corresponding full-state recovery performance for exemplar time instances. Clear patterns may be distinguished for uncertainty estimates. In particular, one can observed much lower uncertainty over the land which corresponds to regions where sensor measurements are available in real-time. Higher uncertainty, particularly for wind-speeds are observed over the oceans. 

\begin{figure}[!h]
\vspace{-0.5cm}
    \centering
    \mbox{
    \subfigure[Temperature]{\includegraphics[width=0.33\textwidth]{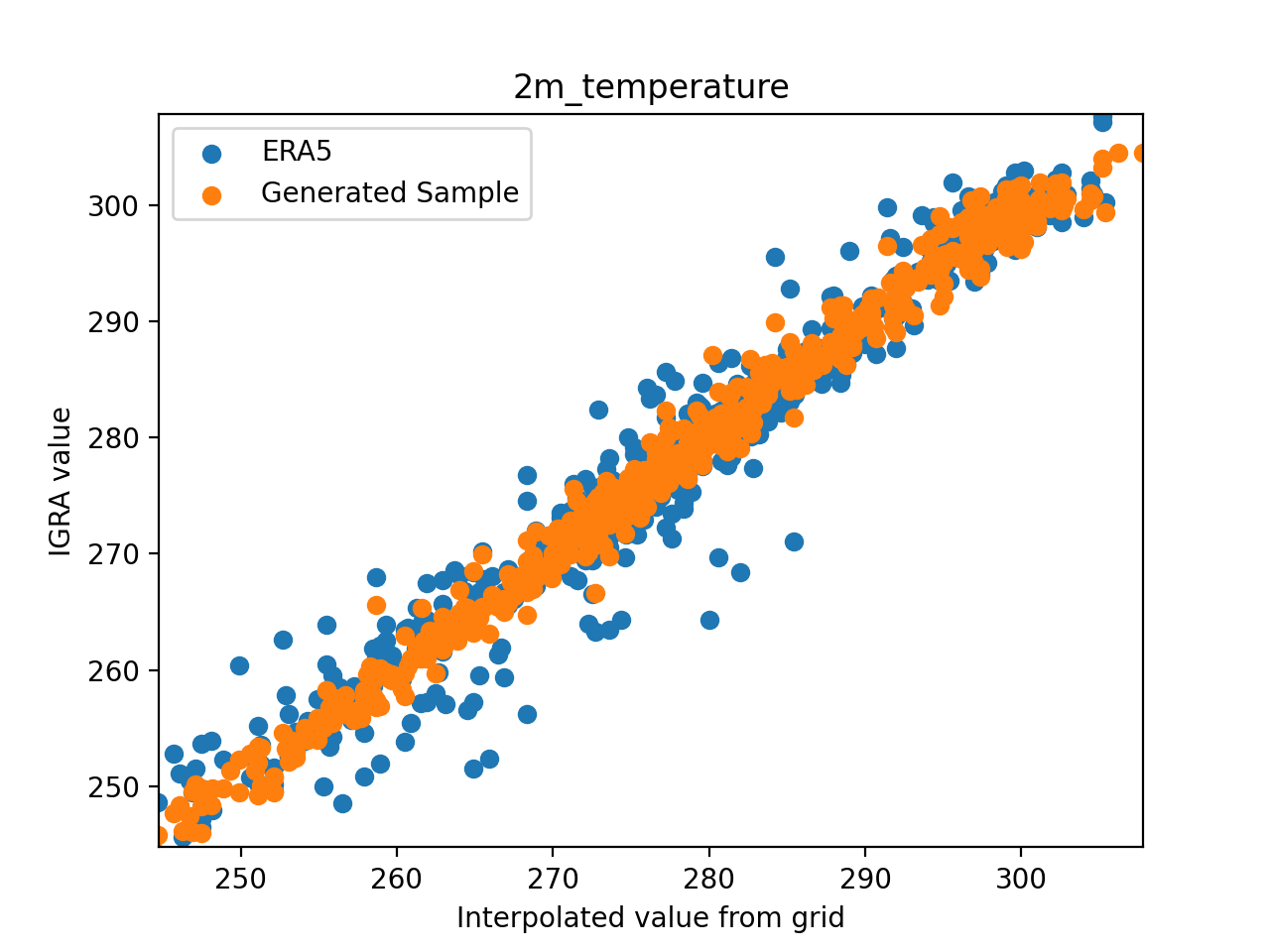}}
    \subfigure[U-Wind]{\includegraphics[width=0.33\textwidth]{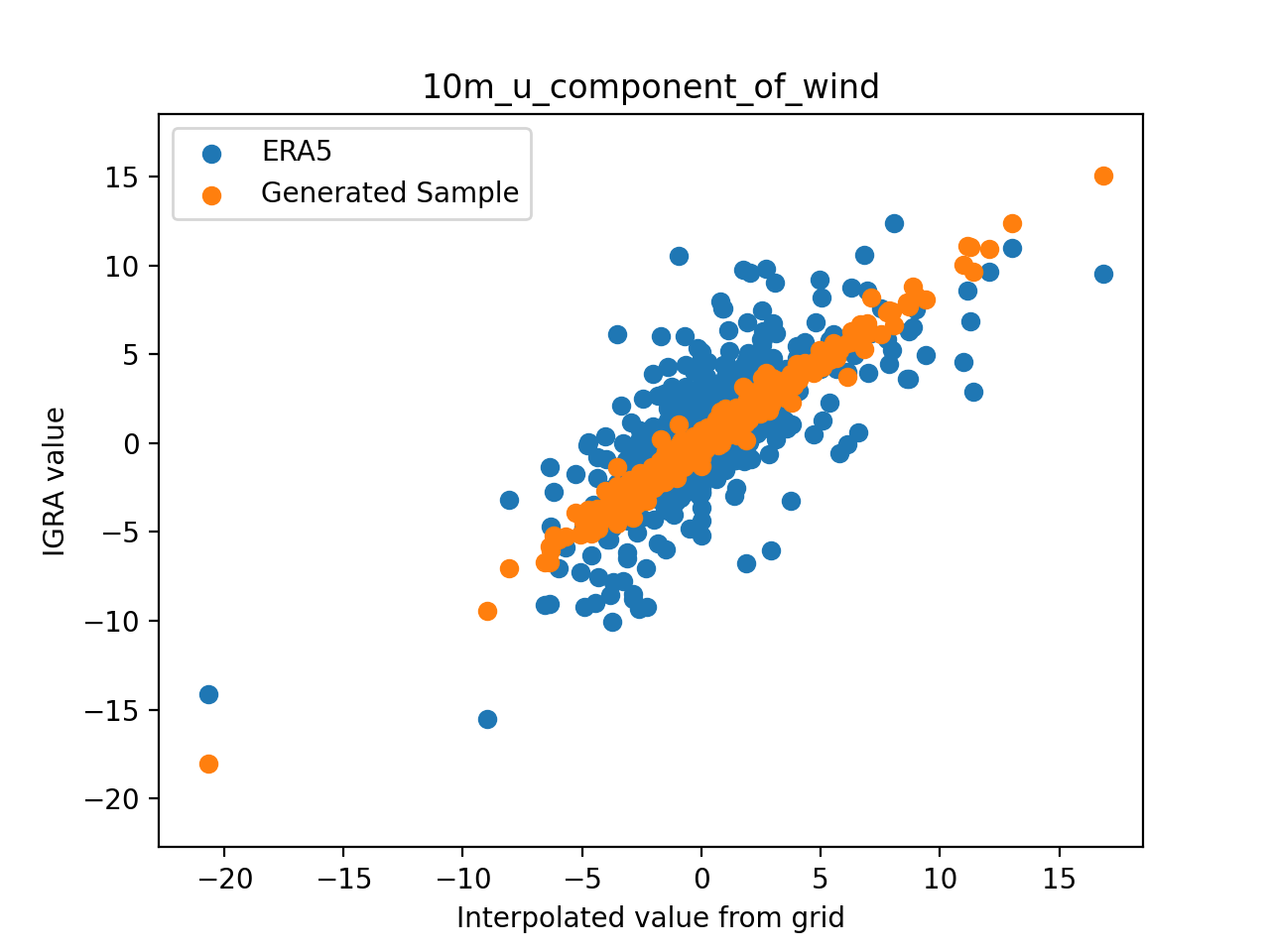}}
    \subfigure[V-Wind]{\includegraphics[width=0.33\textwidth]{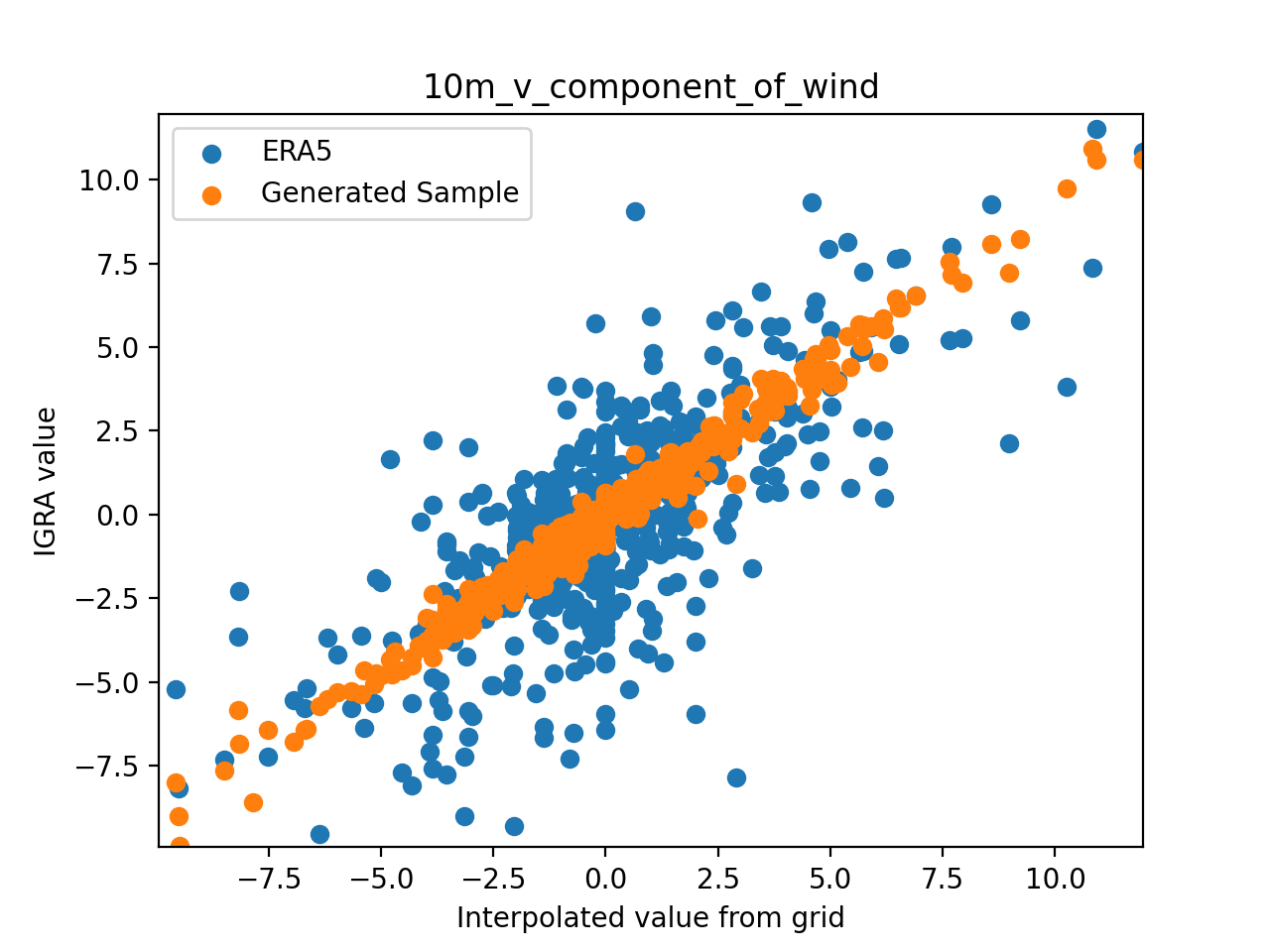}}
    } \\
    \mbox{
    \subfigure[Temperature]{\includegraphics[width=0.33\textwidth]{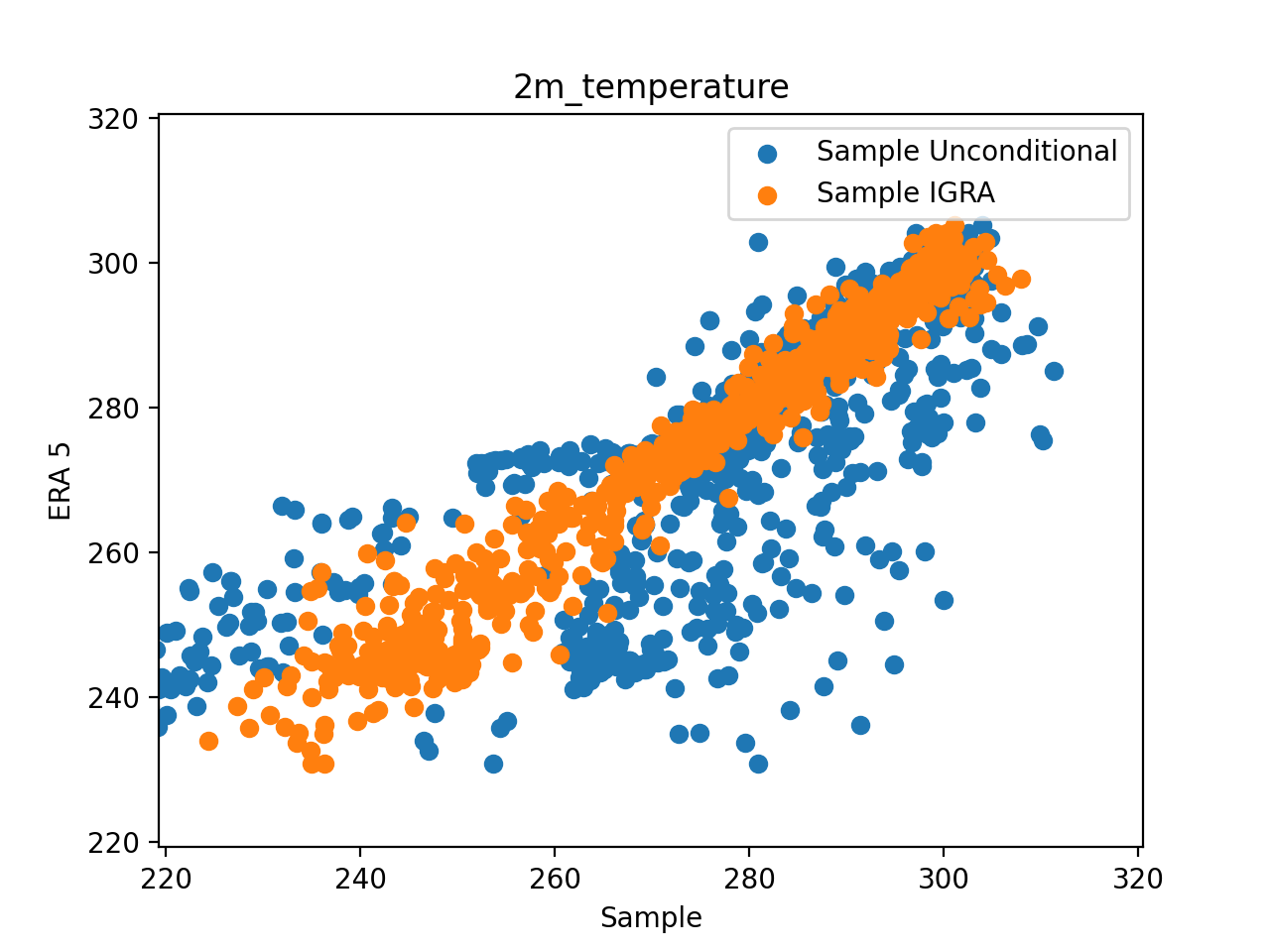}}
    \subfigure[U-Wind]{\includegraphics[width=0.33\textwidth]{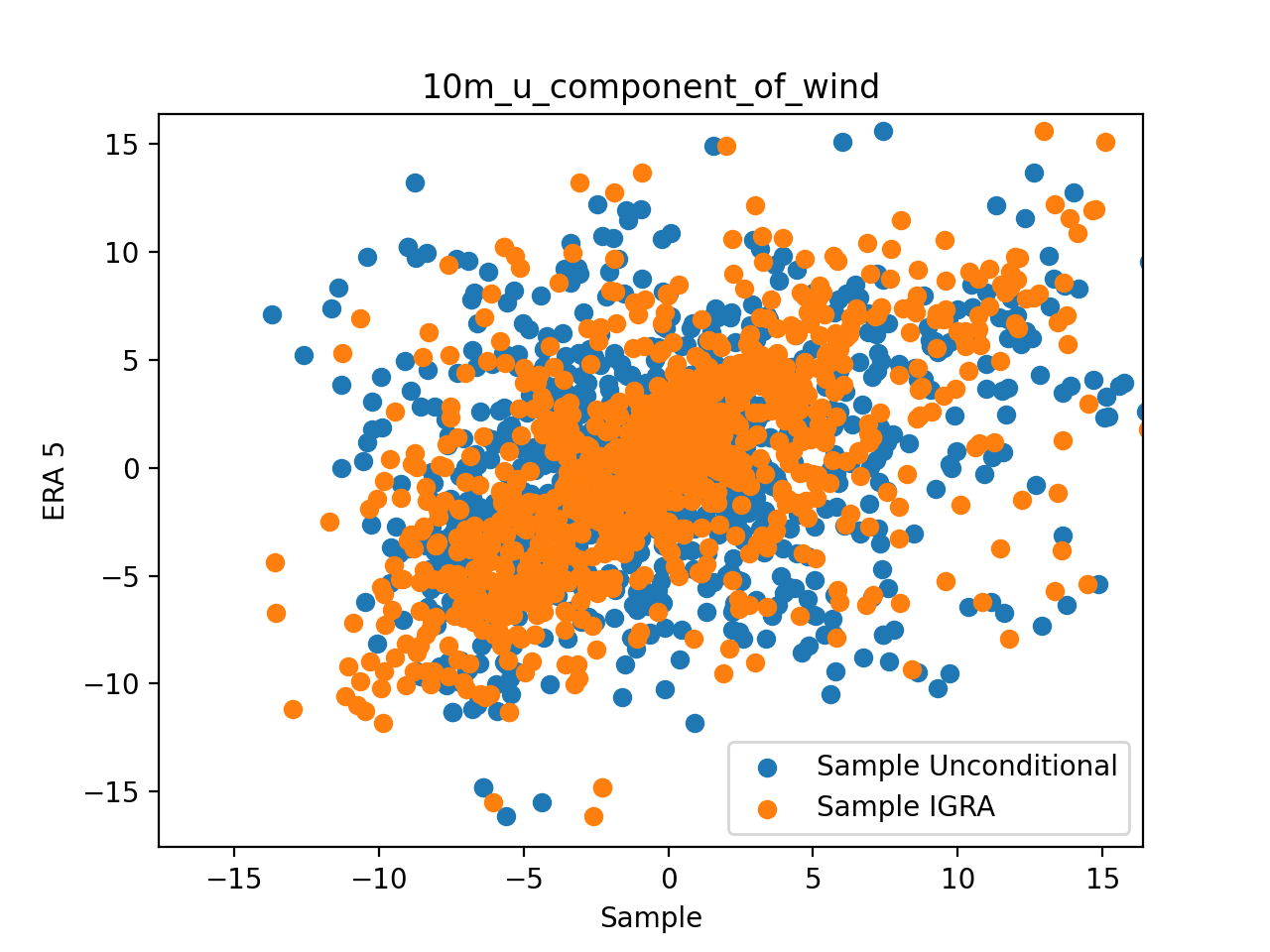}}
    \subfigure[V-Wind]{\includegraphics[width=0.33\textwidth]{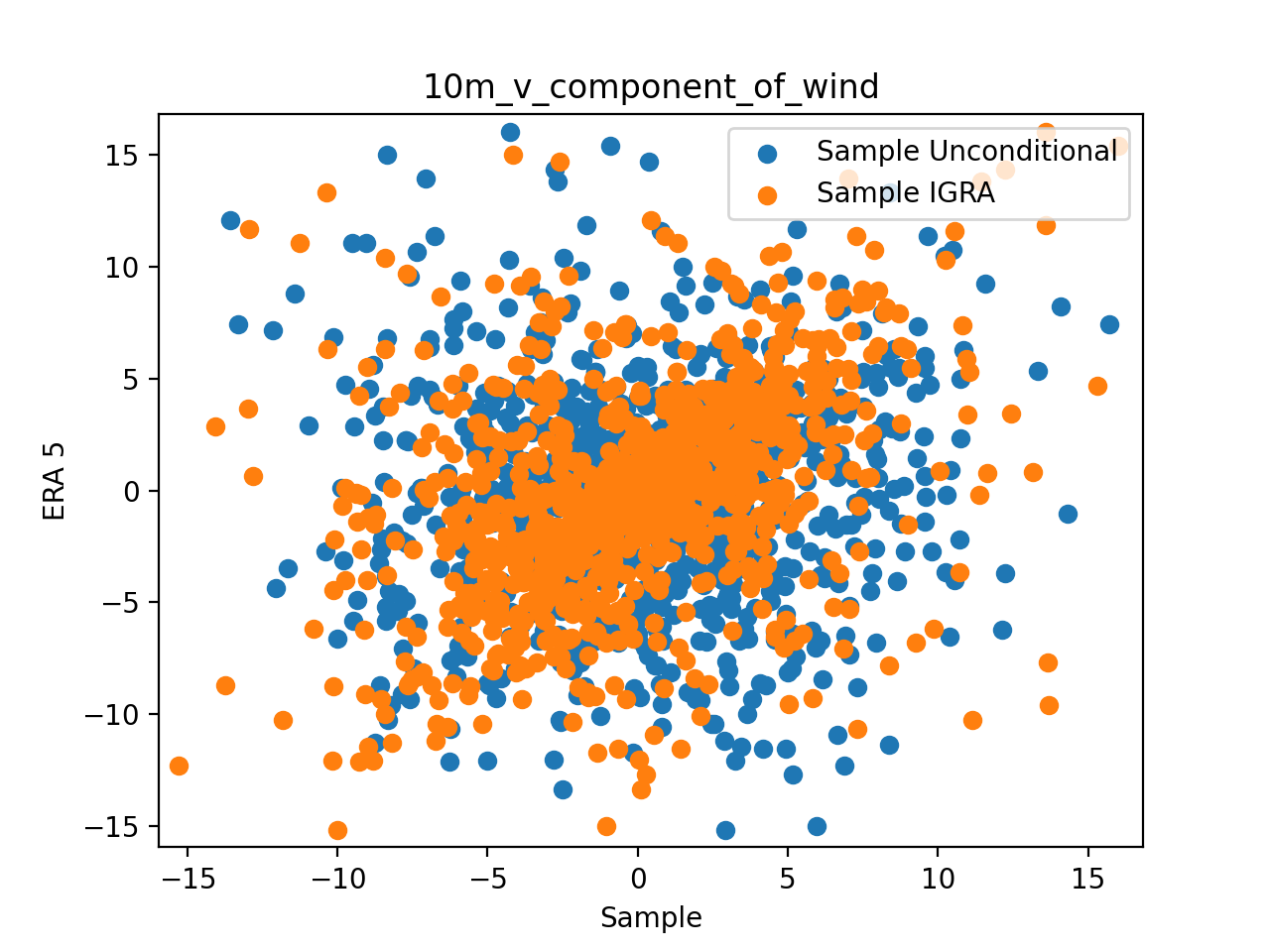}}
    }
    \caption{Scatters at IGRA sensor locations (top) and non-IGRA locations (bottom) from zero-shot sampling using sample-IGRA. The knowledge of IGRA sensor measurements improves the reconstruction at sensor locations significantly as seen above. For locations where observations are not available, some variables, such as temperature are reconstructed more accurately than others such as wind speeds.}
    \label{fig:igra_hr_temp_scatter}
\end{figure}

\begin{figure}[!h]
    \centering
    \mbox{
    \subfigure[Geopotential 500]{\includegraphics[width=0.33\textwidth]{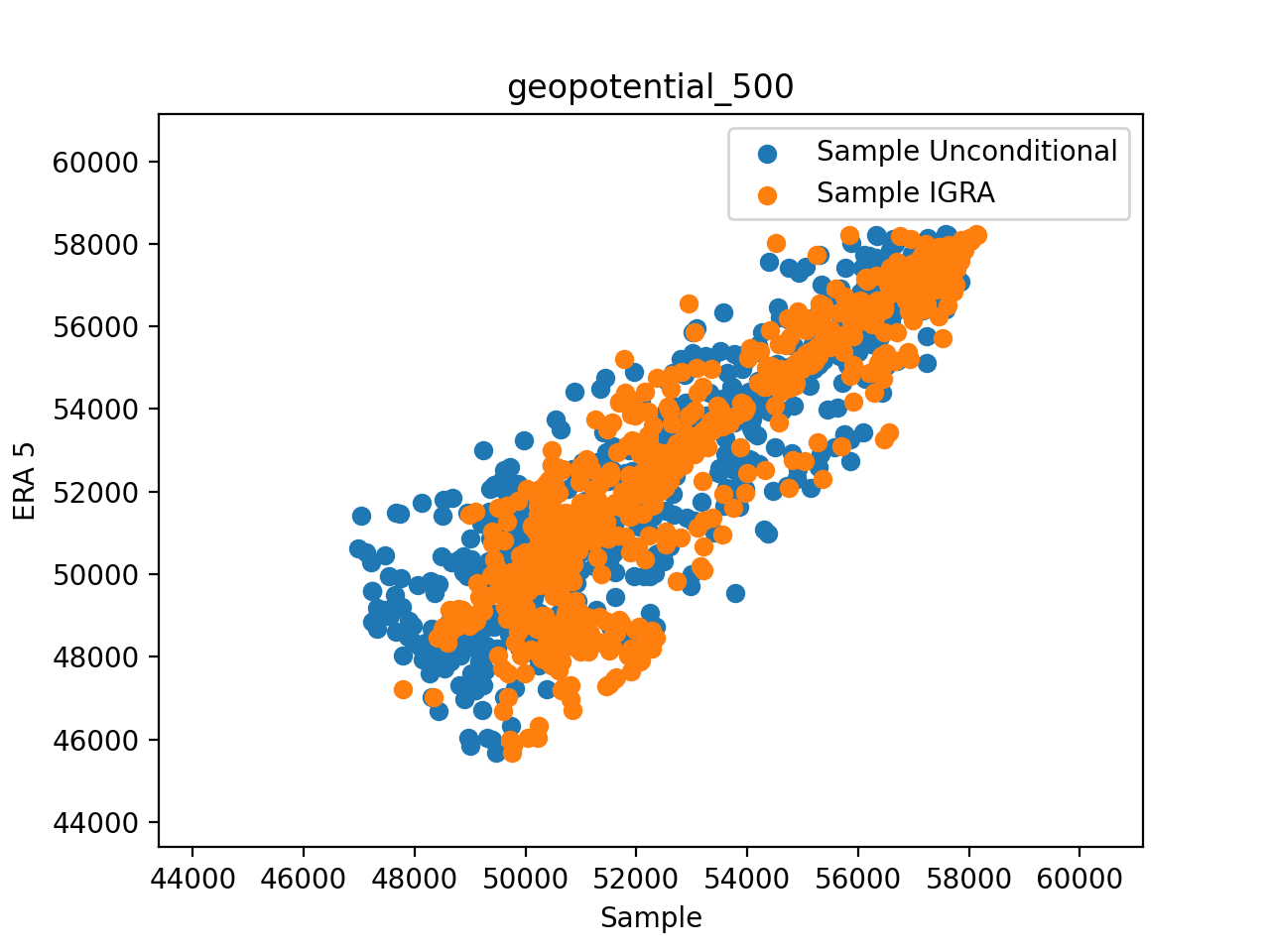}}
    \subfigure[Temperature 850]{\includegraphics[width=0.33\textwidth]{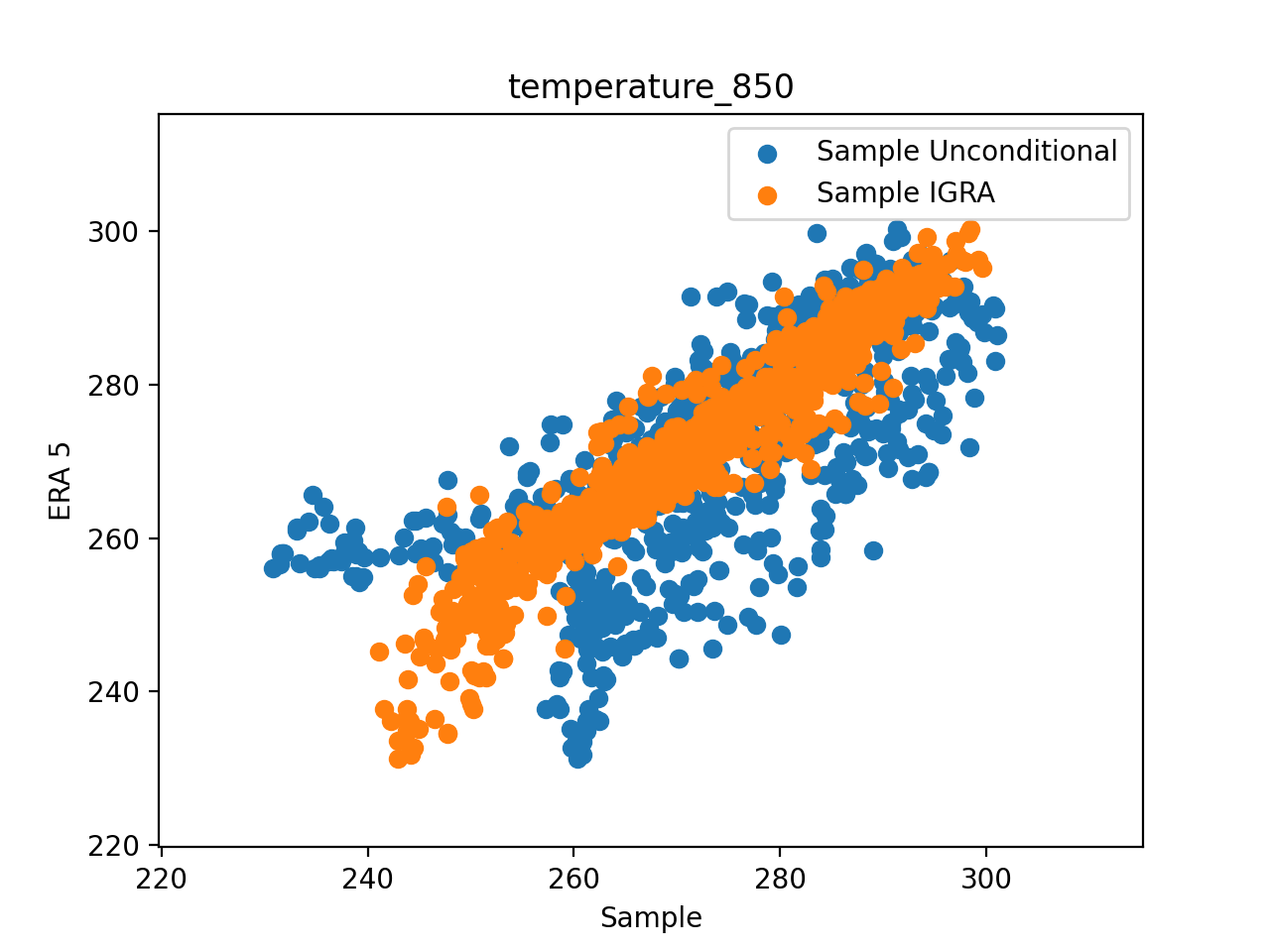}}
    \subfigure[Humidity 850]{\includegraphics[width=0.33\textwidth]{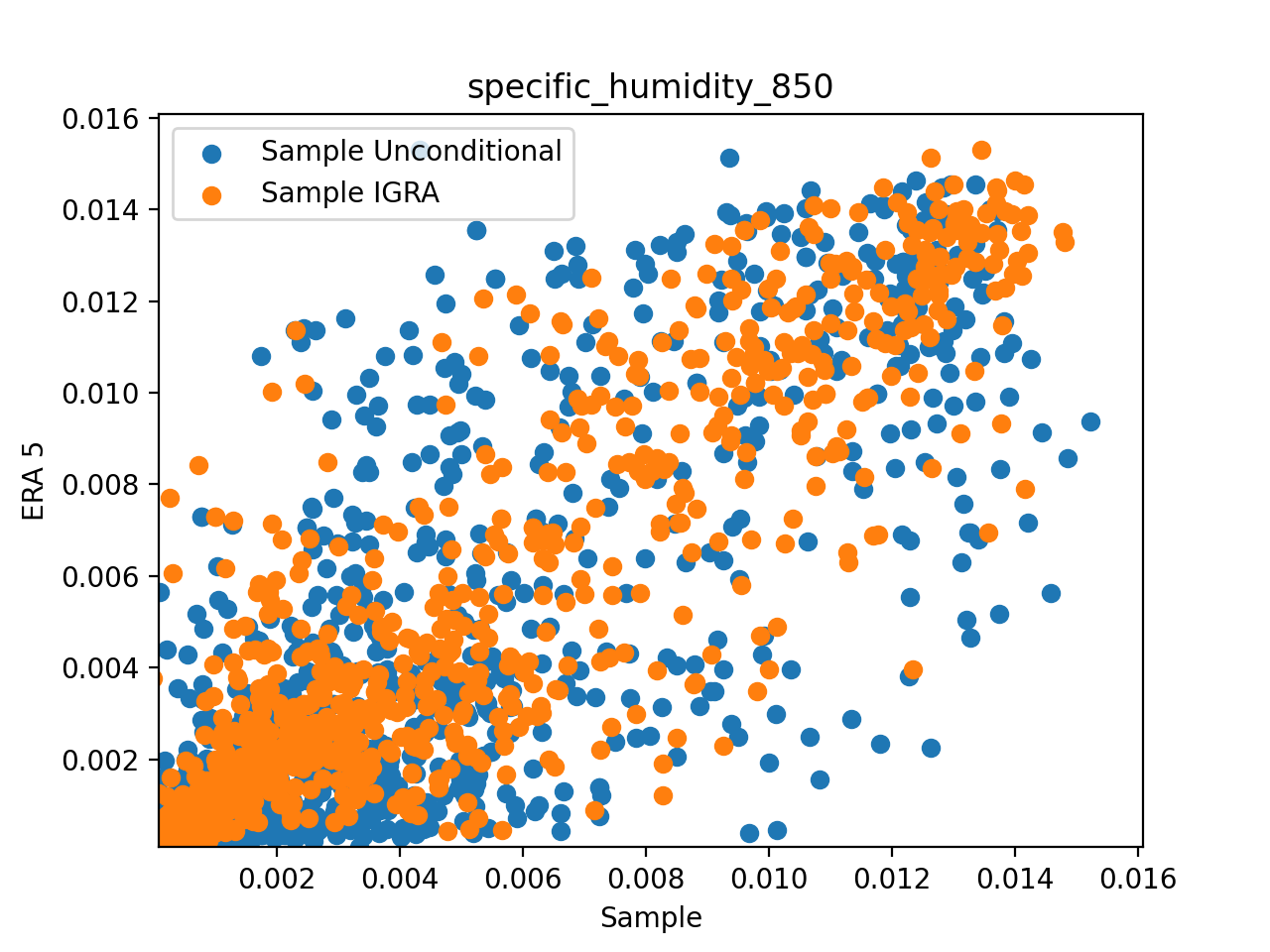}}
    }
    \caption{Scatters for additional fields at non-IGRA locations using zero-shot sampling from sample-IGRA. The knowledge of IGRA sensor measurements improves the reconstruction mostly at sensor locations as seen above.}
    \label{fig:igra_hr_temp_scatter_2}
\end{figure}
\begin{figure}[!ht]
\vspace{-0.4cm}
    \centering
    \mbox{
    \subfigure[2m Temperature]{\includegraphics[width=0.45\textwidth]{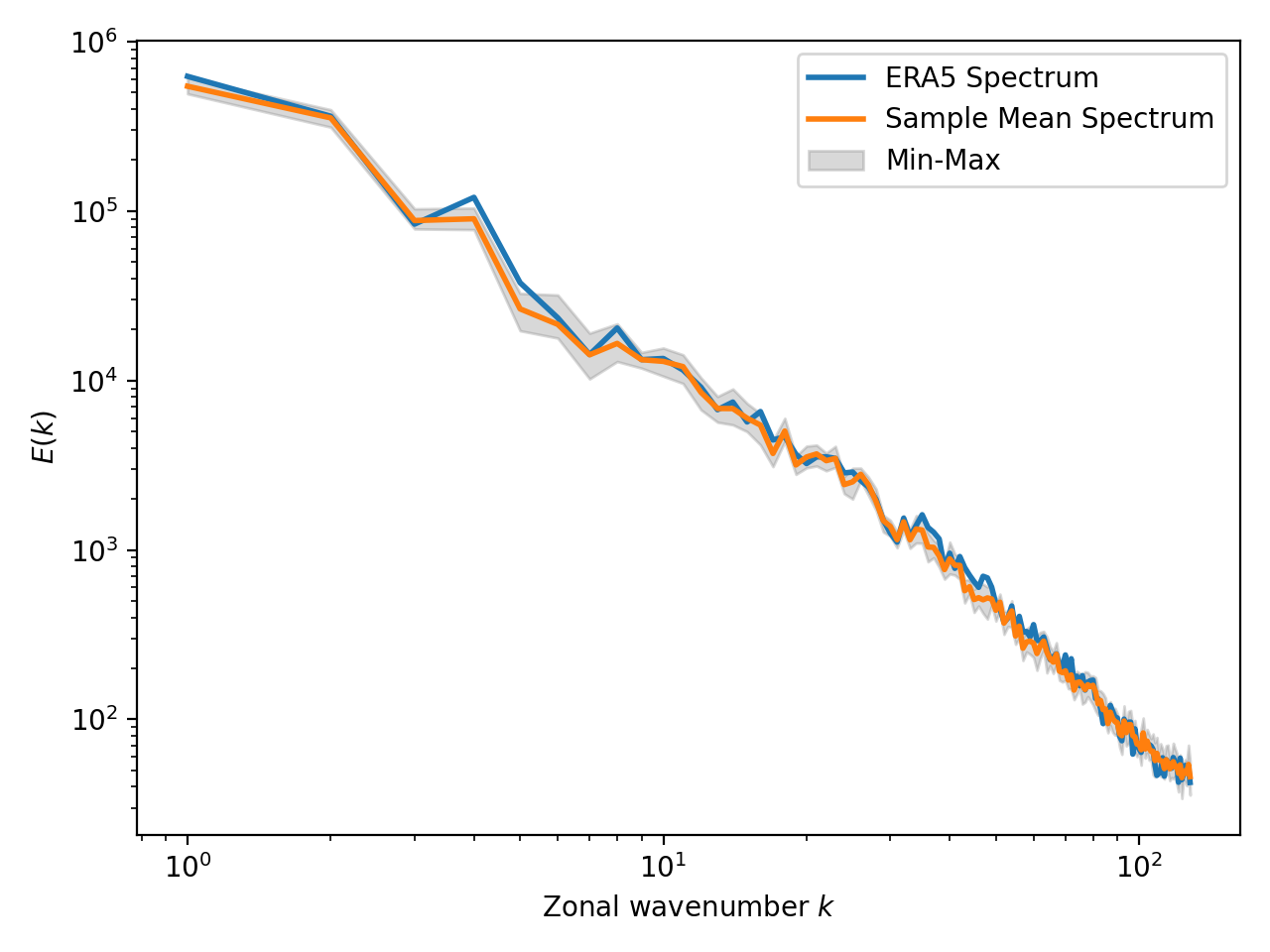}}
    \subfigure[10m u-wind]{\includegraphics[width=0.45\textwidth]{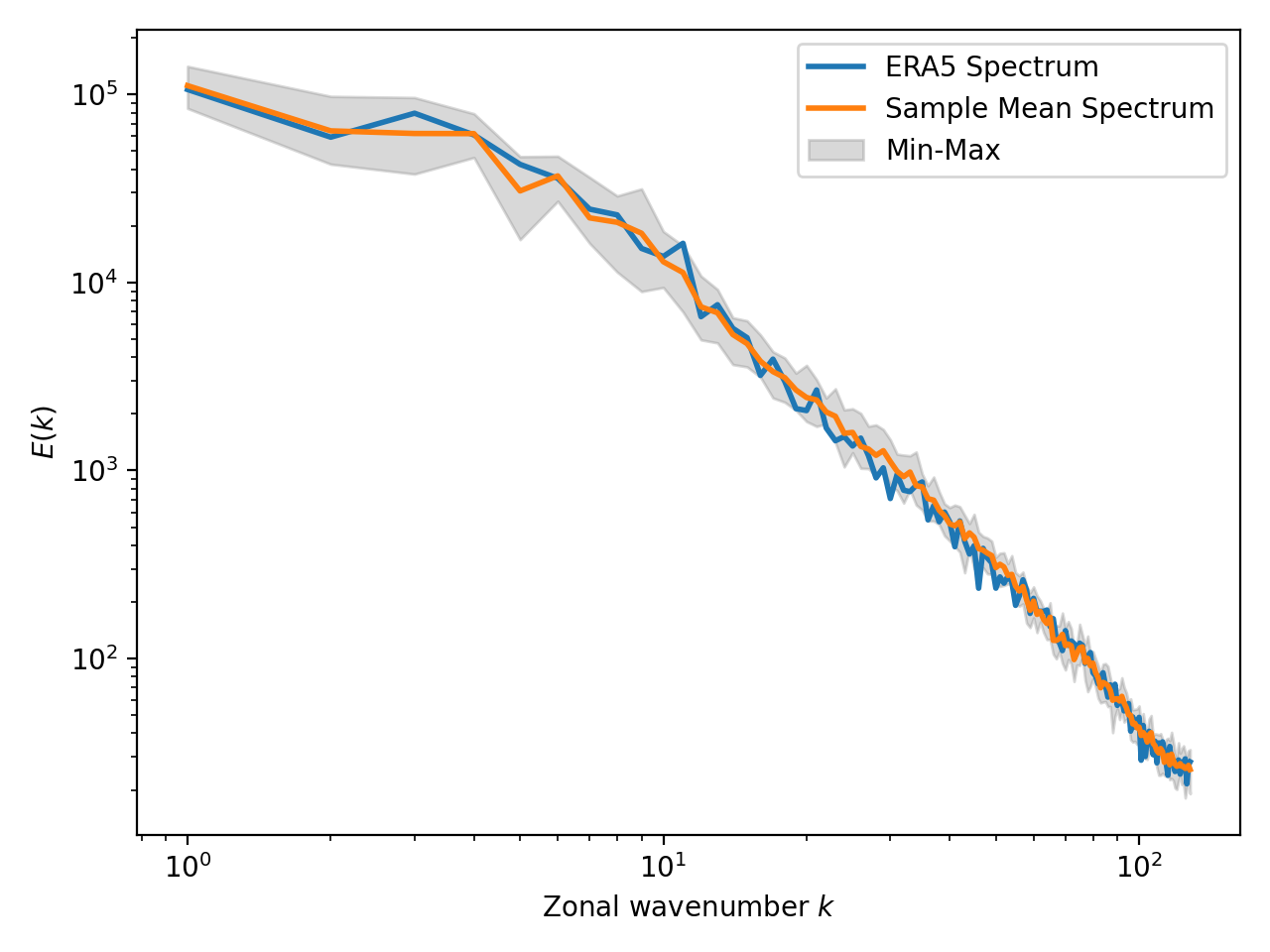}}
    }\\
    \mbox{
    \subfigure[10m v-wind]{\includegraphics[width=0.45\textwidth]{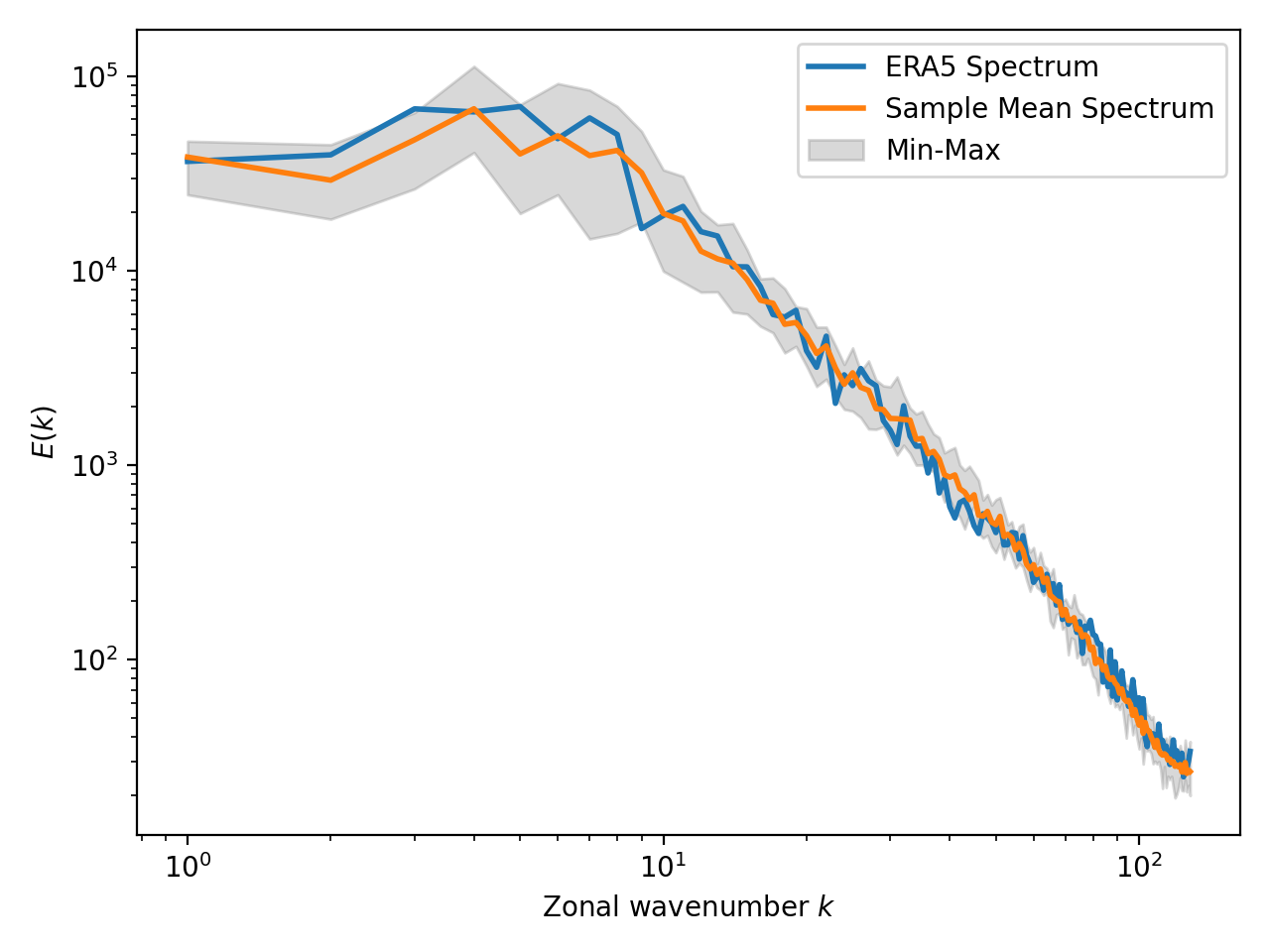}}
    \subfigure[850mb specific humidity]{\includegraphics[width=0.45\textwidth]{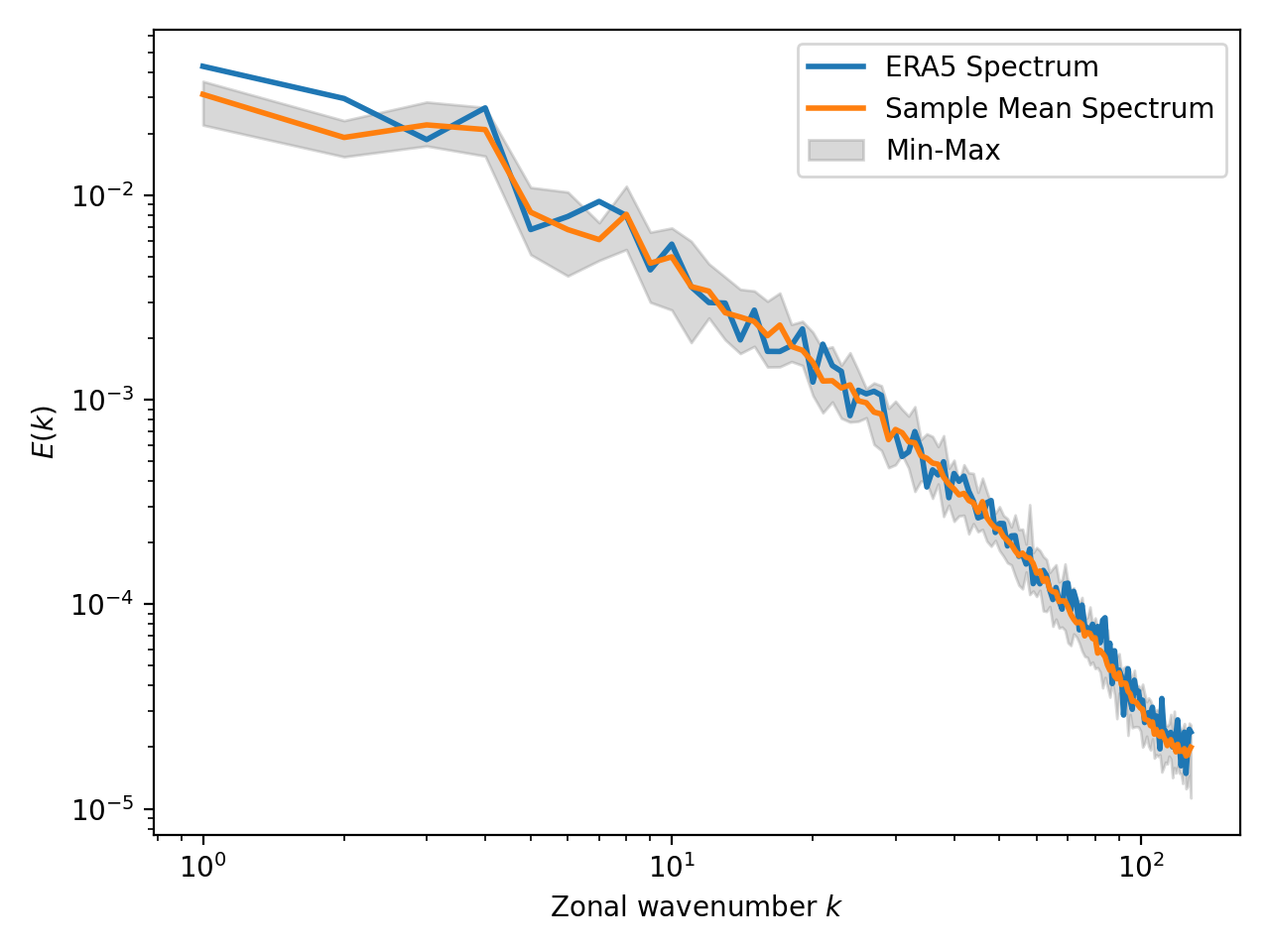}}
    }
    \vspace{-0.1cm}
    \caption{Spectral recovery exhibited by zero-shot sampling from ERA5 generative model when observing sparse observation data from the IGRA dataset (sample-IGRA). Slight cut-off wavenumber errors are observed for the reconstructed flow-fields.}
    \label{fig:igra_hr_spec}
\end{figure}
\begin{figure}[!ht]
\vspace{-0.4cm}
    \centering
    \includegraphics[width=0.95\linewidth]{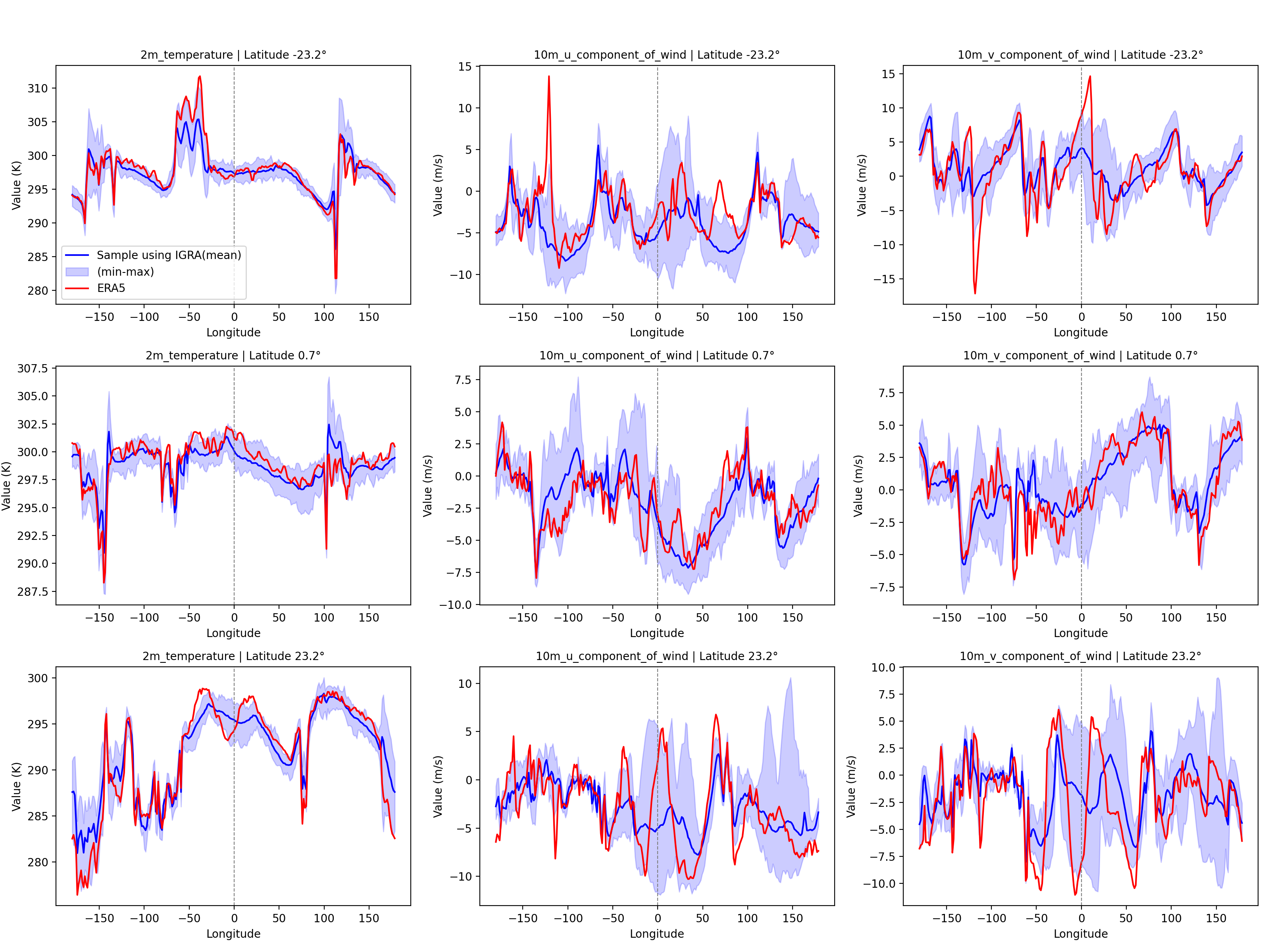}
    \vspace{-0.1cm}
    \caption{Traces for reconstructed flow-fields (including mean and standard deviation) compared with ground truth for the sample-IGRA zero-shot sampling.}
    \label{fig:igra_hr_lineplot}
\vspace{-0.5cm}
\end{figure}
In Figure \ref{fig:igra_hr_temp_scatter}, we plot scatters for three representative variables at IGRA locations, as generated by the score-matching procedure. We expect to see the values generate collapse on a 45$^\circ$ line indicating that IGRA observations have been assimilated well. We also show the ground truth ERA5 values at the specific day in these scatters, where significant deviations are observed for the wind speeds. This indicates that the state recovery procedure has prioritized the likelihood obtained by IGRA observations at this point. In the same figure, we assess the performance of the generative model (after score-matching) for recovering the ground truth ERA5 at non-sampled locations given IGRA observations. Here it is seen that the spread for the generated values at non-IGRA points is generally reduced (in particular for temperature and specific humidity), indicating the IGRA observations have helped constrain the generator to the specific state. The spread from the unconditional samples indicates that while the unconditional generative process is able to provide fields that are ERA5-like, they do not recover the specific flow-fields without score-matching. Furthermore, we also assess the reconstruction quality of the generative super-resolution for additional fields in Figure \ref{fig:igra_hr_temp_scatter_2} at points that are not collocated with IGRA sensing locations. We observe that for some variables, such as geopotential height at 500mb and temperature at 850mb, IGRA observations lead to much improved reconstructions of ERA5. This represents a potential approach for rapid assessment of the quality of observations for recovering specific variables in the atmosphere.

As in the previous section, we also assess the spectral quality of the ERA5 state recovery, given IGRA observations, through angle-averaged kinetic energy spectra as shown in Figure \ref{fig:igra_hr_spec}. In general, similar recovery quality is observed with a slight increase in grid cut-off wavenumber errors for the ensemble means. We also draw the reader's attention to the higher spread in the variance for wind-speed reconstruction indicating how the sensing of IGRA observations may not necessarily correlate with high quality state or spectral recovery for all variables. We believe this to be a promising metric to optimize for improved sensing platforms. Finally, we also plot traces of the generated state recovery for different latitudes across longitudes in Figure \ref{fig:igra_hr_lineplot}, where comparisons with the ground truth ERA5 are also made. One can observe an increased variance from the samples of the generative model indicating greater uncertainty in the state reconstruction. This reflects the intrinsic challenges of reconstructing the reanalysis from partial and spatially anisotropic observations such as IGRA.

\subsection{Multimodal super-resolution}\label{sec:multimodal_sr}
\begin{figure}[!ht]
    \centering
    \mbox{
    \subfigure[2m Temperature]{\includegraphics[width=0.45\textwidth]{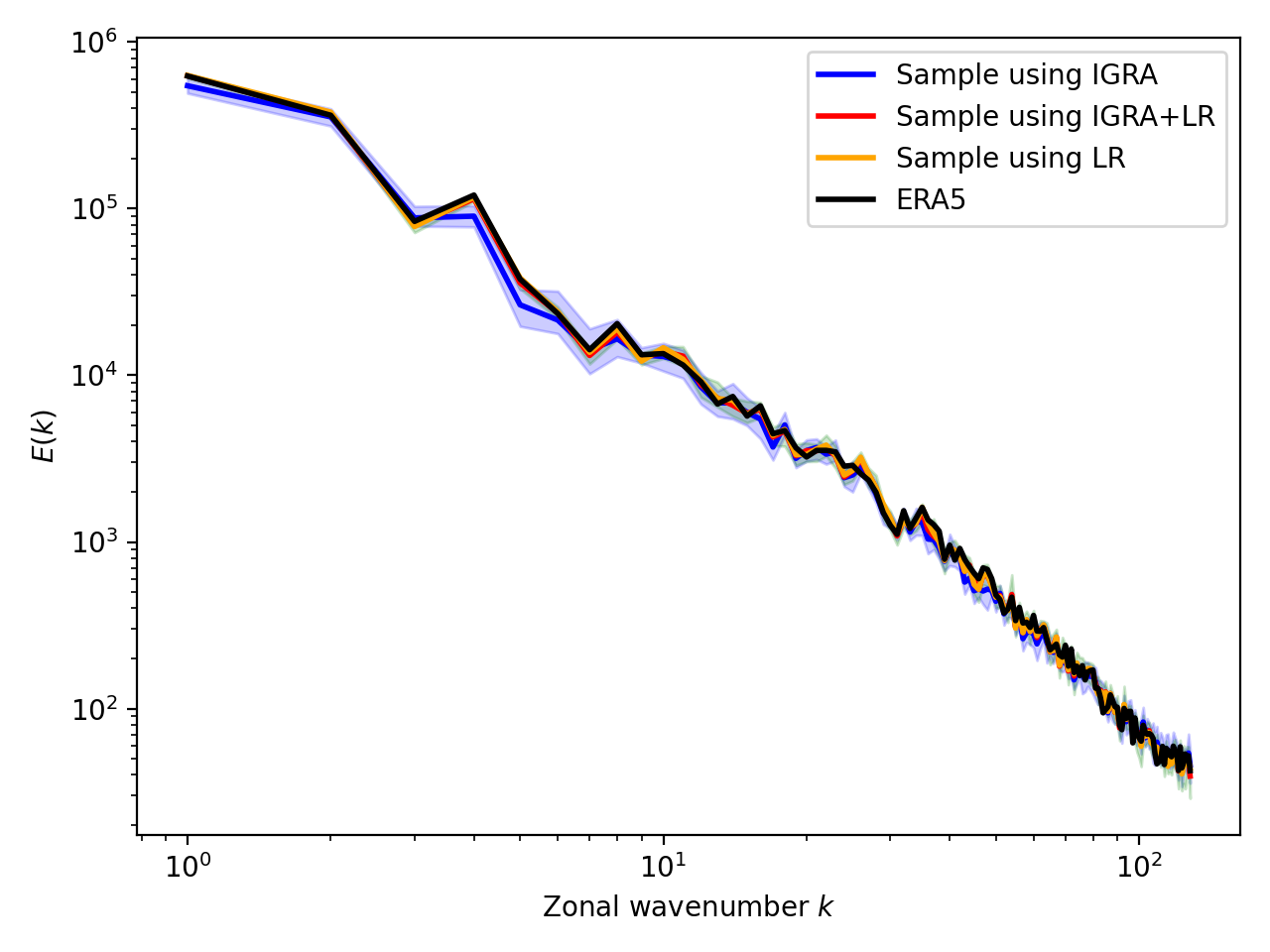}}
    \subfigure[10m u-wind]{\includegraphics[width=0.45\textwidth]{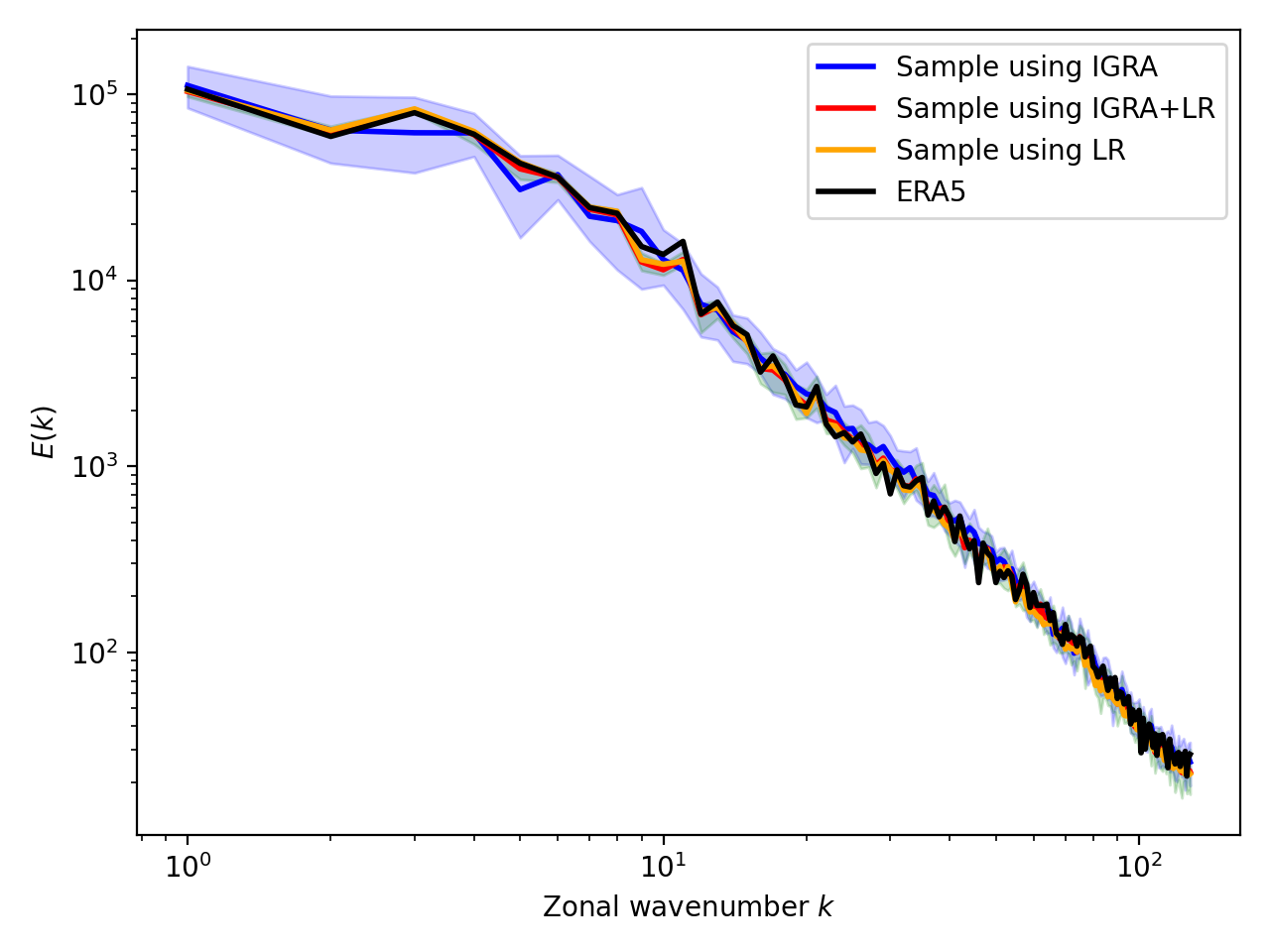}}
    }\\
    \mbox{
    \subfigure[10m v-wind]{\includegraphics[width=0.45\textwidth]{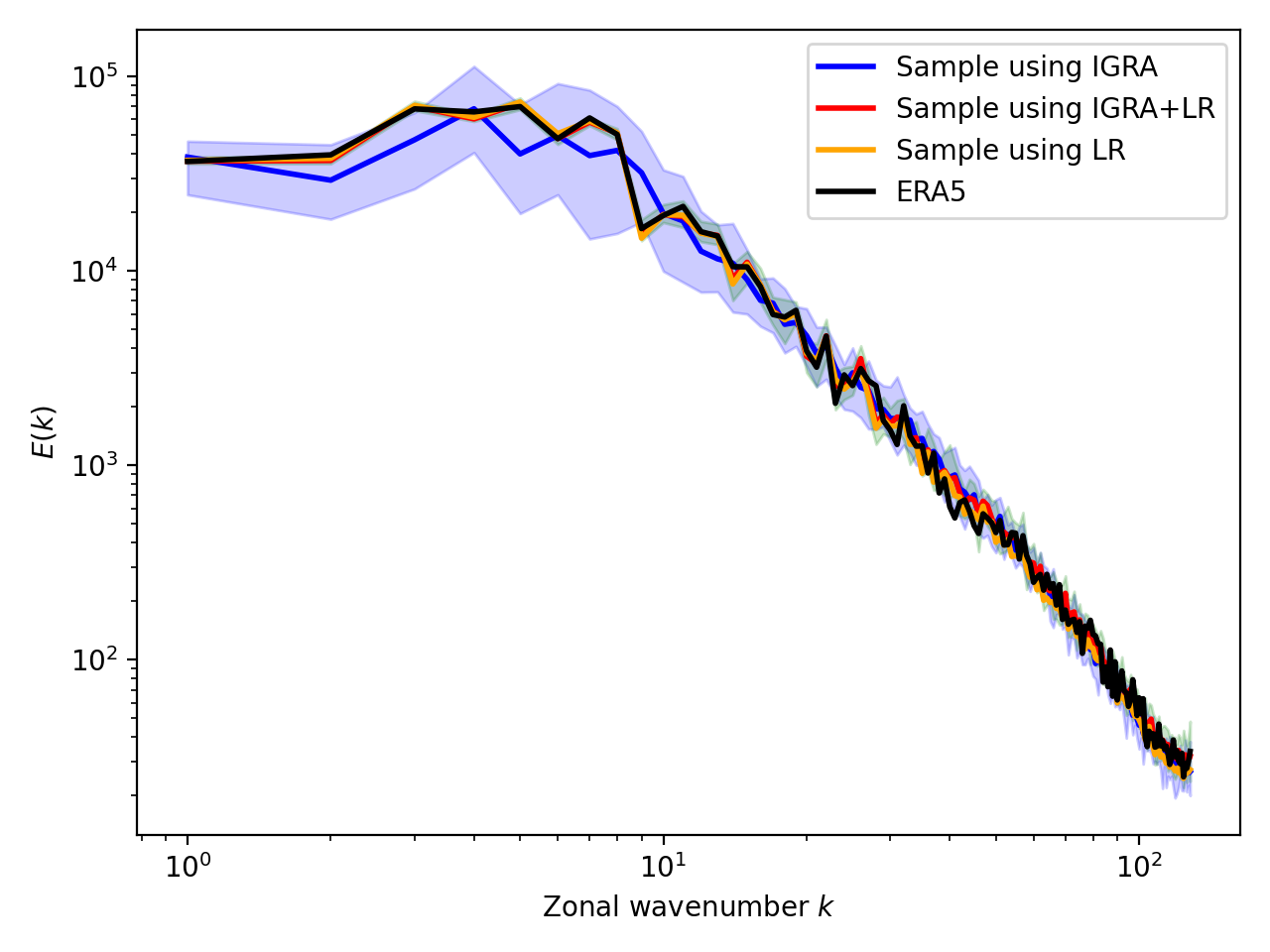}}
    \subfigure[850mb specific humidity]{\includegraphics[width=0.45\textwidth]{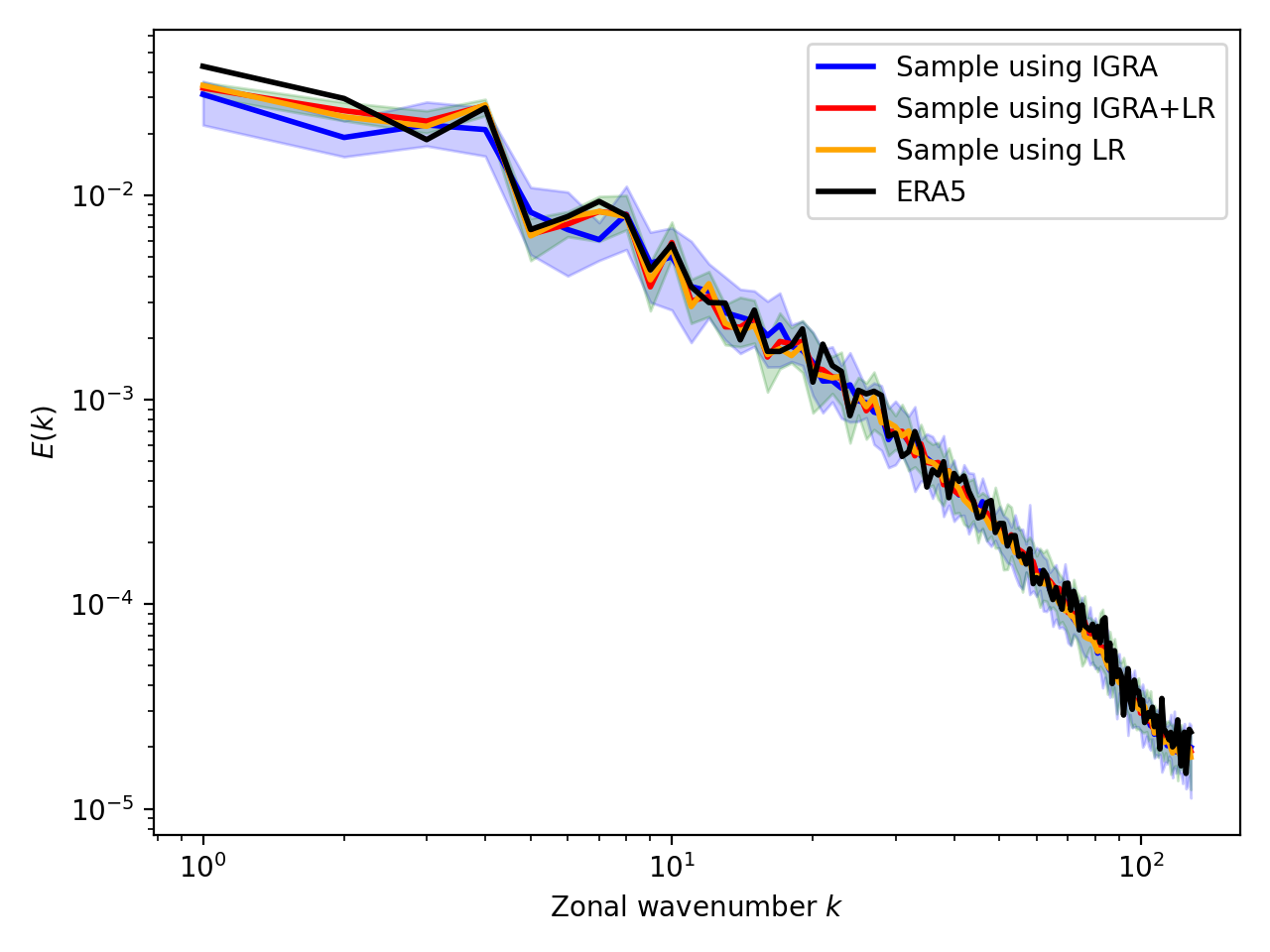}}
    }
    \caption{Spectral recovery exhibited by zero-shot sampling from ERA5 generative model when observing sparse observation data from the IGRA dataset as well as from low-resolution reanalysis (sample IGRA+LR). The biases in the higher wavenumbers obtained from state reconstructions with only IGRA observations are reduced by using low-resolution observations as well.}
    \label{fig:igra_lr_hr_spec}
\end{figure}
\begin{figure}[!ht]
    \centering
    \mbox{
    \subfigure[Surface Temperature]{\includegraphics[width=0.33\textwidth]{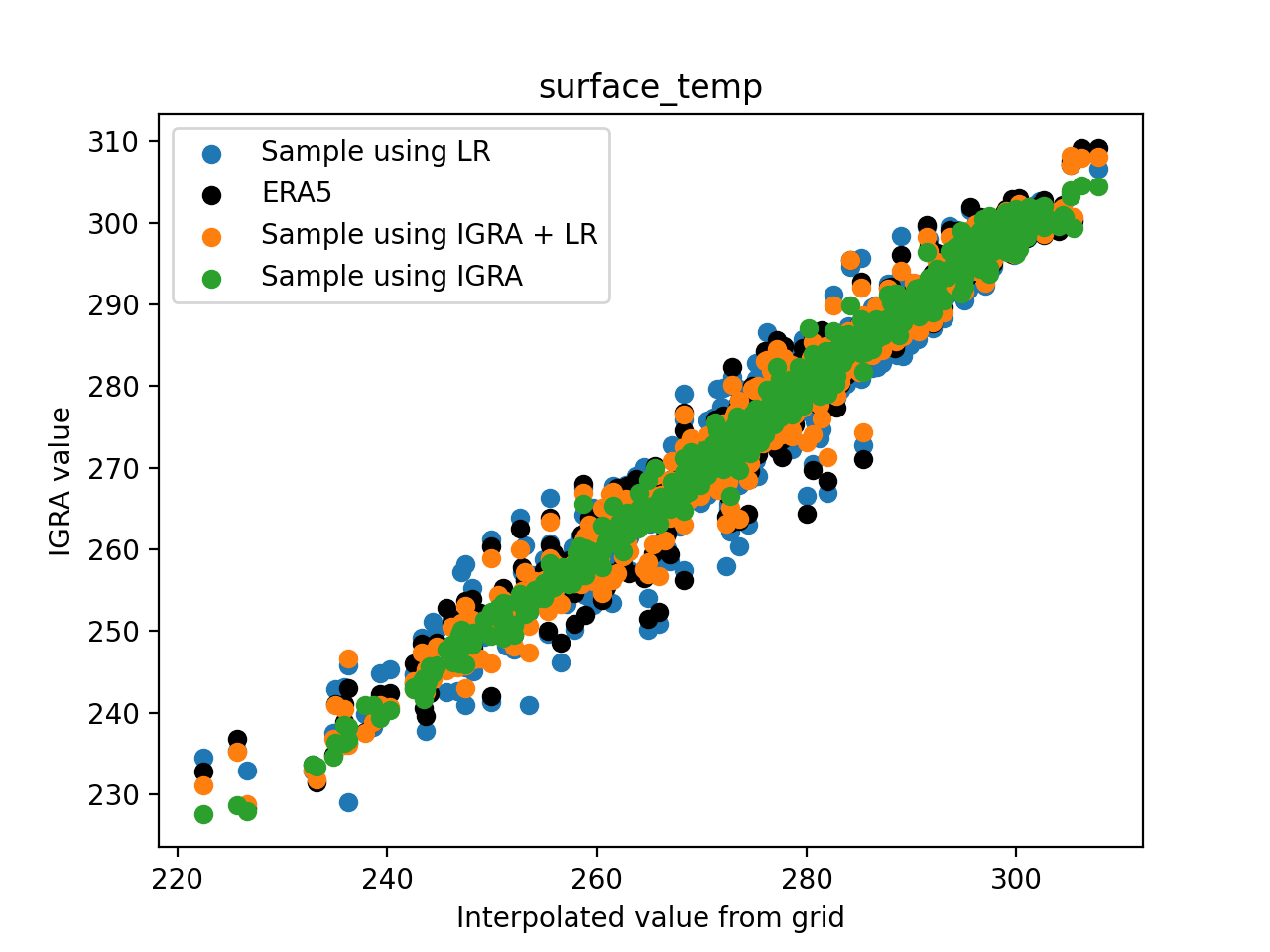}}
    \subfigure[U-Wind]{\includegraphics[width=0.33\textwidth]{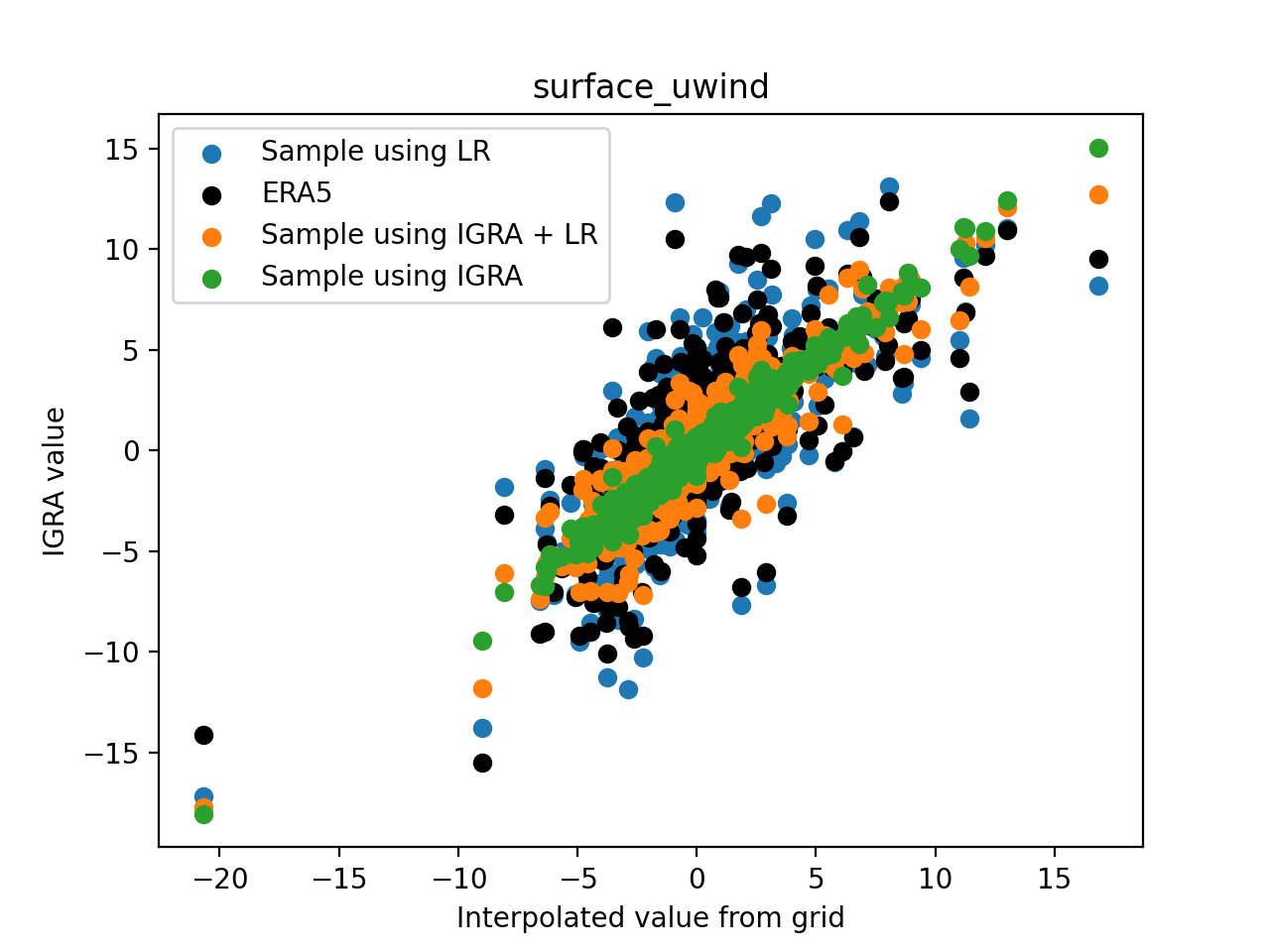}}
    \subfigure[V-Wind]{\includegraphics[width=0.33\textwidth]{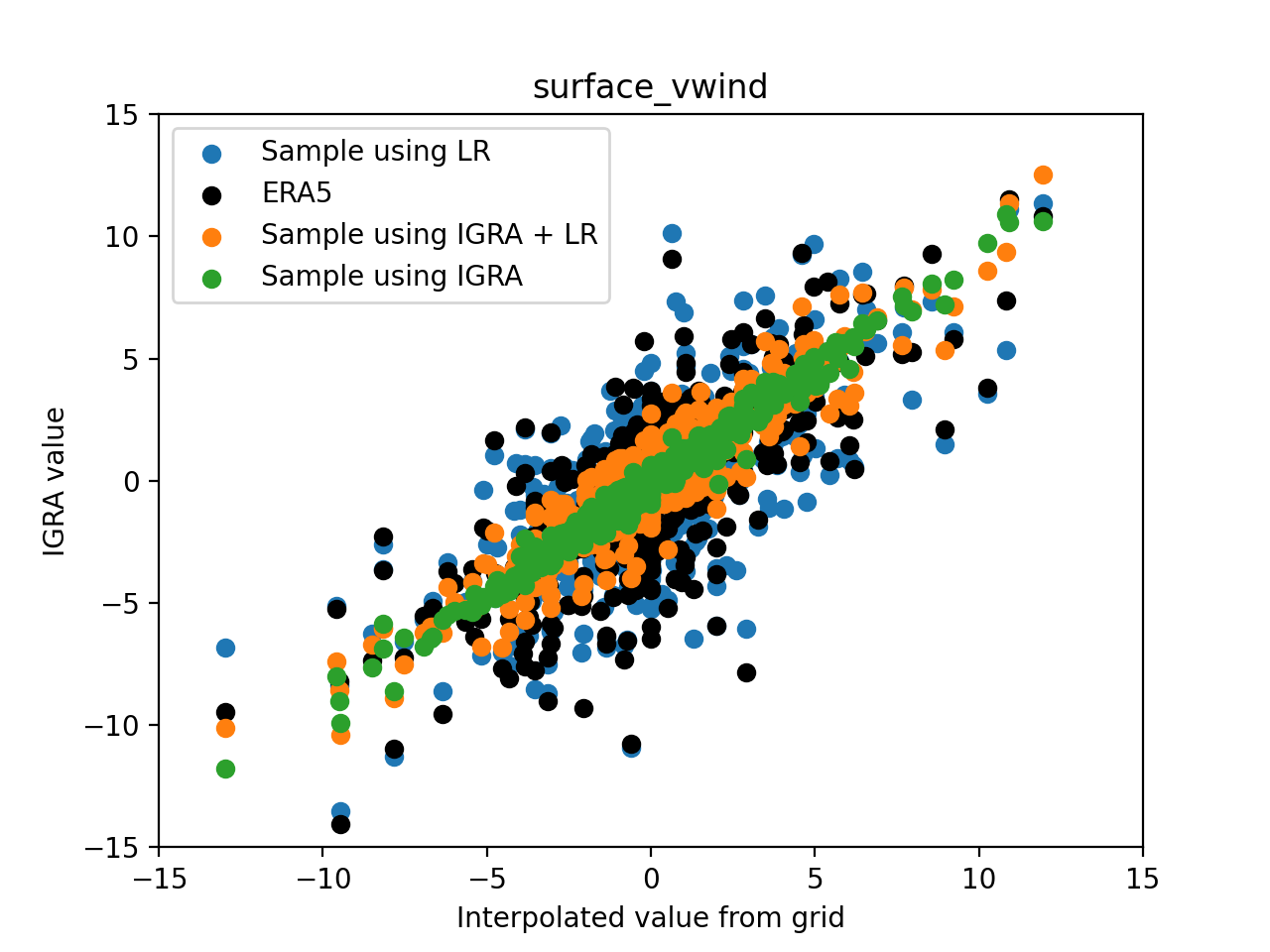}}
    }
    \caption{Scatter, at IGRA sensing locations, from an exemplar reconstruction using zero-shot score matching when using the sample IGRA+LR configuration. At these locations, one can observe a closer agreement to the IGRA values.}
    \label{fig:igra_lr_hr_scatter}
\end{figure}

\begin{figure}[!ht]
    \centering
    \includegraphics[width=0.95\linewidth]{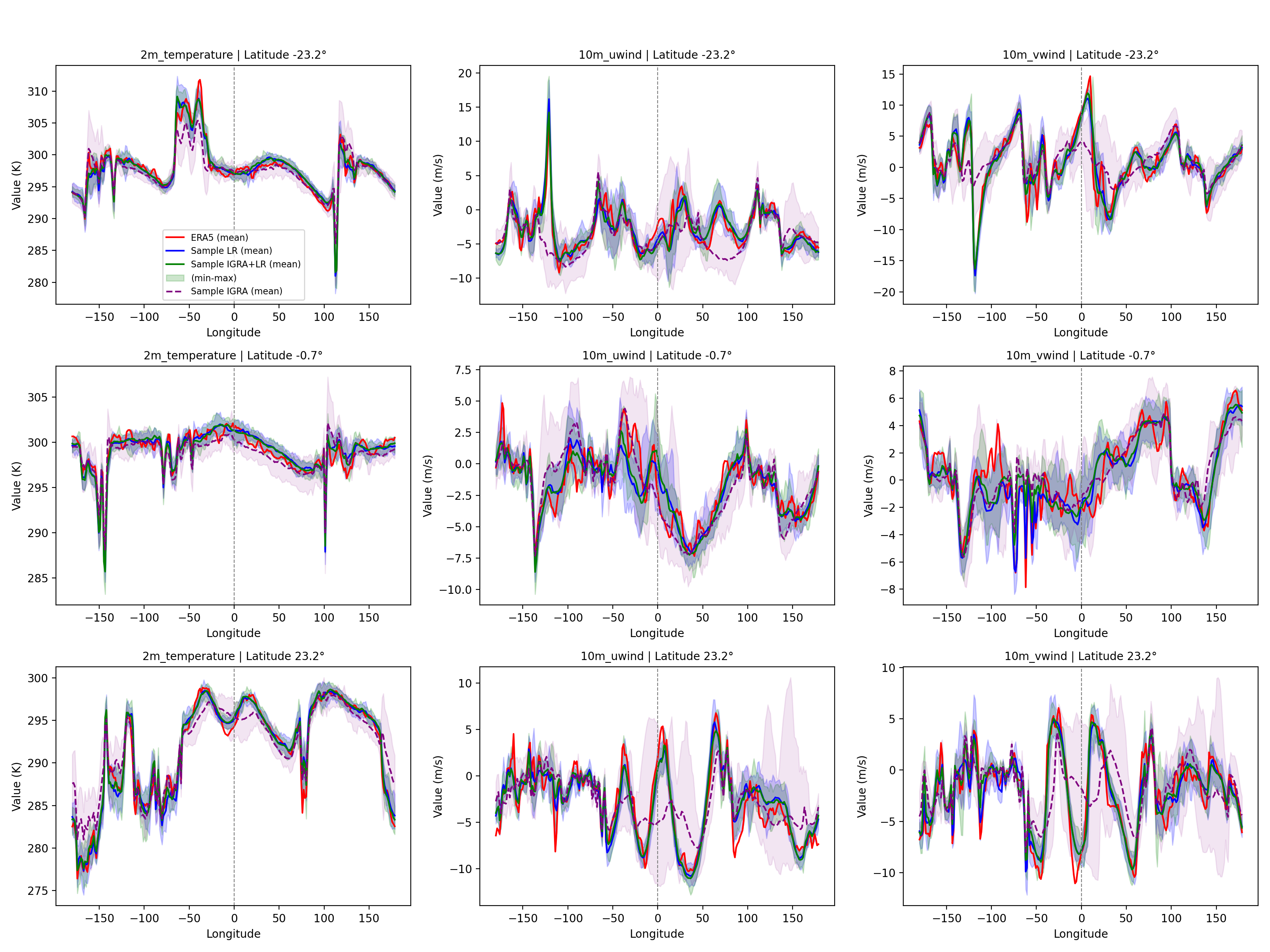}
    \caption{Traces for zero-shot reconstructed flow-fields (including standard deviation as shaded regions) compared with ground truth for state reconstruction with sample-LR, sample-IGRA, and sample-IGRA+LR configurations. The results indicate that the addition of low-resolution ERA5 observations improve those from solely IGRA based reconstructions.}
    \label{fig:igra_lr_hr_lineplot}
\end{figure}

\begin{figure}[!ht]
    \centering
    \mbox{
    \subfigure[Surface Temperature]{\includegraphics[width=0.33\textwidth]{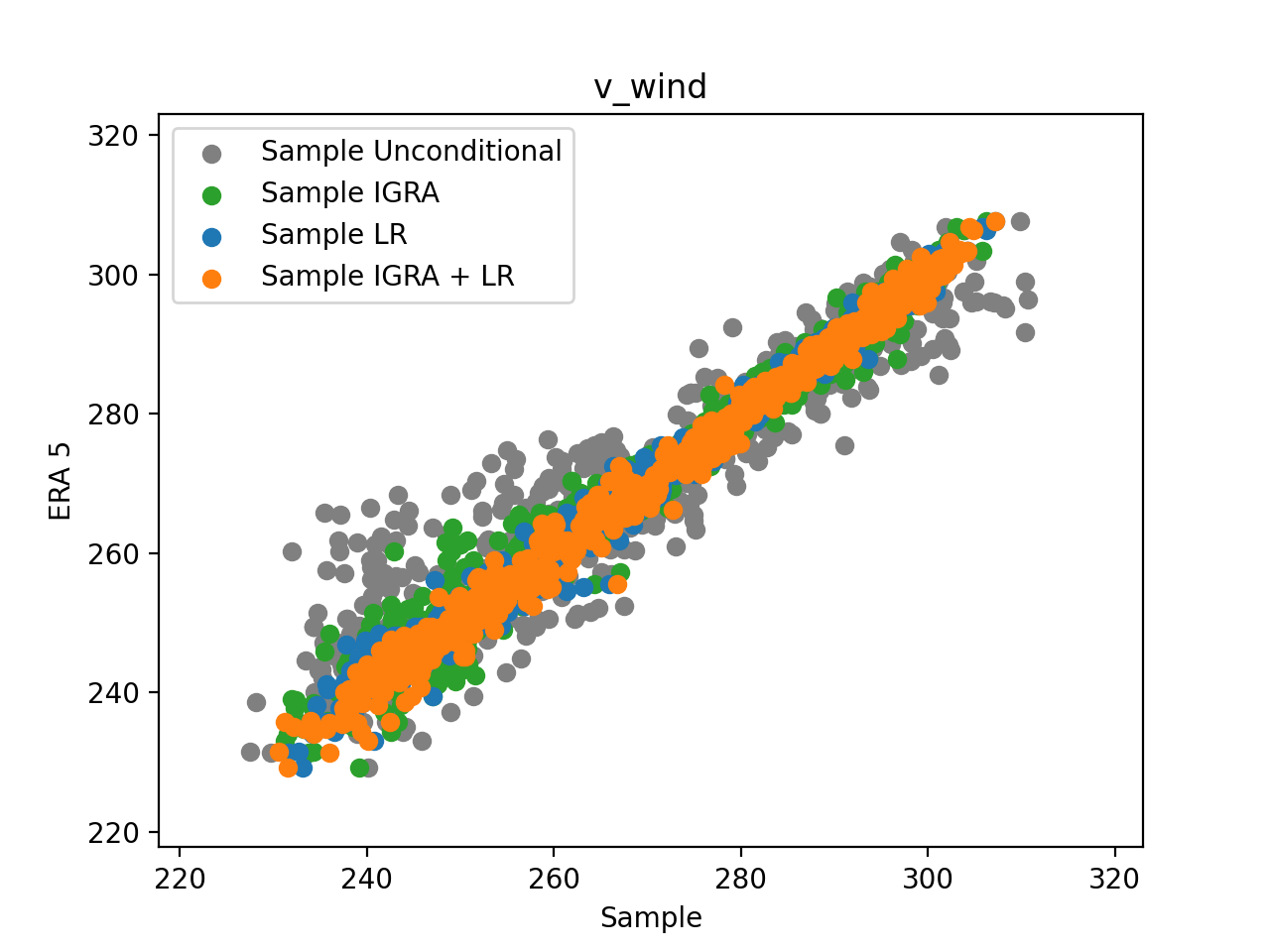}}
    \subfigure[U-Wind]{\includegraphics[width=0.33\textwidth]{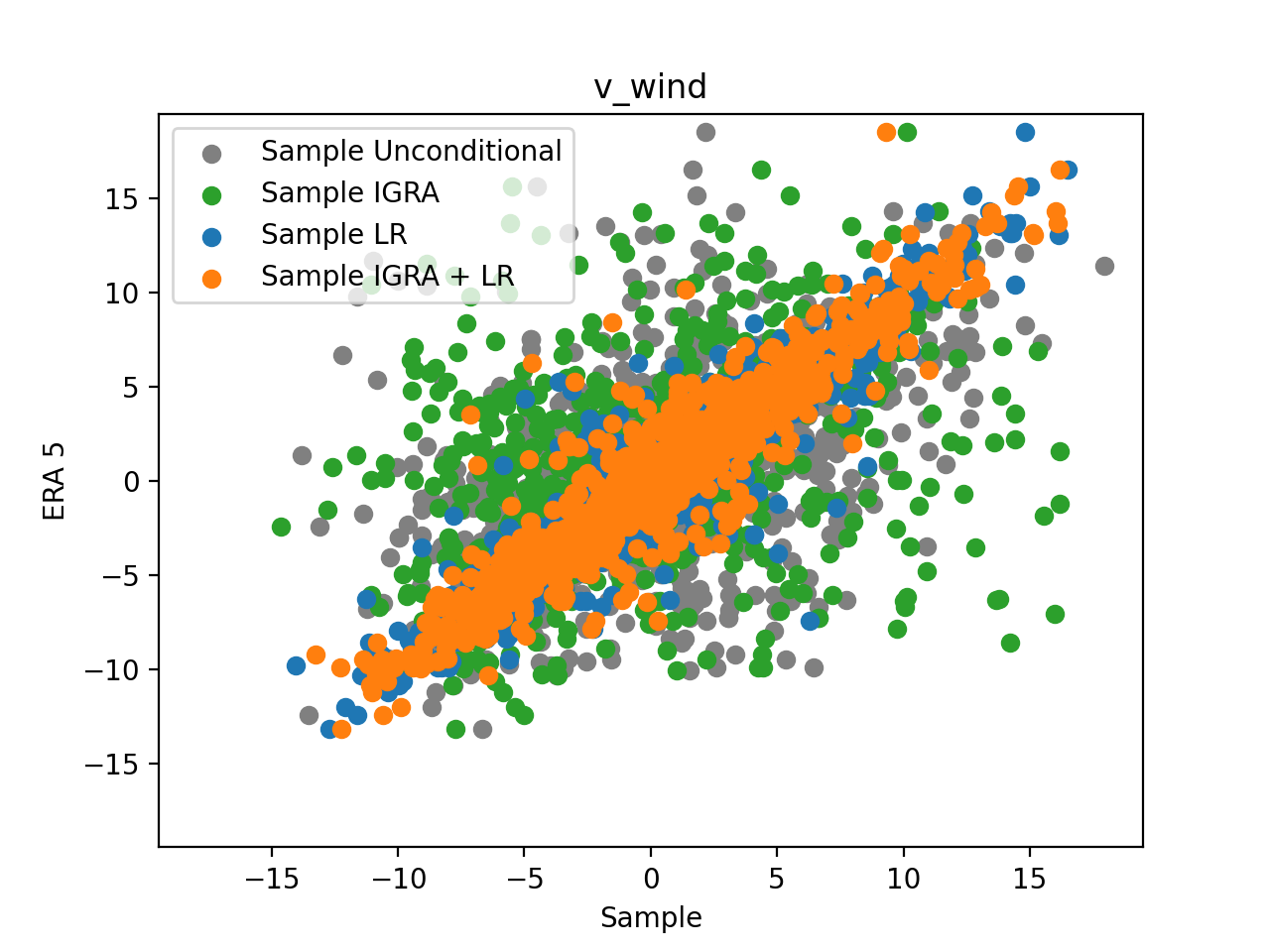}}
    \subfigure[V-Wind]{\includegraphics[width=0.33\textwidth]{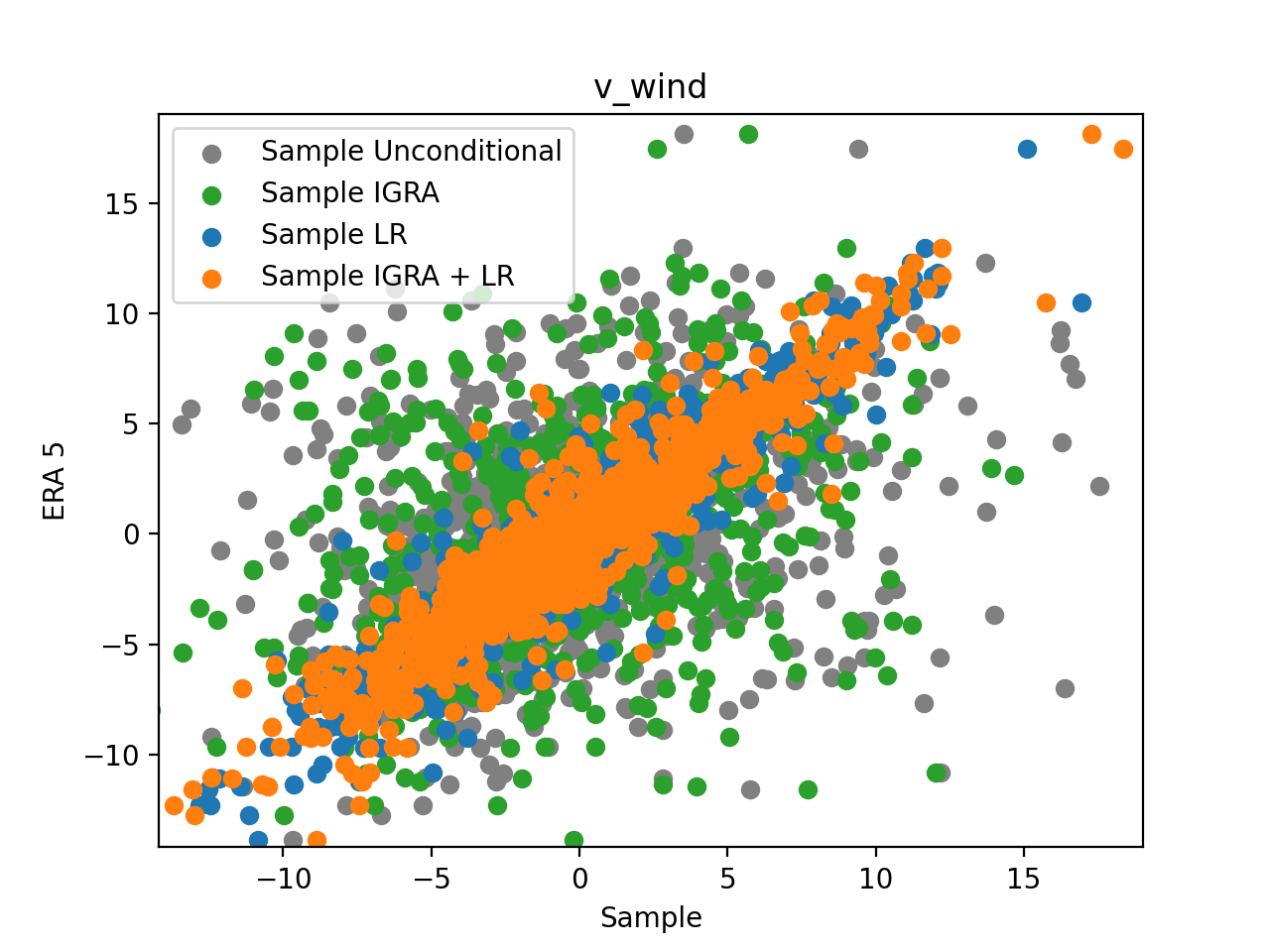}}
    }
    \caption{Scatter, at non-IGRA locations, from an exemplar reconstruction using zero-shot score matching when using the sample IGRA+LR configuration. At these locations, one can observe a closer agreement to the ERA5 values indicate effective multimodal data assimilation.}
    \label{fig:igra_lr_hr_scatter_2}
\end{figure}

In this section, we evaluate one of the greatest strengths of zero-shot sampling from generative models, namely, their ability to assimilate multiple sources of atmospheric data during test time. Therefore, we denote these studies as multimodal super-resolution. Note that modality refers to platforms of measurement as against different data types. First, we perform experiments that rely on observations of both the low-resolution ERA5 observations on a coarse and structured grid, as well as IGRA observations on the unstructured data (previously introduced as the sample-IGRA+LR configuration). In Figure \ref{fig:igra_lr_hr_spec} we show the spectral recovery properties of the multifidelity assimilation versus those obtained solely from IGRA and the structured grid observations. One of the key benefits of using data from the low-resolution ERA5 in addition to IGRA is the significant reduction in uncertainty in the lower wavenumbers for the wind-speed spectra in addition to a correction of the high-wavenumber errors obtained when solely using IGRA. This is also observed in longitudinal traces in Figure \ref{fig:igra_lr_hr_lineplot} where the variance in samples is reduced given more relevant information. When assessing the scatter of the generated values for exemplar state recovery at the IGRA sensor locations, as shown in Figure \ref{fig:igra_lr_hr_scatter}, we observe that the generative model is able to sample in a manner that is closer to the observed IGRA magnitudes despite using both low-resolution ERA5 and IGRA for score-matching. This indicates an ability to balance the influence of various sources of data based on the proximity to their observation locations. This is reinforced by evidence in Figure \ref{fig:igra_lr_hr_scatter_2}, where scatter plots for reconstructions at non-IGRA sensing locations indicate an improved reconstruction of ERA5 now that coarse-grid ERA5 observations are provided during zero-shot score matching. We remark that while using low-resolution observations of reanalysis represents an unrealistic source of data during real-time data fusion, it represents a proof-of-concept of fusing another high-quality source of data to reduce uncertainties during assimilation.
\begin{figure}[!ht]
    \centering
    \mbox{
    \subfigure[2m Temperature (K)]{\includegraphics[trim={0 50 0 50},clip,width=0.5\textwidth]{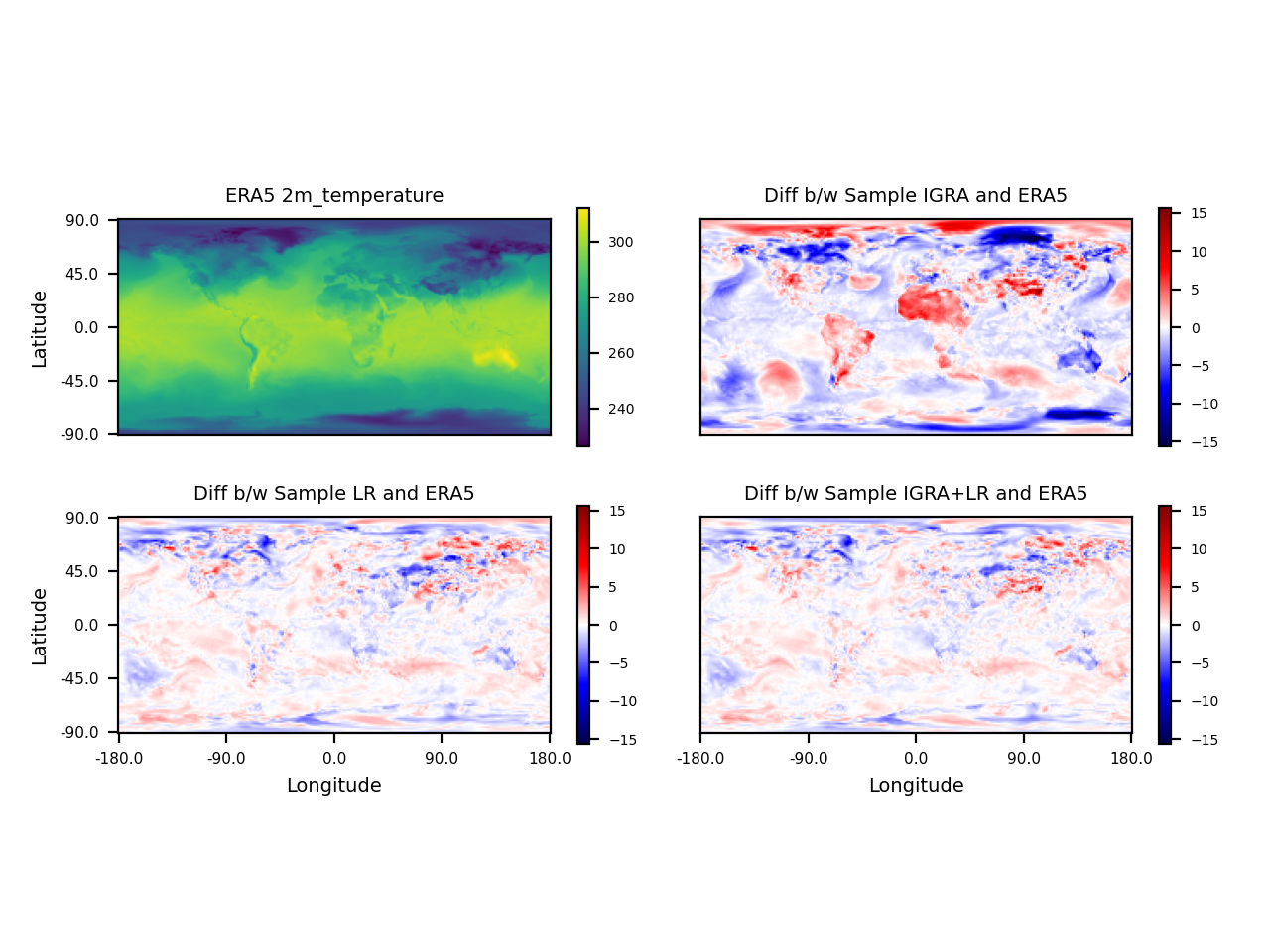}}
    \subfigure[U-Wind (m/s)]{\includegraphics[trim={0 50 0 50},clip,width=0.5\textwidth]{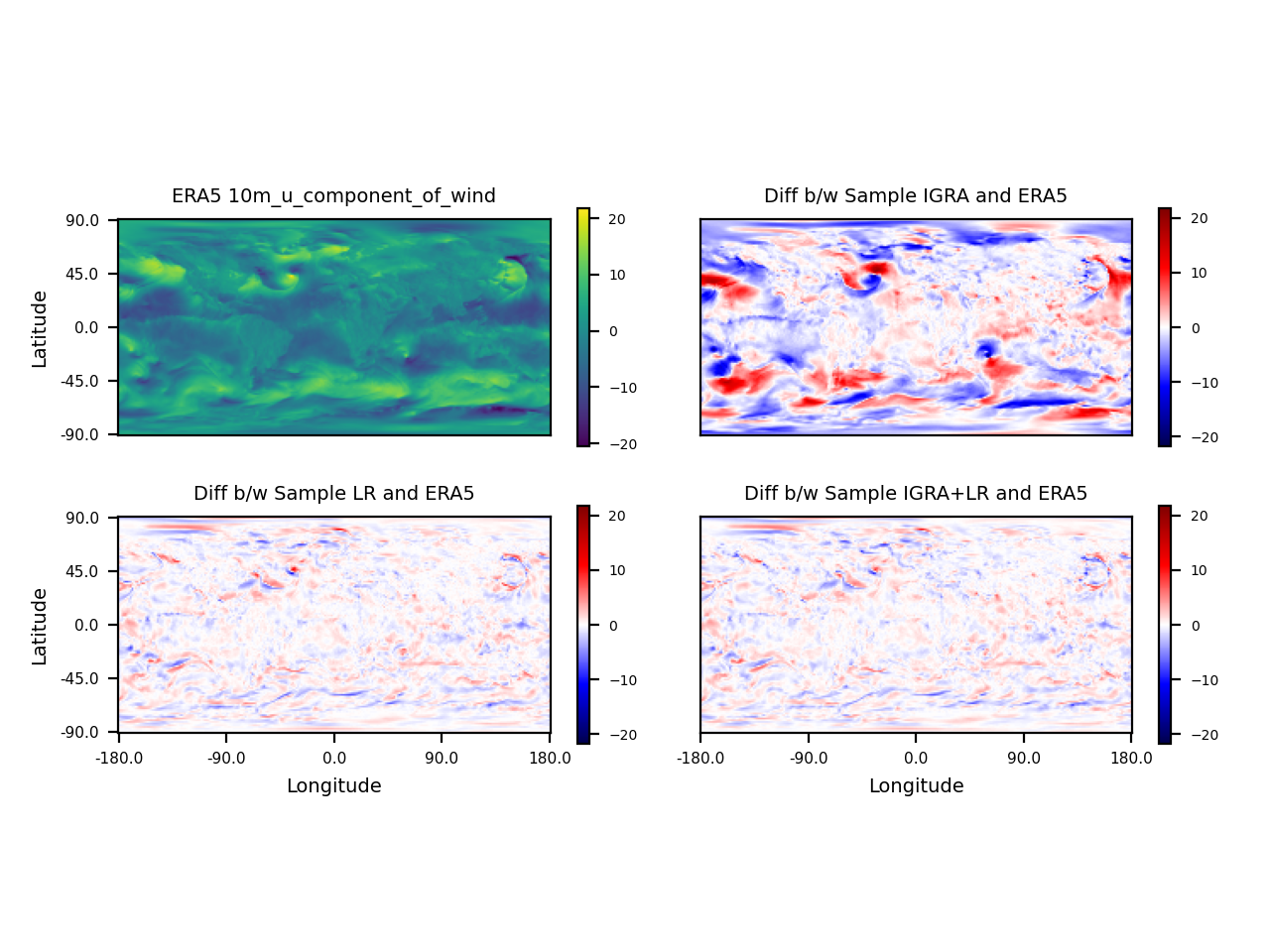}}
    } \\
    \mbox{
    \subfigure[V-Wind (m/s)]{\includegraphics[trim={0 50 0 50},clip,width=0.5\textwidth]{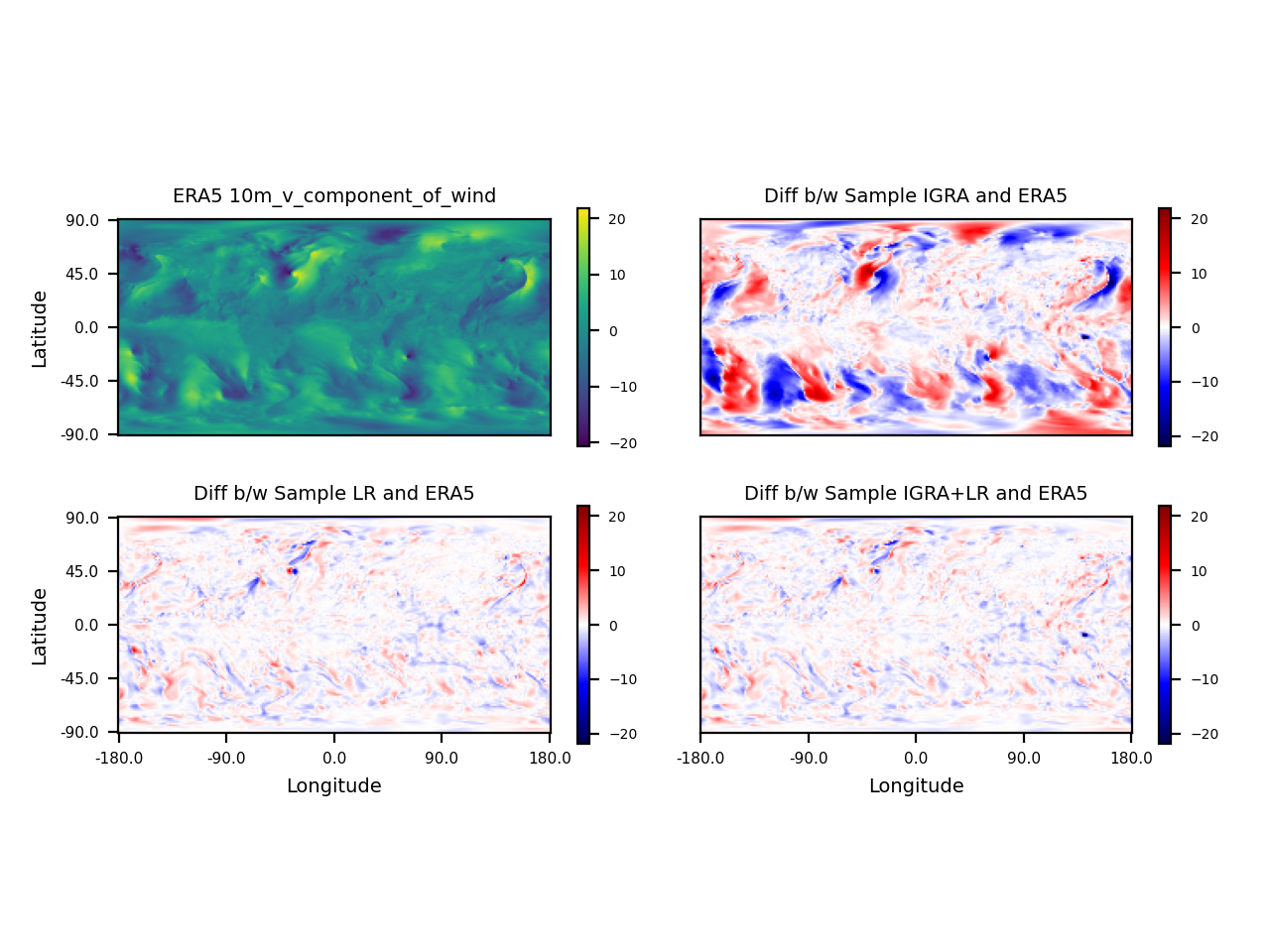}}
    \subfigure[Specific humidity (kg/kg)]{\includegraphics[trim={0 50 0 50},clip,width=0.5\textwidth]{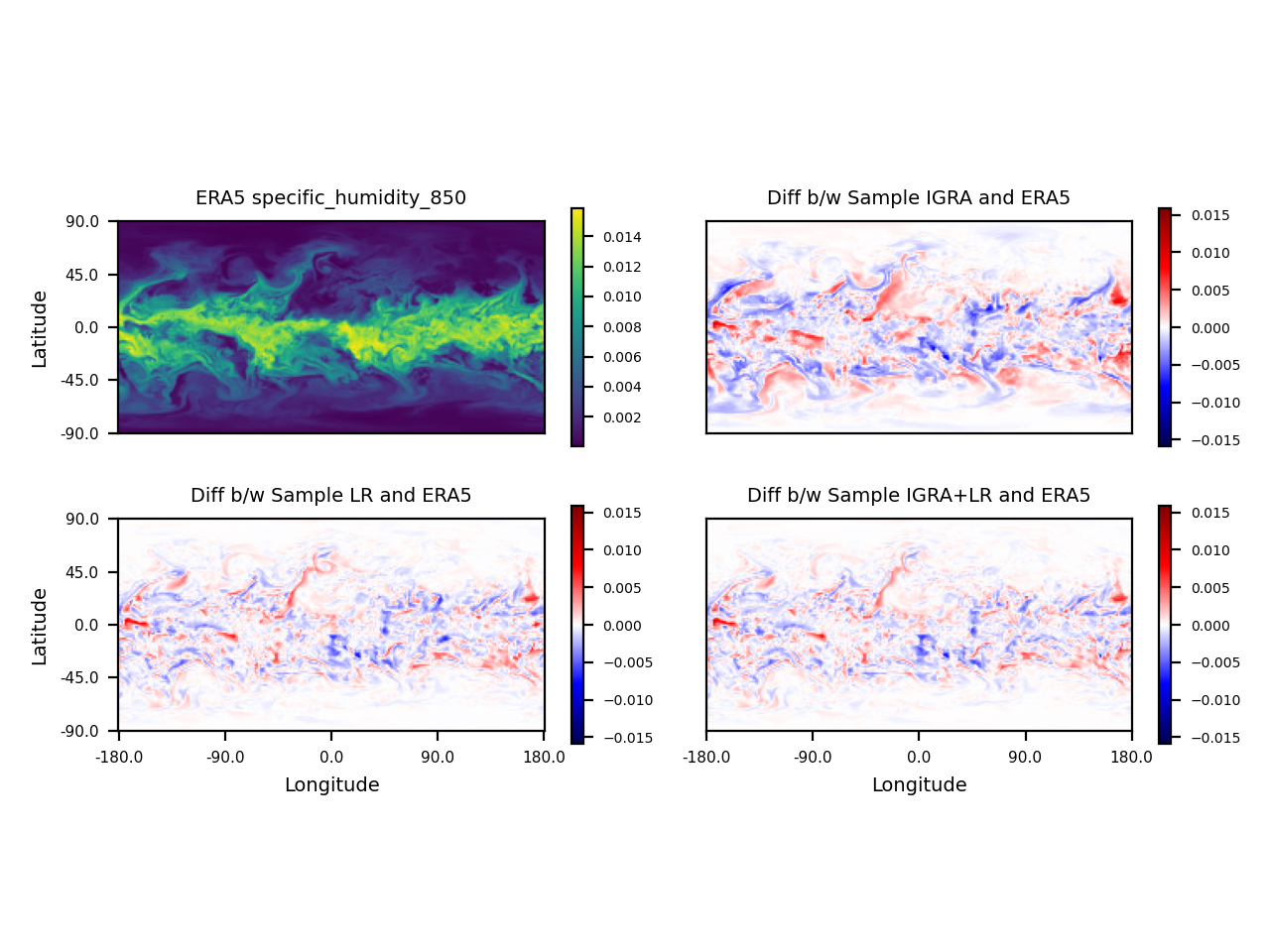}}
    }
    \caption{An assessment of the difference between zero-shot score-based sampling and the ground truth for the various sampling configurations. We observe that the sample-IGRA performance, while competitive for temperature and specific humidity reconstructions, need additional observations from coarse-grid ERA5 for improved wind-speed reconstructions. This provides evidence that combinations can be used for improved reconstructions in zero-shot flow matching.}
    \label{fig:igra_lr_hr_innovation}
\end{figure}

Next, we showcase the model’s ability to assimilate data coming from a dynamical core along with observations coming from a sparse and unstructured data set. For evaluation, LUCIE generates ensembles of 7-day forecasts, initialized at 12:00 UTC on randomly sampled days in 2020, serving as measurements for the EDM model to perform posterior sampling. We also remind the reader that the LUCIE forecasts are provided on a much coarser grid than that of the training data. Data fusion that solely uses LUCIE observations during the sampling process are denoted `LUCIE'. Next, we also use measurements from IGRA dataset in a composite likelihood, as mentioned in Equation \ref{eq:likelihood_Score_lucie}, to perform multimodal data fusion, denoted `IGRA+LUCIE'. We remark that these fusion implementations \emph{do not reinitialize the LUCIE forecast model}. In other words, we simply assimilate precomputed forecasts from a LUCIE deployment on the same initial condition. The hyperparameter $\lambda$ in Equation \ref{eq:likelihood_Score_lucie}, adapts the relative influence of these two datasets. Table \ref{tab:lambda_ablation} shows the existence of an ideal range of $\lambda$ to obtain the optimal RMSE.
\begin{figure}[!ht]
\vspace{-0.4cm}
    \centering
    \mbox{
    \subfigure[IGRA+LUCIE]{\includegraphics[width=0.49\textwidth]{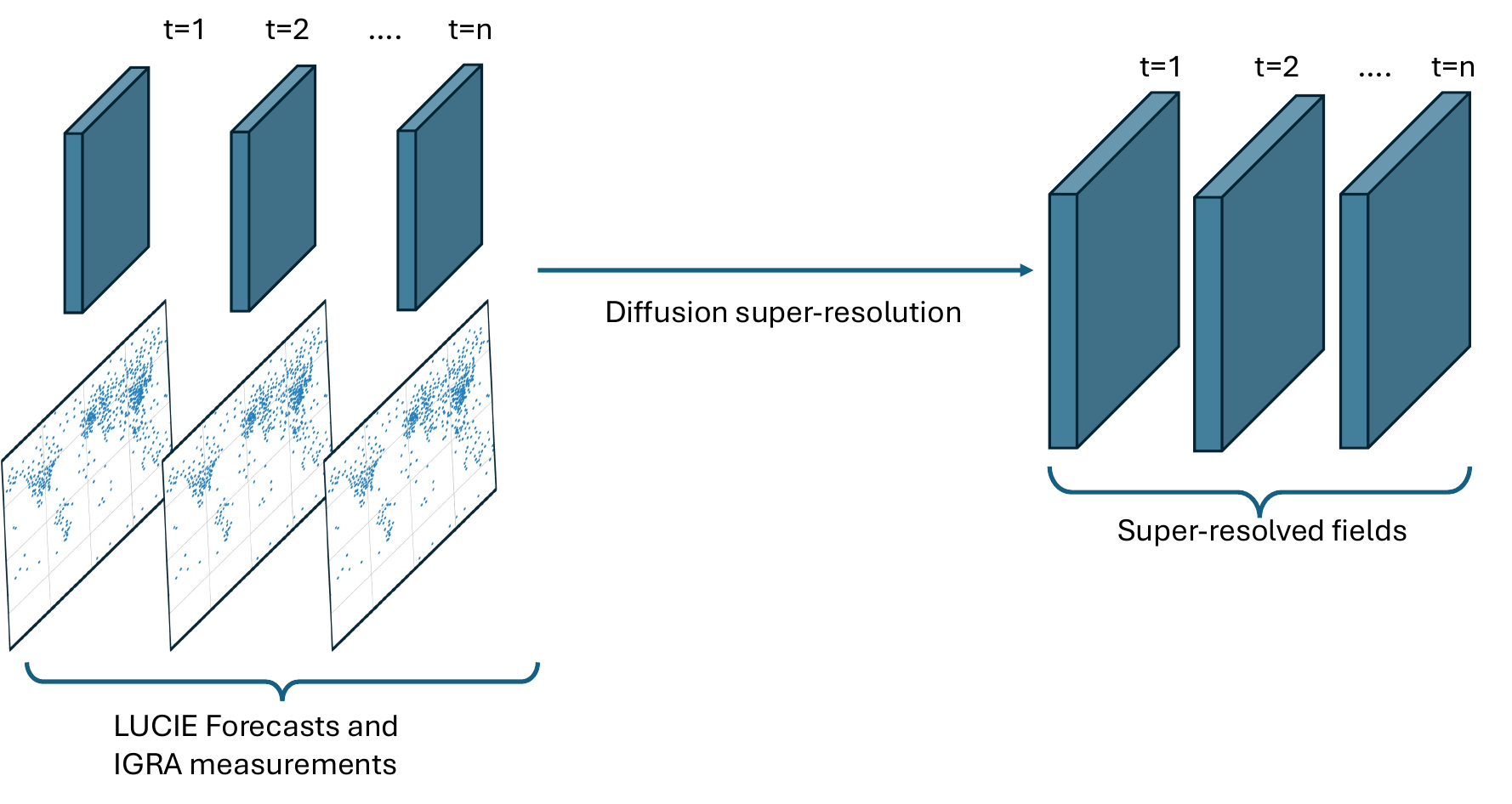}}
    \subfigure[IGRA+LUCIE Reinitialized]{\includegraphics[width=0.49\textwidth]{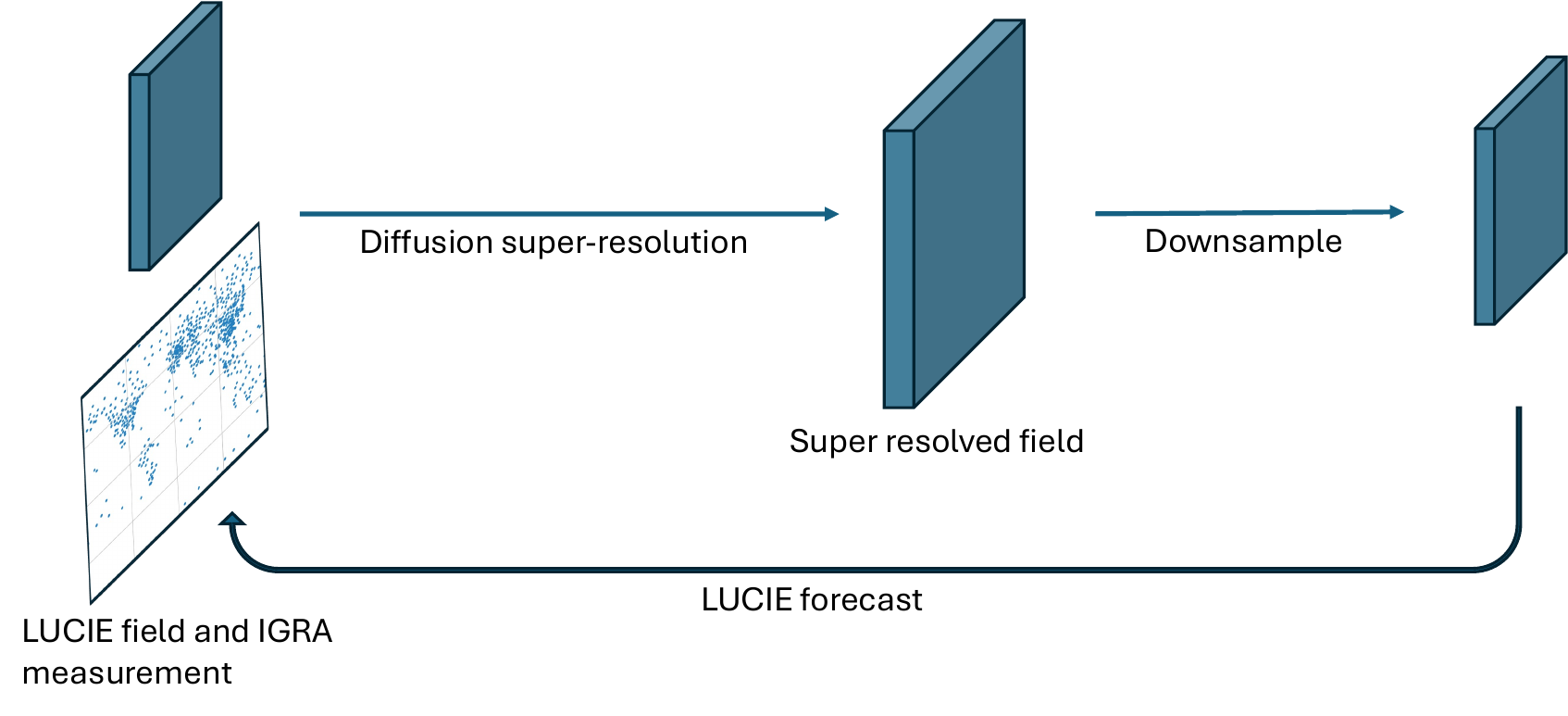}}
    }
    \vspace{-0.1cm}
    \caption{Comparison of IGRA+LUCIE (a) and IGRA+LUCIE Reinitialized (b). In (a), sequences of LUCIE forecasts and measurements from IGRA are fused for super-resolving at each time step independently. In (b), a LUCIE forecast is fused with IGRA measurements and super-resolved, downsampled, and reinitialized for forecast iteratively.}
    \label{fig:lucie_autoreg_schema}
\end{figure}

\begin{figure}[!ht]
    \centering
    \includegraphics[width=0.95\linewidth]{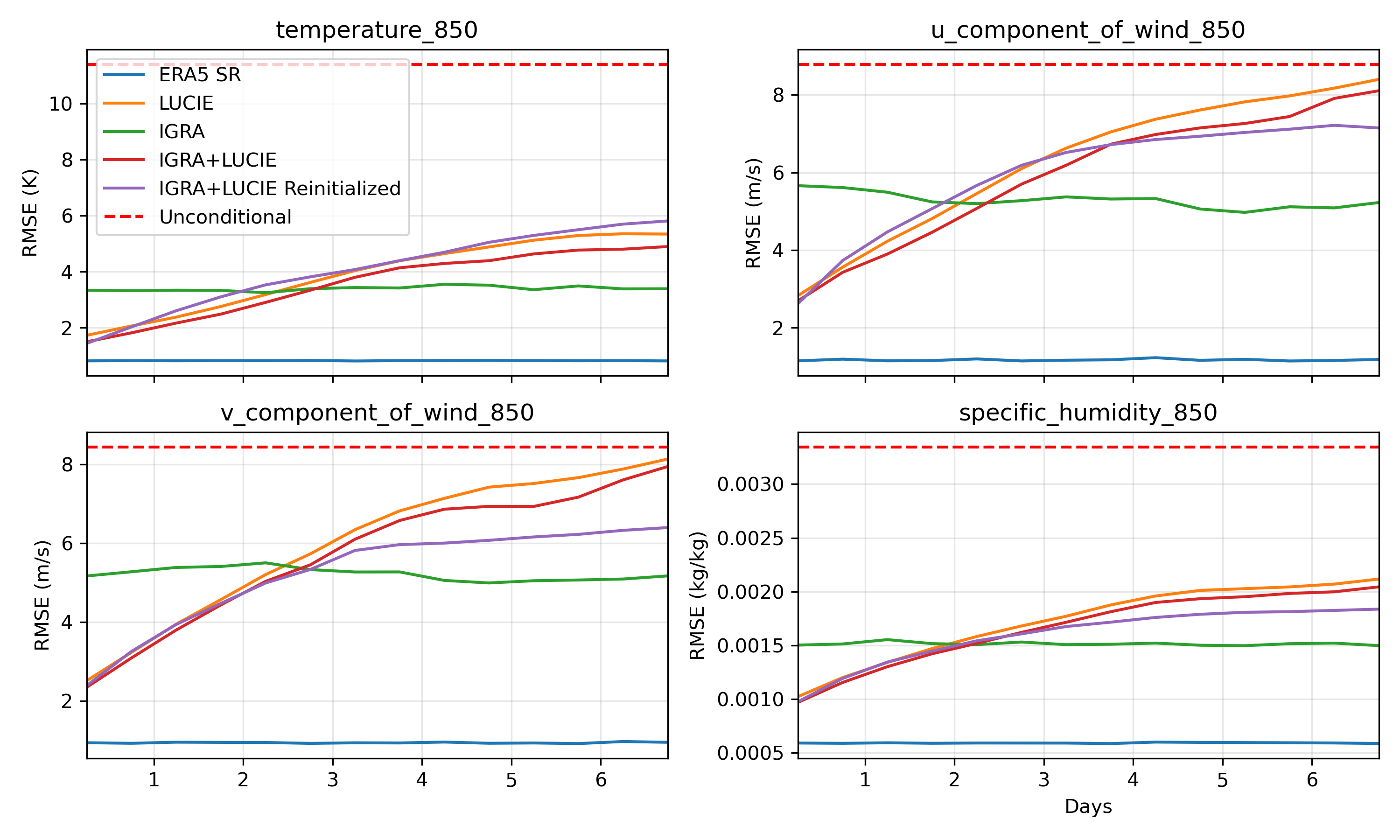}
    \caption{RMSE over 7 days for selected variables. The red dashed line denotes the unconditional baseline that is averaged over 28 random samples. The rest are computed as the RMSE of averaged 16 ensemble members with different initial conditions.}

    \label{fig:lucie_rmse}
\end{figure}
\begin{table}[h!]
\centering
\begin{tabular}{lcccc}
\toprule
$\lambda$ & Day 1 & Day 3 & Day 5 & Day 7 \\
\midrule
0.0 & 1.934 & 3.160 & 4.273 & 4.938 \\
0.5 & 1.678 & 2.828 & 3.919 & 4.241 \\
1.0 & 1.674 & 2.793 & 3.849 & 4.166 \\
2.0 & \textbf{1.599} & \textbf{2.584} & \textbf{3.574} & \textbf{3.825} \\
4.0 & 1.661 & 2.677 & 3.625 & 3.910 \\
\bottomrule
\end{tabular}
\vspace{0.2cm}
\caption{2m temperature RMSE by lead time for different $\lambda$ (Equation \ref{eq:likelihood_Score_lucie}).}
\label{tab:lambda_ablation}
\end{table}

Figure ~\ref{fig:lucie_rmse} shows that sampling through observations of precomputed LUCIE forecasts initially achieves low error but degrades with lead time due to the onset of deterministic chaos. This is an expected trend - we emphasize that observations are not being used to correct our dynamical core, as is common in conventional data assimilation, but a previous forecast from an uncorrected dynamical core is being used to guide the sampling of the posterior of our diffusion model. A true equivalence with data assimilation would need the utilization of these reconstructions as fresh initial conditions of the dynamical core (which we will return to shortly). Next, as observed in Section \ref{sec:IGRA}, sampling with IGRA gives higher accuracy in locations where observations are abundantly available. This causes it to have a relatively constant RMSE but one that is significantly lower than unconditional sampling from the prior. The combined likelihood from the IGRA+LUCIE approach yields improved performance in the sampling, starting closer to LUCIE and showing significant improvement at larger lead time. However, we observe, as shown for when only LUCIE was used in the data fusion process, that growth in emulator errors ultimately causes the reconstructions to have higher errors than merely utilizing IGRA data at each snapshot. We also remark that the fusion step is performed in lock-step with IGRA observations, which are available every 12 hours despite LUCIE being available every 6 hours.

\begin{figure}[!ht]
    \centering
    \includegraphics[width=0.95\linewidth]{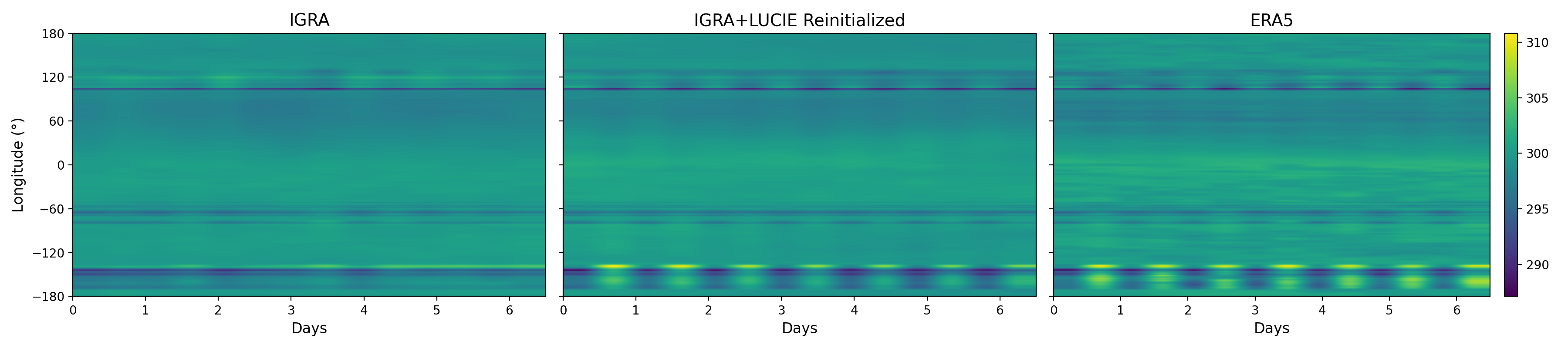}
    \caption{The time series of ensemble mean 2m temperature at 0.7$^{\circ}$ latitude with initial condition at 12:00 UTC January 1st 2020. IGRA+LUCIE Reinitialized is able to recover oscillatory trends in the temperature that are typically missed by IGRA-based reconstructions alone.} 
    \label{fig:p7_time}
\end{figure}

Finally, we introduce a data fusion formulation that implements a reinitialization for LUCIE for each timestep of a super-resolution being performed sequentially in time, denoted `IGRA+LUCIE Reinitialized'. Figure \ref{fig:lucie_autoreg_schema} represents a schematic that contrasts the previous approach to incorporating LUCIE forecasts, versus what is now proposed, where the super-resolution via diffusion sampling is used to reinitialize the LUCIE dynamical core for a one step forecast. This setup represents a scenario that is analogous to classical atmospheric data assimilation, with the important caveat that the covariance matrix is pre-specified for rapid inverse computation. In Figure \ref{fig:p7_time}, we observe that the proposed approach leads to improved or similar performance later in the forecast horizon when compared to solely using precomputed LUCIE or IGRA+LUCIE forecasts. In terms of sampling times mentioned in Table \ref{tab:run_times}, we noticed the inclusion of IGRA increases the sampling time. However, we do not detect any major differences between sampling times of IGRA, IGRA+LUCIE and IGRA+LUCIE Reinitialized. We hypothesize further improved gains, at the cost of reduced computational efficiency, if covariances are updated and inverted for each reinitialization step, but we leave this for a future work due to its computational intractability.

\begin{table}[!h]
\centering
\begin{tabular}{lcc}
\hline
\textbf{Method} & \textbf{Mean Time (s)} & \textbf{Std (s)} \\
\hline
LUCIE & 1.995 & 0.00007 \\
IGRA & 2.298 & 0.00504 \\
IGRA+LUCIE & 2.298 & 0.00073 \\
IGRA+LUCIE Reinitialized & 2.293 & 0.00061\\
\hline
\end{tabular}
\caption{Sampling run times per ensemble member (mean and standard deviations) for different posterior sampling methods.}
\label{tab:run_times}
\end{table}

The RMSE of an unconditional model and the low-resolution ERA5 serve as the the two extremes for our assessments: the unconditional being the worst case and the low-resolution ERA5 providing the best reconstructions. Furthermore, to gauge the temporal consistency of our results, we plot the ensemble mean of 2m temperature at particular latitudes with respect to lead time in Figure \ref{fig:p7_time}. We observe at latitude 0.7$^\circ$ and longitudes from -120 to -180 $^\circ$ that IGRA+LUCIE samples perfectly match the temporal trend with ERA5 as against reconstructions that solely utilize IGRA. This shows that AI-based emulators can provide valuable dynamical information for temporal consistency in independently reconstructed samples from sparse observations.

\section{Discussion and Conclusion}

Rapid fusion of observations from different, real-time platforms has the potential to provide significant gains in the accurate reconstruction of high-dimensional dynamical systems. The motivation of this study was to propose avenues for reducing the significant computational costs for achieving the same when using classical numerical methods for data assimilation. To that end, this study proposed the construction of an unconditional generative model that is trained to match the score of the underlying probability distribution from which arbitrary training data may be sampled. Once this score is approximated with a neural network, a zero-shot score-matching technique built on a Bayesian formulation can be used to rapidly assimilate sparse multifidelity observational data with negligible real-time costs. In this study, we first construct an experiment to recover the full state of a reanalysis dataset given partial observations from a coarse-grid observation of the same. This represents an assessment of the method's ability to recover reanalysis when shown partial information. Next, we perform experiments to assess how the score-matching can recover reanalysis when shown information from a real-world radiosonde dataset as well as snapshots from an atmospheric emulator. These experiments indicate that our proposed method can balance the influence of the various observations during the reanalysis reconstruction process. Finally, we perform an experiment that is conceptually closer to real-world atmospheric data assimilation, where the instantaneous reconstruction of the atmosphere is re-utilized as an initial condition for the emulator forecast. These results show that improved gains in accuracy may be obtained when a dynamical core is reinitialized for a one-step forecast in a sequential data fusion task. Specifically, we note that the best practices of atmospheric data assimilation, such as improved covariance matrices can lead to significant gains on top of the proof-of-concept proposed here. However, we note that offline precomputation of the forecasts from a dynamical core also provide competitive results at much reduced computational costs. More importantly, to the best of our knowledge, this work represents a combined data assimilation and super-resolution task that allows for the use of observations from various dynamical models and observations on varying grids. When compared to adjoint-based methods, our approach does not require the computation of gradients through a multistep dynamical core roll-out. When compared with ensemble methods, our approach allows for the assimilation of dynamical predictions on a different grid from the final output which is that of the training data resolution. 

In all experiments, we observe that the score-matching technique generates reconstructions for the atmosphere that outperform an unconditional sampling from the pretrained diffusion model indicating an information gain due to the fusion process. This indicates that zero-shot score matching can be used as an effective and computationally inexpensive tool to accurately perform data fusion from data of varying sources and spatiotemporal fidelity. Based on the promising results from this study, future extensions of this framework will include the integration of additional dynamical considerations for diffusion-based forecasting, i.e., performing lead-time-dependent score-matching with a diffusion model that generates trajectories, integrating both Eulerian and Lagrangian observations within a common fusion paradigm, and optimal sensor placement for high-quality state recovery.

 \section*{Acknowledgments}

 This research used resources of the Argonne Leadership Computing Facility, which is a U.S. Department of Energy Office of Science User Facility operated under contract DE-AC02-06CH11357. DC, HG, and RM acknowledge support from DOE ASCR award "Inertial neural surrogates for stable dynamical prediction". DC, HG, RM also acknowledge the support of computing resources through an AI For Science allocation at the National Energy Research Scientific Computing Center (NERSC) and from Penn State Institute for Computational and Data Sciences (ICDS). RM acknowledges the support of a YIP from ARO Modeling of Complex Systems (PM: Robert Martin).

%Bibliography
\bibliographystyle{unsrt}  
\bibliography{references}

\end{document}